\documentclass[a4paper,fleqn]{cas-dc}

\usepackage[numbers]{natbib}
\usepackage{balance}
\usepackage[]{mdframed}
\usepackage{hyperref}
\usepackage[labelformat=simple]{subcaption}

\DeclareCaptionLabelFormat{subcaptionlabel}{\normalfont(\textbf{#2}\normalfont)}
\def\tsc#1{\csdef{#1}{\textsc{\lowercase{#1}}\xspace}}
\tsc{WGM}
\tsc{QE}
\tsc{EP}
\tsc{PMS}
\tsc{BEC}
\tsc{DE}
\begin{document}
\let\WriteBookmarks\relax
\def\floatpagepagefraction{1}
\def\textpagefraction{.001}
\shorttitle{ML-Driven Performance Optimization of Tunable HTL-Free PSCs with MWCNT}
\shortauthors{Ihtesham et~al.}

\title [mode = title]{Simultaneous Optimization of Efficiency and Degradation in Tunable HTL-Free Perovskite Solar Cells with MWCNT-Integrated Back Contact Using a Machine Learning-Derived Polynomial Regressor}

\author[1]{Ihtesham Ibn Malek}[%type=editor,
                        %auid=000,bioid=1,
                       % role=Researcher,
                        orcid=0009-0003-6747-0189
                        ]

%\cormark[1]


\ead{shanto.bin.malek@gmail.com}

\begin{comment}
\ead[url]{www.cvr.cc, cvr@sayahna.org}
\end{comment}

\credit{Conceptualization, Data curation, Formal analysis, Methodology, Software, Visualization, Writing – original draft, Writing – review and editing}

\affiliation[1]{organization={Department of Electrical and Electronic Engineering, Bangladesh University of Engineering and Technology},
    %addressline={}, 
    city={Dhaka},
    % citysep={}, % Uncomment if no comma needed between city and postcode
    postcode={1205}, 
    country={Bangladesh}}

\author[1]{Hafiz Imtiaz}

\ead{hafizimtiaz@eee.buet.ac.bd}

\credit{Data curation, Investigation, Resources, Supervision, Validation, Visualization, Writing – review and editing}

\author[1]{Samia Subrina}[%
   %role=Co-ordinator,
   ]

\cormark[1]

\ead{samiasubrina@eee.buet.ac.bd}
%\ead[URL]{www.sayahna.org}

\credit{Data curation, Investigation, Resources, Supervision, Validation, Visualization, Writing – review and editing}

\cortext[cor1]{Corresponding author}
%\cortext[cor2]{Principal corresponding author}

\begin{abstract}
Perovskite solar cells (PSCs) without a hole transport layer (HTL) offer a cost-effective and stable alternative to conventional architectures, utilizing only an absorber layer and an electron transport layer (ETL). This study presents a machine learning (ML)-driven framework to optimize the efficiency and stability of HTL-free PSCs by integrating experimental validation with numerical simulations. Excellent agreement is achieved between a fabricated device and its simulated counterpart at a molar fraction \( x = 68.7\% \) in \(\mathrm{MAPb}_{1-x}\mathrm{Sb}_{2x/3}\mathrm{I}_3\), where MA is methylammonium. A dataset of 1650 samples is generated by varying molar fraction, absorber defect density, thickness, and ETL doping, with corresponding efficiency and 50-hour degradation as targets. A fourth-degree polynomial regressor (PR-4) shows the best performance, achieving RMSEs of 0.0179 and 0.0117, and \( R^2 \) scores of 1 and 0.999 for efficiency and degradation, respectively. The derived model generalizes beyond the training range and is used in an L-BFGS-B optimization algorithm with a weighted objective function to maximize efficiency and minimize degradation. This improves device efficiency from 13.7\% to 16.84\% and reduces degradation from 6.61\% to 2.39\% over 1000 hours. Finally, the dataset is labeled into superior and inferior classes, and a multilayer perceptron (MLP) classifier achieves 100\% accuracy, successfully identifying optimal configurations.
\end{abstract}

% \includegraphics[width=\textwidth]{figs/GrAbs.jpg}

% \begin{highlights}  
% \item A close match with experimental result is observed at \( x = 68.7\% \) in MAPb\(_{1-x}\)Sb\(_{2x/3}\)I\(_3\), validating numerical models.  
% \item A dataset of 1650 samples is generated by varying four key fabrication parameters with their corresponding efficiency and degradation.  
% \item 4th degree polynomial regressor achieves RMSEs of 0.0179, 0.0117, and \( R^2 \) of 1, 0.999 for efficiency and degradation predictions.  
% \item Efficiency improves from 13.7\% to 16.84\%, while degradation reduces from 6.61\% to 2.39\% over 1000 hours.  
% \item The ML-guided approach enhances both efficiency and stability, improving the commercial viability of HTL-free PSCs.  
% \end{highlights}

\begin{keywords}
HTL-free PSCs \sep 4th degree polynomial regressor (PR-4) \sep L-BFGS-B method \sep MLP \sep Superior performance
\end{keywords}

\maketitle

\flushbottom

\section{Introduction}

Solar energy is increasingly recognized as a crucial component in the transition toward sustainable and renewable energy sources \cite{saleh2024challenges, maka2022solar, izam2022sustainable}. It offers immense potential for reducing dependence on fossil fuels, mitigating environmental impacts, and providing clean, accessible power \cite{maka2022solar, algarni2023contribution}. However, fluctuations in power system parameters caused by disturbances, weather patterns, and other natural factors pose challenges to maintaining grid stability and reliability \cite{ward2013effect, malek2020gic, rahman2021gic}. Solar energy technology, with its inherently low inertia, can help address these challenges by enabling faster response and dynamic grid support \cite{ahmed2023dynamic, makolo2021role}. For large-scale deployment, developing efficient and stable solar cells is essential to ensuring reliable power generation \cite{li2020towards} and seamless grid integration \cite{oshilalu2024innovative}.

Given the growing interest in renewable energy, perovskite solar cells (PSCs) have gained considerable attention in recent years due to their high power conversion efficiencies (PCEs) 
\cite{malek2024machine, gratzel2017rise}, low manufacturing costs \cite{li2024manufacturing, rong2018challenges}, and flexible fabrication processes \cite{yang2019recent}. One of the key advantages of PSCs is their tunable properties, which make them highly adaptable to a variety of applications \cite{jeon2014solvent}. Perovskite materials, particularly those in the MAPb\(_{1-x}\)Sb\(_{2x/3}\)I\(_3\) family, offer the ability to fine-tune the band structure by adjusting the composition parameter \(x\). This parameter controls the ratio of methylammonium lead iodide (MAPbI\(_3\)) to antimony (Sb) in the perovskite structure. By varying \(x\), the electronic properties of the material, such as the bandgap, electron affinity, etc., can be significantly influenced. Adjusting \(x\) shifts the band structure, enabling the material to absorb sunlight more effectively across different parts of the spectrum. This adjustment flexibility in designing these materials enables the PSCs to operate with improved efficiency in diverse environmental conditions \cite{miah2024band}.

A promising advancement in PSC design is the development of hole transport layer (HTL)-free devices \cite{zhou2020highly}. HTLs, typically used in conventional PSCs, are responsible for transporting holes from the perovskite layer to the electrode, but they also introduce additional complexity and cost to the device structure \cite{tress2014role}. HTL-free PSCs offer several benefits, including reduced fabrication costs, simplified device architecture, and potentially improved stability \cite{ma2024hole}. The elimination of the HTL can minimize the number of interfaces that might otherwise lead to degradation, such as interface recombination, ion migration, and chemical instability. By reducing these interfaces, HTL-free PSCs can achieve better long-term stability and reduce the risk of performance degradation under real-world operating conditions \cite{al2020engineered}.

\begin{figure*}
	\centering
	\includegraphics[width=.7\textwidth]{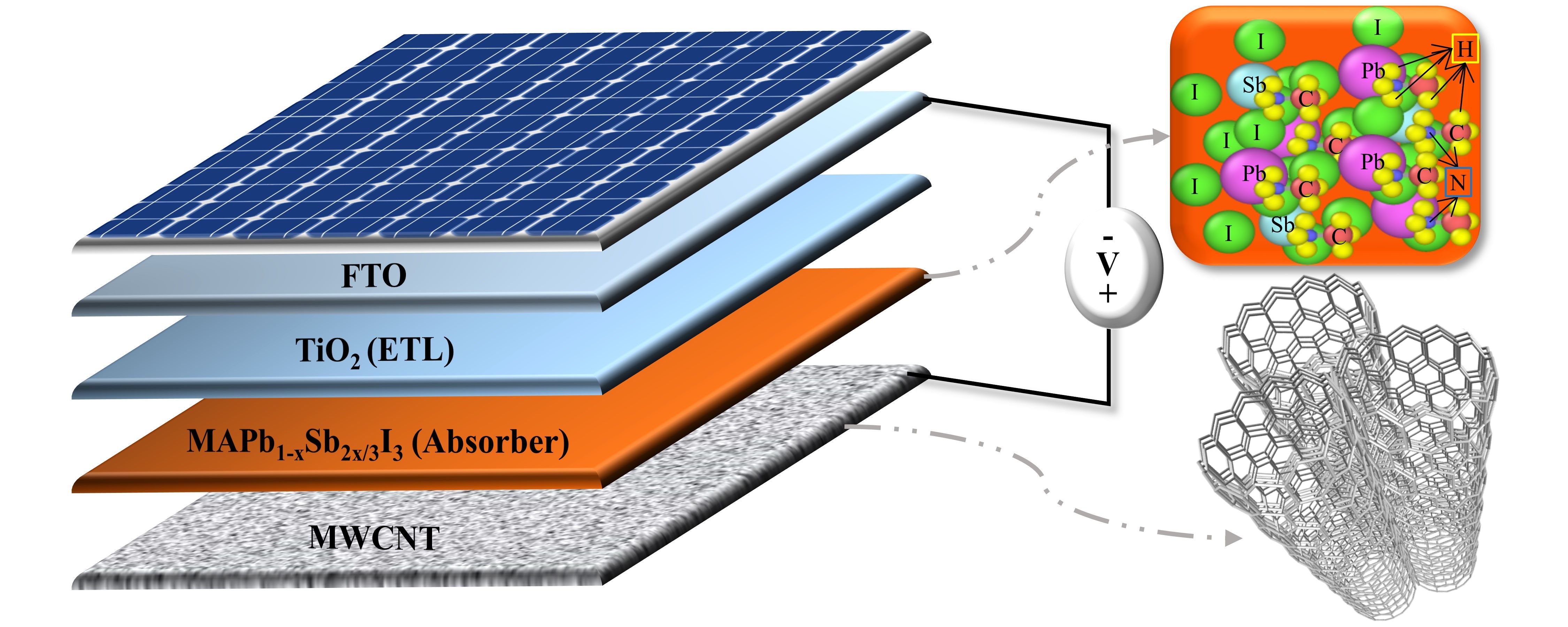}
	\caption{Perovskite solar cell structure used in this study.}
	\label{FIG:1}
\end{figure*}

The electron transport layer (ETL), plays a critical role in PSCs by facilitating the efficient transport of electrons from the perovskite layer to the external circuit \cite{pan2020advances}. Titanium dioxide (TiO\(_2\)) is widely used as an ETL due to its excellent electron mobility, high stability under light and heat exposure, and its ability to form a good interface with the perovskite absorber. As the ETL, TiO\(_2\) helps ensure high performance and minimizes charge recombination \cite{fatima2024critical}. In the back contact of PSCs, Multi-Walled Carbon Nanotubes (MWCNT) are often incorporated to further enhance charge collection and conductivity \cite{mohammed2023improving}. MWCNTs are highly conductive, possess a large surface area, and exhibit excellent mechanical properties, making them ideal for improving the interface between the perovskite absorber and the electrode \cite{zheng2017boron}.

Despite the promising performance of PSCs, their commercialization is hindered by stability issues, particularly under prolonged exposure to environmental stressors, such as moisture, oxygen, and temperature variations \cite{li2020towards}. Degradation mechanisms, including ion migration and chemical instability, can significantly reduce their lifespan and efficiency, making stability a critical factor for large-scale commercialization \cite{mazumdar2021stability}. Both PCE and stability must be optimized simultaneously to improve performance and long-term reliability. Traditional methods for enhancing stability, such as material modification \cite{garcia2018benchmarking, mammeri2023paths}, encapsulation \cite{matteocci2016encapsulation}, and non-systematic device architecture optimization \cite{grancini2017one}, have provided some solutions. However, these approaches often come with trade-offs, increasing production costs or complicating the manufacturing process, which limits the scalability and market adoption of PSCs.

In this context, machine learning (ML) has emerged as a powerful tool to address the challenges of performance optimization and stability enhancement. ML models can analyze large datasets, relating material properties, device configurations, and performance metrics to provide predictive insights into optimal device design. By leveraging these algorithms, key parameters influencing both efficiency and stability can be identified, enabling the design of high-performing, stable PSCs. Additionally, ML can uncover complex relationships between material properties and device behavior, potentially accelerating the development of more efficient and stable PSCs compared to traditional trial-and-error methods \cite{malek2024machine, mammeri2023paths, oboh2022artificial}.

In this study, we integrate SCAPS (Solar Cell Capacitance Simulator), a widely used 1-D simulator \cite{burgelman2000modelling}, with an ML-based regressor to optimize performances in HTL-free PSCs. SCAPS simulates the transport of charge carriers and the effects of varying key parameters, such as absorber thickness, dopant concentration, and material properties on device performance and stability. The resulting simulation data is used to train an ML model that predicts the optimal combination of parameters for high efficiency and minimal degradation. While some studies have used ML to optimize PSC performance \cite{malek2024machine, oboh2022artificial}, few have focused on optimizing both efficiency and stability in HTL-free devices.

To address this research gap,  we present a novel method for enhancing the PCE and stability of HTL-free PSCs using an ML-derived polynomial regressor or order 4 (PR-4). More specifically, our approach uniquely addresses performance and longevity simultaneously. SCAPS and numerical simulations are employed to fine-tune parameters by matching simulated data with experimental results. This data is then used to generate a dataset by varying key fabrication parameters, which is analyzed and used to train various machine learning models, including classical, ensemble, and neural network-based regressors. The derived polynomial regressor equation is applied as the objective function to optimize both efficiency and degradation of the cells. The data is also labeled into two classes, which are used to train different ML classifiers for classifying whether the optimized result ensures superior performance. We demonstrate that our approach not only enhances PCE but also minimizes degradation over time. The regressor optimizes performance metrics, while the classifier identifies configurations that yield superior performance, ensuring both optimization and assessment of the cells, offering a promising solution for the commercialization of HTL-free PSCs..

\section{Methodology}

In this section, we describe the SCAPS simulation model used for the PSC analysis, including the characteristics curves, performance metrics, and degradation analysis. Additionally, we outline the machine learning and optimization techniques employed to enhance the model's accuracy and efficiency.

\begin{figure}[t]
    \centering
    \includegraphics[width=\linewidth]{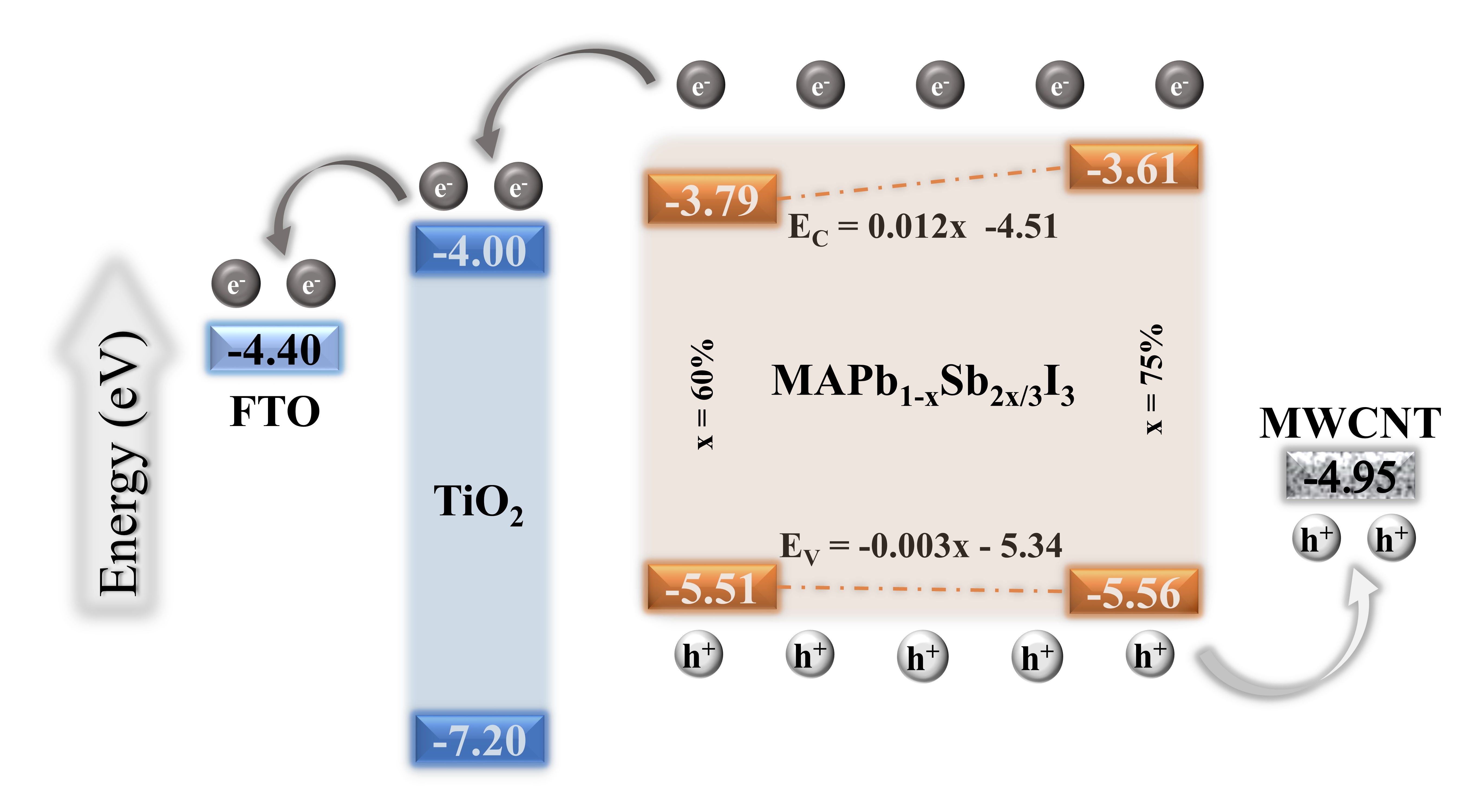}  
    \caption{Energy band structure of different layers.}
    \label{fig:eg}
\end{figure}

\subsection{Simulation Model in SCAPS}
The PSC simulation model is implemented using the SCAPS-1D simulator to solve the drift-diffusion, Poisson, and continuity equations in a one-dimensional domain using the finite difference method \cite{burgelman2000modelling, laidouci2023performance}. The governing equations are discretized and solved iteratively. The Poisson equation, relating the electric field, is given by Eq. (1), and the continuity equations for electrons and holes are given by Eq. (2) and (3), with the current densities for electrons and holes are shown in Eq. (4) and (5), respectively.

\begin{equation}
    \frac{E_{i+1} - E_i}{\Delta x} = - \frac{q}{\epsilon} \left( p_i - n_i + N_{D,i}^+ - N_{A,i}^- \right),
\end{equation}

\begin{equation}
    \frac{J_{n,i+1} - J_{n,i}}{\Delta x} + q (G_{n,i} - R_{n,i}) = 0,
\end{equation}
\begin{equation}
    \frac{J_{p,i+1} - J_{p,i}}{\Delta x} + q (G_{p,i} - R_{p,i}) = 0,
\end{equation}

\begin{equation}
    J_{n,i} = q n_i \mu_n E_i + q D_n \frac{n_{i+1} - n_i}{\Delta x},
\end{equation}
\begin{equation}
    J_{p,i} = q p_i \mu_p E_i - q D_p \frac{p_{i+1} - p_i}{\Delta x},
\end{equation}

where \( E \) is the electric field, \( p \) and \( n \) are hole and electron densities, and \( N_D^+ \) and \( N_A^- \) are ionized donor and acceptor concentrations. The index \(i\) represents the spatial grid point, and \( \Delta x \) is the spatial step size. \( J_n \) and \( J_p \) are the current densities for electrons and holes, respectively, \( G_n \) and \( G_p \) are the generation rates, and \( R_n \) and \( R_p \) are the recombination rates. \( \mu_n \) and \( \mu_p \) are the mobilities, and \( D_n \) and \( D_p \) are the diffusion coefficients. The total current density is the sum of \( J_n \) and \( J_p \). In SCAPS-1D simulation, recombination mechanisms such as Shockley-Read-Hall, Auger, and radiative recombination are incorporated to model the current-voltage characteristics of the PSC. The simulation is performed under steady-state conditions, solving these equations iteratively across the device structure.

\begin{table}[tb]
    \centering
    \caption{Simulation parameters for the SCAPS model.}
    \label{tab:1}
    \begin{tabular}{lcccc}
        \hline
        \textbf{Variable} & \textbf{Minimum} & \textbf{Maximum} & \textbf{Steps} & \textbf{Scale} \\
        \hline
        $x (\%)$ & 60 & 75 & 11 & Linear \\
        $N_{\text{ABS}}$ (cm$^{-3}$) & $1 \times 10^{14}$ & $9 \times 10^{14}$ & 5 & Log \\
        $T_{\text{ABS}}$ (nm) & 250 & 500 & 6 & Linear \\
        $N_{\text{ETL}}$ (cm$^{-3}$) & $6 \times 10^{16}$ & $18 \times 10^{16}$ & 5 & Log \\
        \hline
    \end{tabular}
\end{table}

The simulated photovoltaic device shown in Figure~\ref{FIG:1}, structured as FTO/TiO$_2$/MAPb$_{1-x}$Sb$_{2x/3}$I$_3$/MWCNTs, is analyzed under standard AM 1.5 G solar illumination at 300 K \cite{mohammed2023improving}. The initial series resistance ($R_S$) and shunt resistance ($R_{Sh}$) are set at 1 $\Omega \text{cm}^2$ \cite{moone2024performance} and 1 k$\Omega \text{cm}^2$ \cite{almora2024degradation}, respectively. Neutral defects, positioned at the mid-bandgap with an energy level of 0.1 eV, exhibit a Gaussian distribution, and the capture cross-section for both the carriers is \(2 \times 10^{-15} \ \text{cm}^2\). Charge carrier recombination is governed by Shockley-Read-Hall (SRH) and Auger mechanisms, with an SRH carrier lifetime of 250 ns \cite{mohammed2023improving} and an Auger recombination coefficient of \(1 \times 10^{-26} \ \text{cm}^6\text{s}^{-1}\) \cite{al2023understanding}. The surface recombination velocity is parameterized following \cite{malek2024machine}. Electrode work functions are set at 4.4 eV for the front contact (FTO) \cite{mohammed2023improving} and 4.95 eV for the back contact (MWCNTs) \cite{shiraishi2001work}, ensuring efficient carrier extraction.

The FTO/TiO$_2$/MAPb$_{1-x}$Sb$_{2x/3}$I$_3$/MWCNTs structure, with the band diagram shown in Figure~\ref{fig:eg}, ensures efficient charge extraction. FTO facilitates electron transfer from TiO$_2$, which, as an ETL, prevents hole backflow. The perovskite absorber, MAPb$_{1-x}$Sb$_{2x/3}$I$_3$, exhibits a tunable bandgap from 1.72 eV at $x = 60\%$ to 1.95 eV at $x = 75\%$ \cite{miah2024band}, where we consider a linear shift in the conduction and valence bands for this small range of $x$, as illustrated in the figure. MWCNTs enhance hole extraction due to their optimal work function, ensuring efficient charge collection \cite{malek2024machine}. This alignment minimizes recombination, thereby improving overall photovoltaic performance. 

The SCAPS simulation parameters, apart from those listed in Table \ref{tab:1}, are assumed to be fixed and sourced from \cite{miah2024band, mohammed2023improving, fooladvand2023single}. Table \ref{tab:1} presents the key variable parameters used in the simulation, selected based on the analysis in \cite{malek2024machine}, which highlights the impact of variations in the band structure (bandgap energy and electron affinity), absorber defect density ($N_{\text{ABS}}$), absorber thickness ($T_{\text{ABS}}$), and transport layer doping density as critical factors influencing PSC performance. The molar fraction, $x$, in MAPb$_{1-x}$Sb$_{2x/3}$I$_3$ is varied linearly from 60\% to 75\%, directly modulating the band structure. Since the structure lacks a HTL, only the electron transport layer doping concentration ($N_{\text{ETL}}$) is explored on a logarithmic scale to evaluate its influence on overall device efficiency. The thickness of the ETL is not considered in this study due to its negligible impact on device performance \cite{malek2024machine}.

Accurately modeling PSC degradation requires identifying its primary cause. Ion migration within the perovskite layer is a dominant factor, progressively introducing defects that impair performance \cite{zhao2021effects}. As defect density increases, power conversion efficiency (PCE) declines but eventually stabilizes, defining the device’s defect tolerance. This process can be quantitatively described by Eq. (6) \cite{nardone2014towards}, which models defect density evolution over time (\(t\)):  

\begin{equation}
N = N_0 \left(1 + \frac{t}{\tau}\right)^{1/2},
\end{equation}

where \( N_0 = 1 \times 10^{14} \ \text{cm}^{-3} \) is the initial defect density, and \( \tau \) represents the characteristic degradation time. So, for \( \tau = 0.9 \ \text{h} \), over 50 hours, \( N \approx 7.5N_0 \). The degradation in efficiency (\( \Delta \)) after 50 hours can be calculated using the initial efficiency (\( \eta_0 \)) and the efficiency at 50 hours (\( \eta_{50\mathrm{h}} \)), as given by \cite{hartono2023stability}.

\[
\Delta_{50h} = \frac{\eta_0 - \eta_{\text{50h}}}{\eta_0} \times 100\%.
\]

\subsection{Characteristic Curves and Performance Parameters}

\begin{figure}[tb]
    \centering
    \includegraphics[width=\linewidth]{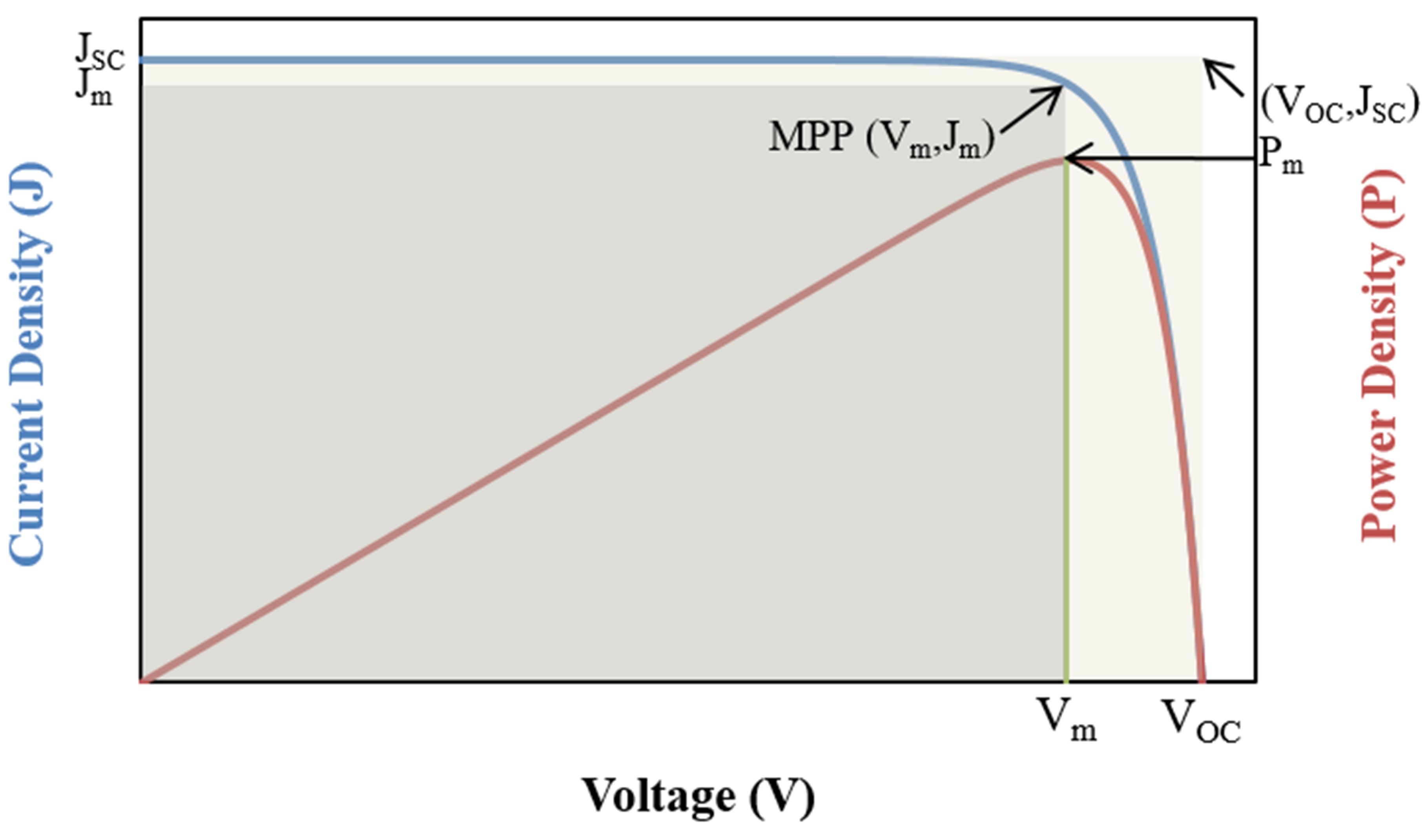}  
    \caption{J-V and P-V characteristic curves}
    \label{fig:jv_pv_curve}
\end{figure}

By varying key parameters in SCAPS, we obtain the characteristic current density–voltage (J–V) and power density–voltage (P–V) curves, along with performance metrics such as short-circuit current density ($J_{SC}$), open-circuit voltage ($V_{OC}$), fill factor ($FF$), power conversion efficiency ($\eta$), and stability. To reconstruct these characteristic curves using these parameters, an equivalent circuit model is useful. The characteristic curves are shown in Figure~\ref{fig:jv_pv_curve}, with the corresponding equation given by Eq. (7) \cite{nelson2003physics}:

\begin{equation}
    J = J_{ph} - J_0 \left( e^{\frac{V + J R_s}{nV_T}} - 1 \right) - \frac{V + J R_s}{R_{sh}}
\end{equation}

For the short-circuit condition (\( V = 0 \)), we obtain the expression for \( (J_{SC}=)~f_{J_{SC}} (J_{ph}, J_0, n)\) as a function of the unknown variables \( (J_{ph}, J_0, n) \), as shown in Eq. (8):

\begin{equation}
    f_{J_{sc}} = J_{ph} - J_0 \left( e^{\frac{J_{sc} R_s}{nV_T}} - 1 \right) - \frac{J_{sc} R_s}{R_{sh}}
\end{equation}

Similarly, for the open-circuit condition (\( J = 0 \)), Eq. (9) provides the expression for \( (V_{OC}=)~f_{V_{OC}} (J_{ph}, J_0, n) \):

\begin{equation}
    f_{V_{oc}} = J_{ph}R_{sh} - J_0R_{sh} \left( e^{\frac{V_{oc}}{nV_T}} - 1 \right)
\end{equation}

The power equation is given by:

\[
P = JV \quad \text{or} \quad D_p = \frac{dP}{dV} = J + V \cdot \frac{dJ}{dV}
\]

Next, for \( \frac{dJ}{dV} \), taking the derivative of both sides of Eq. (7) gives:

\[
\frac{dJ}{dV} = - J_0 e^{\frac{V + J R_s}{nV_T}} \cdot \frac{1 + R_s \frac{dJ}{dV}}{nV_T} - \frac{1 + R_s \frac{dJ}{dV}}{R_{sh}}
\]

\[
or, \frac{dJ}{dV} = -\frac{J_0 e^{\frac{V + J R_s}{nV_T}} \cdot \frac{1}{nV_T} + \frac{1}{R_{sh}}}{1 + \frac{J_0 R_s}{nV_T} e^{\frac{V + J R_s}{nV_T}} + \frac{R_s}{R_{sh}}}
\]

At the maximum power point (MPP), \( P_m = J_m V_m \); when \( D_P = 0 \), the corresponding current and voltage are \( J_m \) and \( V_m \). Finally, substituting \( \frac{dJ}{dV} \) for maximum power into the expression for \( ({D_P}=0=)~f_{D_P} (J_{ph}, J_0, n, V_m)\), we obtain Eq. (10):

\begin{equation}
    \begin{split}
        f_{D_P} = \left( J_{ph} - J_0 \left( e^{\frac{V_m + P_m R_s / V_m}{nV_T}} - 1 \right) - \frac{V_m + P_m R_s / V_m}{R_{sh}} \right) \\
        - V_m \frac{ J_0 e^{\frac{V_m + P_m R_s / V_m}{nV_T}} \cdot \frac{1}{nV_T} + \frac{1}{R_{sh}}}{1 + \frac{J_0 R_s}{nV_T} e^{\frac{V_m + P_m R_s / V_m}{nV_T}} + \frac{R_s}{R_{sh}}}
    \end{split}
\end{equation}

Finally, for maximum power, Eq. (11) gives the expression for \( (P_{m}=)~ f_{P_m} (J_{ph}, J_0, n, V_m) \):

\begin{equation}
    f_{P_m} = \left( J_{ph} - J_0 \left( e^{\frac{V_m + P_m R_s / V_m}{nV_T}} - 1 \right) - \frac{V_m + P_m R_s / V_m}{R_{sh}} \right) V_m
\end{equation}

Now, for the four unknown values \( (J_{ph}, J_0, n, V_m) \), we have four equations (8)-(11) that can be solved using the Newton-Raphson method as given in Eq. (12) \cite{akram2015newton}, with the Jacobian matrix J. The initial unknown values are set as \( J_{ph}, J_0, n, V_m \), with initial guesses \( J_{sc} \), \( 10^{-10} \), \( 1 \), and \( 90\% \) of \( V_{oc} \), respectively.

\begin{equation}
    {\begin{bmatrix}
J_{ph} \\
J_0 \\
n \\
V_m
\end{bmatrix}}_{k+1}
={\begin{bmatrix}
J_{ph} \\
J_0 \\
n \\
V_m
\end{bmatrix}}_k 
+
J^{-1}_k
{\begin{bmatrix}
J_{sc} - f_{J_{sc}} \\
V_{oc} - f_{V_{oc}} \\
D_P - f_{D_P} \\
P_m - f_{P_m}
\end{bmatrix}}_k
\end{equation}

\(
    where, J = \begin{bmatrix}
\frac{\partial f_{J_{sc}}}{\partial f_{J_{ph}}} & \frac{\partial f_{J_{sc}}}{\partial f_{J_0}} & \frac{\partial f_{J_{sc}}}{\partial f_n} & \frac{\partial f_{J_{sc}}}{\partial f_{V_m}} \\
\frac{\partial f_{V_{oc}}}{\partial f_{J_{ph}}} & \frac{\partial f_{V_{oc}}}{\partial f_{J_0}} & \frac{\partial f_{V_{oc}}}{\partial f_n} & \frac{\partial f_{V_{oc}}}{\partial f_{V_m}} \\
\frac{\partial f_{D_P}}{\partial f_{J_{ph}}} & \frac{\partial f_{D_P}}{\partial f_{J_0}} & \frac{\partial f_{D_P}}{\partial f_n} & \frac{\partial f_{D_P}}{\partial f_{V_m}} \\
\frac{\partial f_{P_m}}{\partial f_{J_{ph}}} & \frac{\partial f_{P_m}}{\partial f_{J_0}} & \frac{\partial f_{P_m}}{\partial f_n} & \frac{\partial f_{P_m}}{\partial f_{V_m}}
\end{bmatrix}
\)

\subsection{Polynomial Regression in ML}

\begin{figure}[tb]
    \centering
    \includegraphics[width=\linewidth]{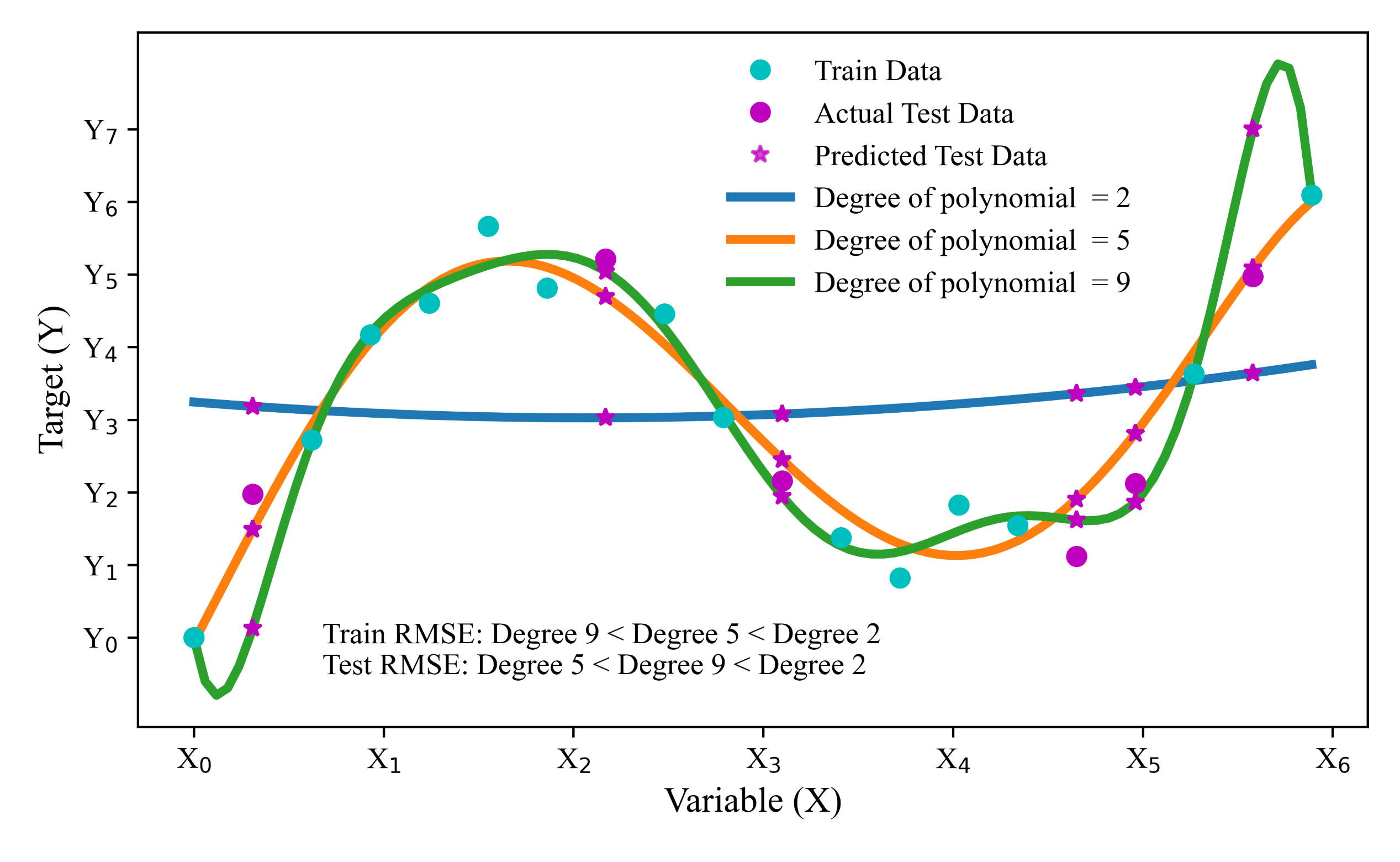}
    \caption{Polynomial regression fits with varying degrees.}
    \label{fig:poly_regression}
\end{figure}

Polynomial regression is a supervised machine learning technique where the relationship between the independent variable \(X\) and the dependent variable \(Y\) is modeled as an \(n\)-degree polynomial, as shown in Eq. (13) \cite{peckov2012machine}.

\begin{equation}
Y = \sum_{i=0}^{n\ge2} W_i X^i    
\end{equation}

Here, \(W_i\) are the coefficients, and \(n\) is the polynomial degree. Polynomial regression is a simpler alternative to neural networks for modeling non-linear relationships \cite{cheng2018polynomial}. Unlike linear regression (LR), which assumes a linear relationship between X and Y, polynomial regression can capture more complex relationships.

In machine learning, polynomial regression fits data that linear models cannot, using training data to determine the coefficients. The focus is on prediction accuracy, optimized through techniques like cross-validation, regularization, and hyperparameter tuning, rather than adhering to statistical assumptions. In contrast, statistical regression relies on assumptions such as normally distributed errors and constant error variance \cite{draper1998applied}. Machine learning methods do not explicitly enforce these assumptions but still aim to maximize prediction accuracy, even when the data does not meet statistical criteria.

Figure \ref{fig:poly_regression} illustrates polynomial regression of varying degrees on the same set of points, as an example. As the degree increases, model complexity grows, improving the fit to training data but also risking overfitting, where the model captures noise rather than the true relationship. As shown, for degrees 2, 5, and 9, the training root mean square error (RMSE) increases with the degree of the polynomial. However, for the test case, the maximum error occurs at the minimum degree, while the minimum error does not occur at the maximum degree. This suggests overfitting of the training data, introducing significant error for the first and last test samples. Therefore, the 5th degree polynomial works best, yielding the minimum RMSE for the test data. This highlights the effectiveness of the model with lower complexity in achieving higher prediction accuracy.

In this work, two separate linear regression models were trained: one for predicting PCE (\(\eta (\%)\)) and another for predicting degradation (\(\Delta (\%)\)). Finally, the polynomial equations for \(\eta (\%)\) and \(\Delta (\%)\) were derived and presented, where the coefficients correspond to the impact of each polynomial feature on the respective output. As shown in Table \ref{tab:poly_regression_ml_params}, the polynomial regression model utilizes several key parameters, including the data features, target variables, model settings, and values for test size, polynomial degree, and scaling.

\begin{table}[tb]
\centering
\caption{Machine Learning Parameters for Polynomial Regression Model}
\begin{tabular}{ll}
\hline
\textbf{Parameter}           & \textbf{Description}                                                \\ \hline
Independent features                   & 4                          \\ 
Dependent targets                   & 2                    \\ 
Train to test ratio           & 80 : 20                            \\ 
Degree of Polynomial & 4                                     \\
Scaler for feature normalization & StandardScaler                              \\ 

Performance evaluation metrics & R² and RMSE                        \\ \hline
\end{tabular}

\label{tab:poly_regression_ml_params}
\end{table}

\subsection{L-BFGS-B Optimization Method}

\begin{figure}[tb]
    \centering
    \includegraphics[width=\linewidth]{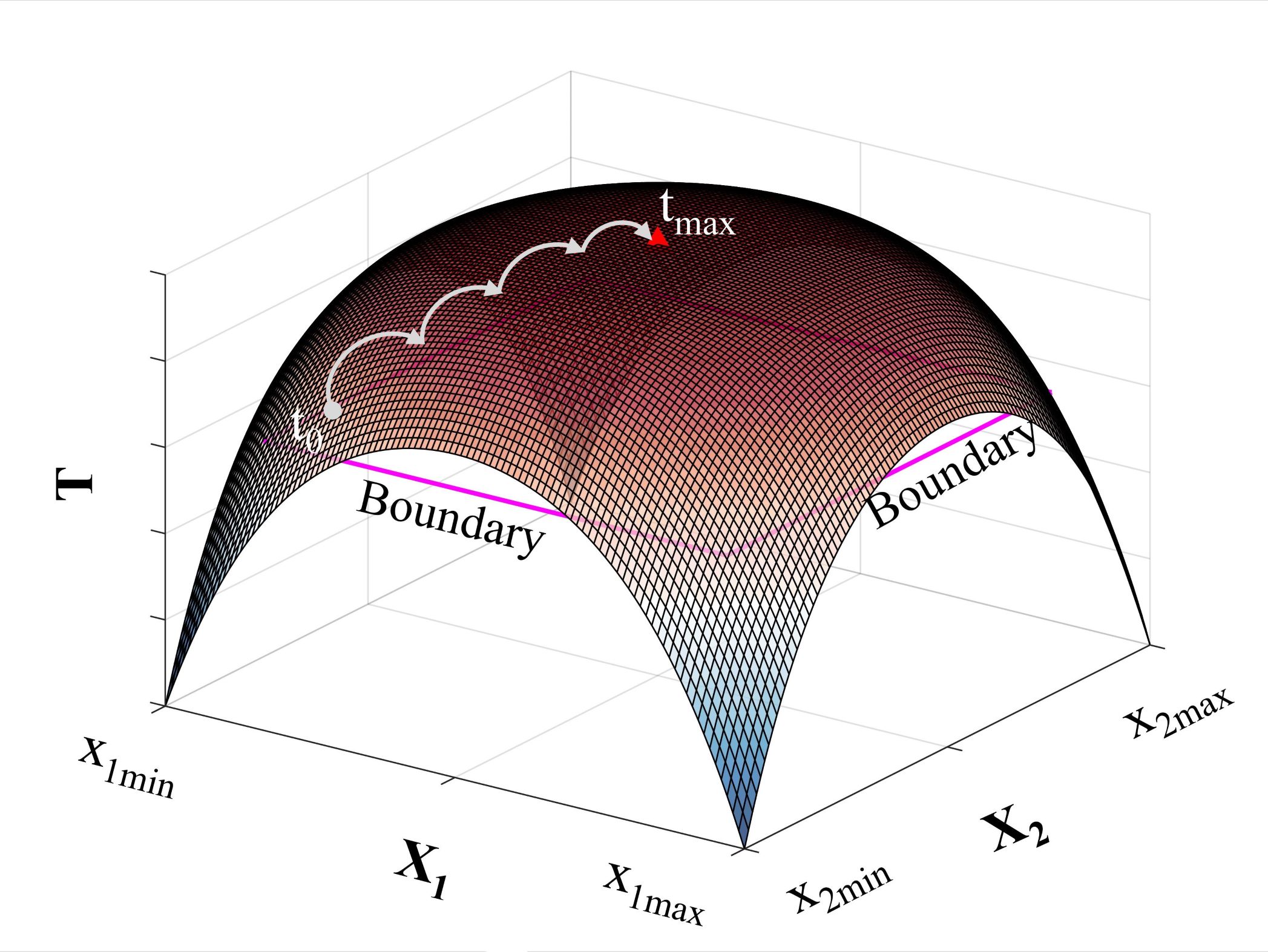}
    \caption{The process of L-BFGS-B optimization}
    \label{fig:opt}
\end{figure}

Limited-memory Broyden–Fletcher–Goldfarb–Shanno with Box constraints (L-BFGS-B) is a quasi-Newton method for large-scale optimization with bound constraints. It approximates the Hessian matrix, which represents second-order derivatives of the objective function, improving memory efficiency while ensuring optimization within variable bounds \cite{byrd1995limited}. The update step is shown in Eq. (14):

\begin{equation}
    \mathbf{x}_{k+1} = \mathbf{x}_k - \alpha_k \mathbf{H}_k^{-1} \nabla f(\mathbf{x}_k)
\end{equation}

where \( \mathbf{H}_k^{-1} \) is the inverse Hessian approximation, and \( \nabla f(\mathbf{x}_k) \) is the gradient. For the objective function \( T(X_1, X_2) \), as depicted in Figure~\ref{fig:opt}, the Hessian matrix is:

\[
\mathbf{H}(X_1, X_2) =
\begin{bmatrix}
\frac{\partial^2 T}{\partial X_1^2} & \frac{\partial^2 T}{\partial X_1 \partial X_2} \\
\frac{\partial^2 T}{\partial X_2 \partial X_1} & \frac{\partial^2 T}{\partial X_2^2}
\end{bmatrix}
\]

This matrix describes the curvature of \( T(X_1, X_2) \). L-BFGS-B uses an approximation of the Hessian to reduce memory usage, updated iteratively using gradient and position differences.

The optimization enforces variable bounds with a projected gradient approach: \(l_i \leq x_i \leq u_i\); where \( l_i \) and \( u_i \) are the bounds. In this work, L-BFGS-B optimizes a polynomial regression model by minimizing a weighted sum of \( \Delta \) and \( \eta \) as the objective function (Eq. 15), ensuring physically meaningful parameters, where the sign of the weighted sum is chosen to maximize \( \eta \) and minimize \( \Delta \). For simplicity, equal importance is assigned to both targets by setting the weights of $\eta$ and $\Delta$ as \( w_{\eta} = w_{\Delta} = 0.5 \). However, these weights can be adjusted depending on the specific application and the desired emphasis on each target. For instance, in cost-sensitive rural electrification projects, longer operational stability may be prioritized to minimize maintenance, leading to a higher weight on $\Delta$ \cite{diarra2002solar}. Conversely, in space-constrained urban installations where maximizing energy yield per unit area is essential, higher initial efficiency may be given greater importance by assigning a higher weight on $\eta$ \cite{etukudoh2024solar}. Similarly, grid-integrated systems may require a balanced optimization of both metrics to ensure reliable and efficient power output.

\begin{equation}
    f (x, N_{ABS}, T_{ABS}, N_{ETL}) = w_{\eta}\eta - w_{\Delta}\Delta
\end{equation}

\section{Results and Discussions}

This section begins by validating the simulation results through parameter tuning in SCAPS, using experimental data for comparison. Once the parameters are optimized, SCAPS is employed to generate the dataset. The dataset is then labeled and analyzed to extract valuable insights, followed by the evaluation of ML model performances. Finally, we present the optimized outcomes, achieving a high PCE with minimal degradation, ensuring superior performance.

\begin{table}[b]
\centering
\caption{Layer Properties Used in Simulation \cite{malek2024machine, miah2024band, mohammed2023improving, oboh2022artificial, fooladvand2023single}}
\label{table:layer_properties}
\begin{tabular}{lll}
\hline
\textbf{Layer Properties} & \textbf{ETL} & \textbf{Absorber} \\  
\hline
Thickness (nm) & 50 \cite{malek2024machine} & 290 \cite{malek2024machine} \\  
Energy Gap (eV) & 3.2 \cite{oboh2022artificial} & 1.85 \cite{miah2024band} \\  
Electron Affinity (eV) & 4.0 \cite{mohammed2023improving} & 3.67 \cite{miah2024band} \\  
Dielectric Permittivity & 9 \cite{mohammed2023improving} & 6.5 \cite{mohammed2023improving} \\  
DOS$_{CB}$ (cm\textsuperscript{-3}) & 2.0 × 10\textsuperscript{17} \cite{mohammed2023improving} & 1.66 × 10\textsuperscript{19} \cite{fooladvand2023single} \\  
DOS$_{VB}$ (cm\textsuperscript{-3}) & 6.0 × 10\textsuperscript{17} \cite{mohammed2023improving} & 5.41 × 10\textsuperscript{19} \cite{fooladvand2023single} \\  
$\mu_n$ (cm\textsuperscript{2}/Vs) & 100 \cite{mohammed2023improving} & 50 \cite{fooladvand2023single} \\ 
$\mu_p$ (cm\textsuperscript{2}/Vs) & 25 \cite{mohammed2023improving} & 50 \cite{fooladvand2023single} \\  
$N_{Acceptor}$ (cm\textsuperscript{-3}) \cite{mohammed2023improving} & 0 & 1 × 10\textsuperscript{18} \\  
$N_{Donor}$ (cm\textsuperscript{-3}) \cite{mohammed2023improving} & 1.65 × 10\textsuperscript{17} & 1 × 10\textsuperscript{18} \\  
$N_{Defect}$ (cm\textsuperscript{-3}) \cite{malek2024machine} & 1 × 10\textsuperscript{15} & 1 × 10\textsuperscript{14} \\  
\hline
\end{tabular}
\end{table}

\begin{figure}[tb]
    \centering
    \begin{subfigure}{\linewidth}
        \includegraphics[width=\textwidth]{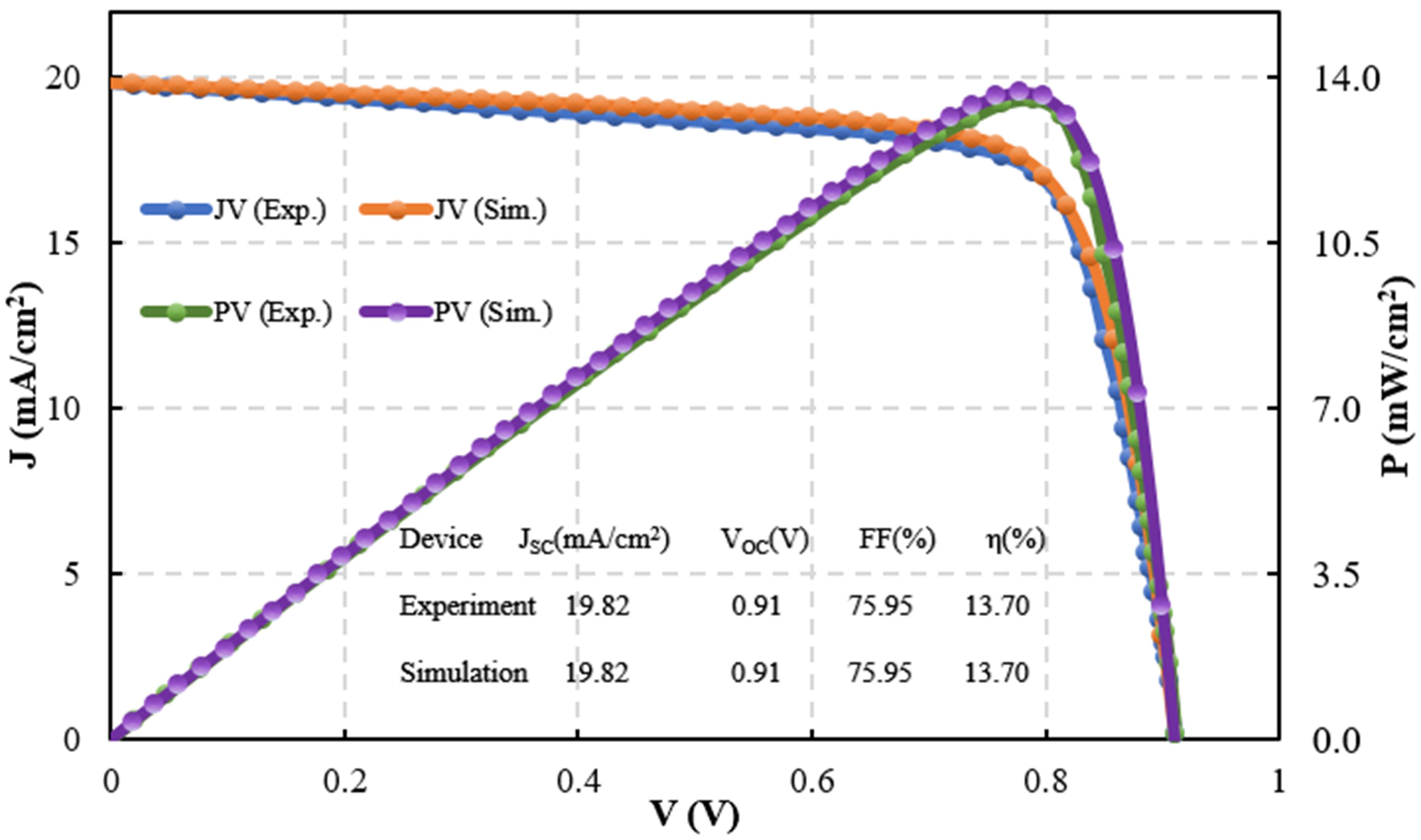}
        \caption{J-V and P-V curves for an experimental CNT-PSC \cite{mohammed2023improving} and the simulation in this work.}
    \label{fig:jv_pv}
    \end{subfigure}
    \begin{subfigure}{\linewidth}
        \includegraphics[width=\textwidth]{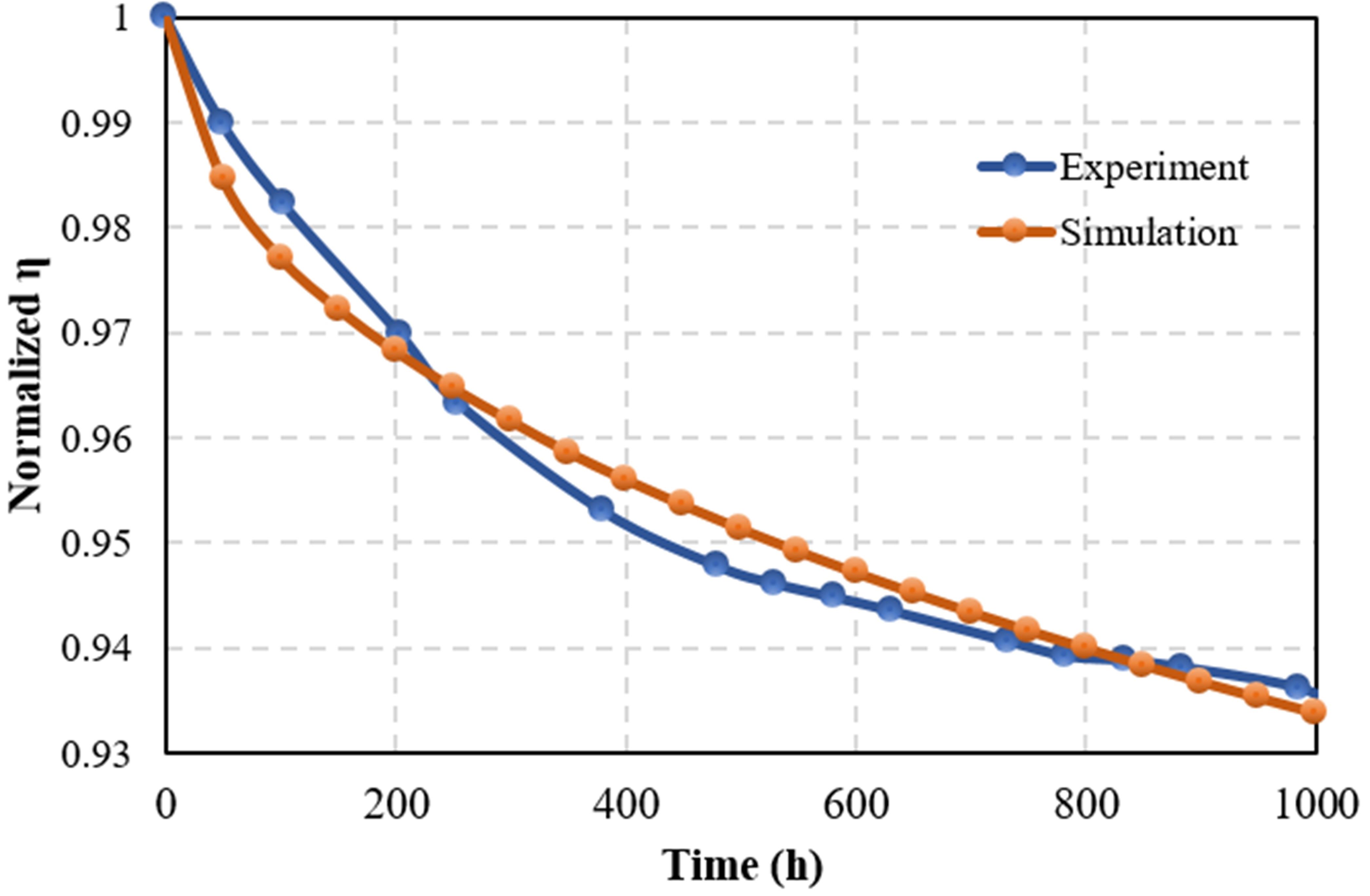}
        \caption{Stability curve showing PCE degradation over time for an experimental CNT-PSC \cite{luo2018all} and the simulation in this work.}
        \label{fig:stability_curve}
    \end{subfigure}
    \caption{Validation curves.}
    \label{fig:expvl}
\end{figure}

\subsection{Experimental Validation of Simulation}

We validate the accuracy of the simulation model by comparing the simulated and experimental results for current density, power density, and performance degradation over time. Figure~\ref{fig:jv_pv} illustrates the variation of current density with voltage (J-V) on the primary y-axis and power density with voltage (P-V) on the secondary y-axis for a carbon nanotube-based perovskite solar cell (CNT-PSC)~\cite{mohammed2023improving}, compared with our simulated model. Both the J-V and P-V characteristics exhibit a near-perfect match with the experimental results when using the parameter values listed in Table~\ref{table:layer_properties}. 
Furthermore, the key photovoltaic performance parameters—J\textsubscript{SC}, V\textsubscript{OC}, FF, and \(\eta\)—fully match the experimental results, yielding a root mean square error (RMSE) of zero in the simulated curves. Any minor discrepancies observed are attributed to the value extraction process of experimental data.

To assess long-term stability, the degradation of PCE over time was analyzed by comparing the simulated stability curve with another CNT-PSC \cite{luo2018all}. Figure~\ref{fig:stability_curve} presents the stability comparison, demonstrating a strong agreement with an RMSE of 0.003. The degradation parameter \(\Delta\) was found to be 6.5\% experimentally and 6.6\% in our simulation after 1000 hours, corresponding to an absolute error of 1.54\%. This result was obtained for \(\tau = 0.9\) hour (54 minutes) and an initial defect density \(N_0 = 5 \times 10^{13}\) cm\textsuperscript{-3}, confirming the accuracy of the degradation modeling approach. Consequently, the time required to degrade to \( p\% \) of the initial PCE, denoted as \( T_p \) \cite{khenkin2020consensus}, is approximately 450 hours experimentally and 500 hours in the simulation for \( T_{95} \). The time difference is negligible for \( T_{94} \) and \( T_{97} \), further confirming the consistency between experimental and simulated results. These results confirm that the simulated model accurately reproduces both the electrical characteristics and long-term stability of CNT-PSC devices, demonstrating its reliability for further dataset generation.

\subsection{Dataset}

\begin{figure*}[b]
    \centering
    \begin{subfigure}{0.4\textwidth}
        \caption{}
        \includegraphics[width=\textwidth]{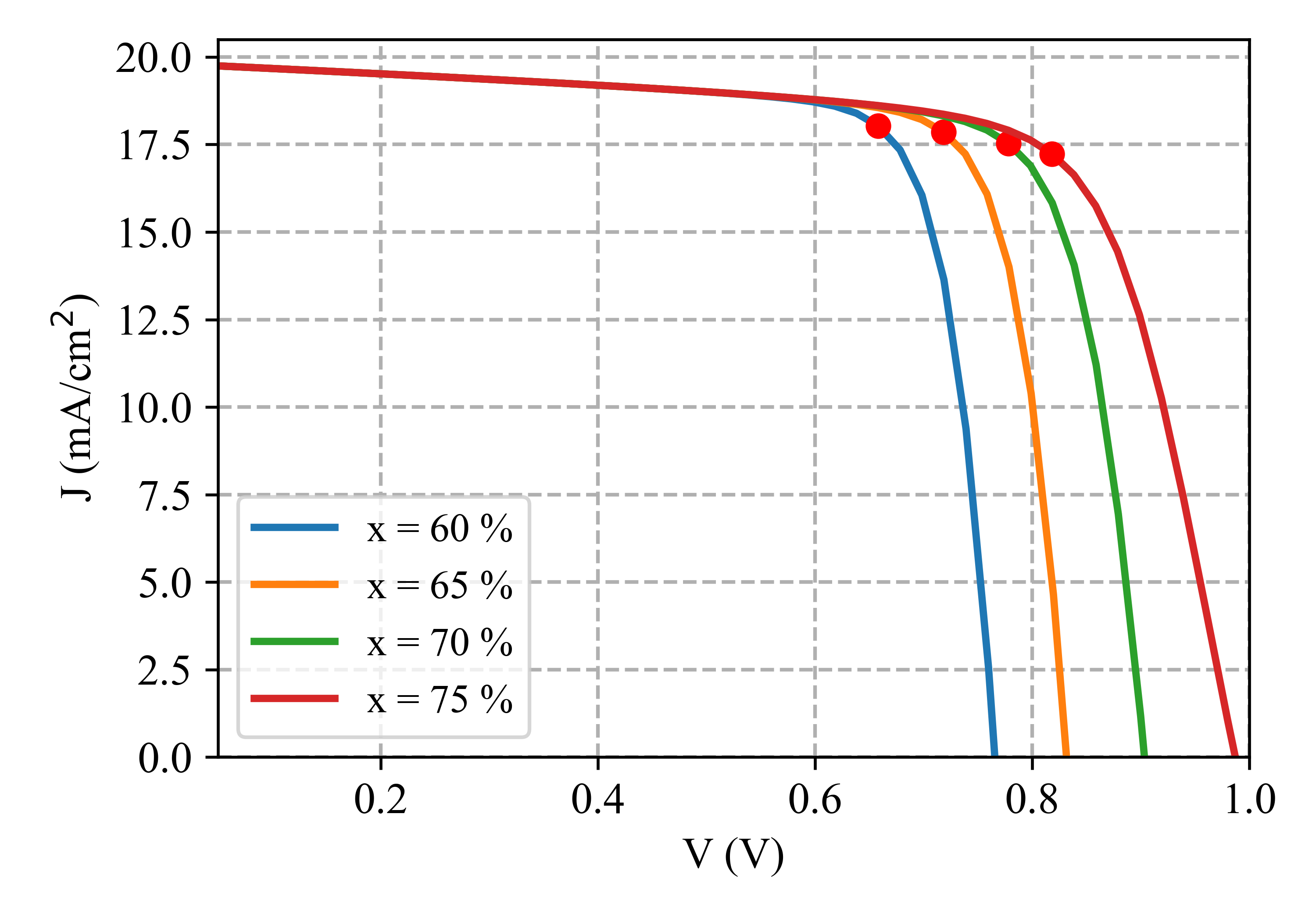}
        \label{fig:7a}
    \end{subfigure}
    \begin{subfigure}{0.4\textwidth}
        \caption{}
        \includegraphics[width=\textwidth]{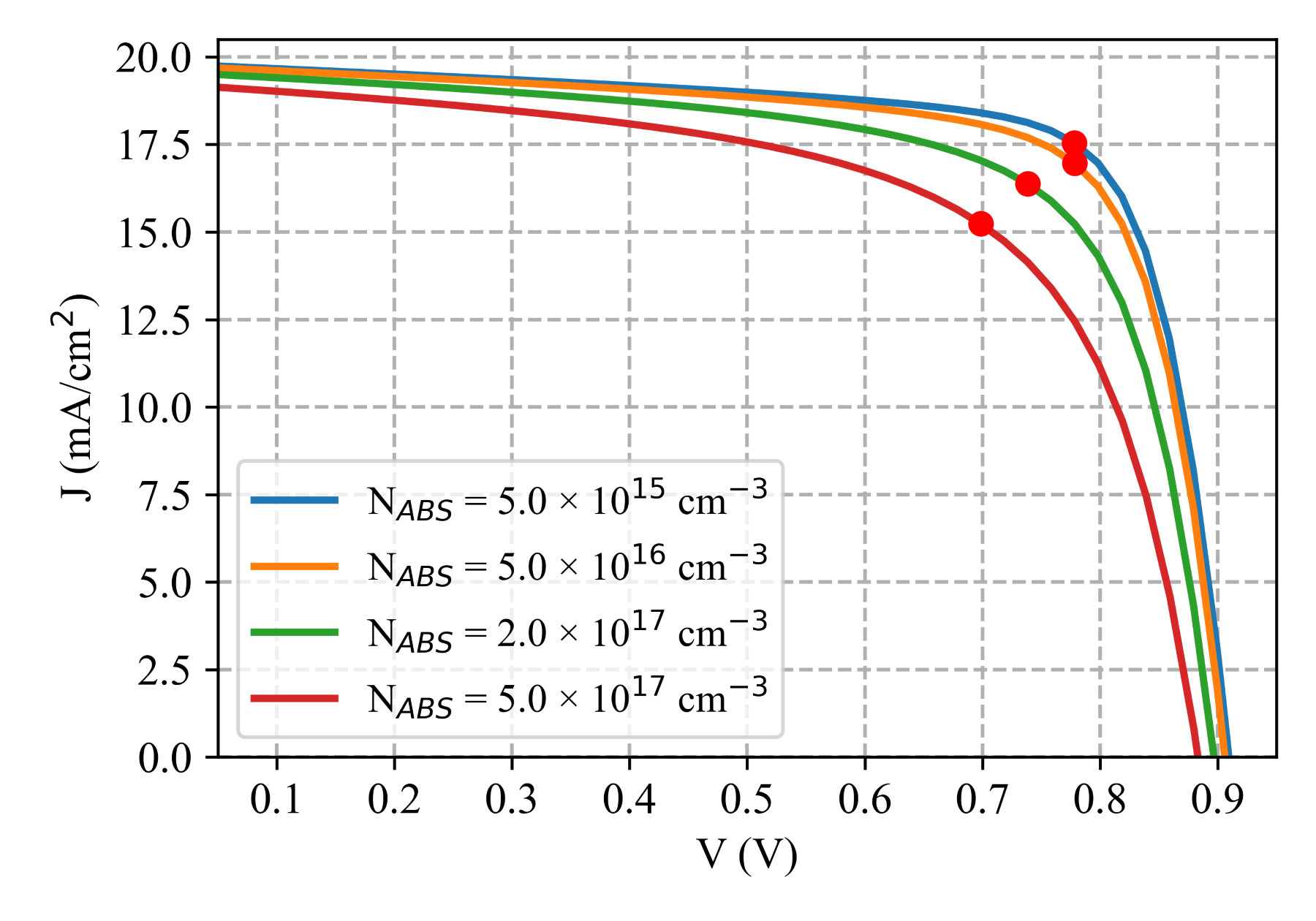}
        \label{fig:7b}
    \end{subfigure}
    \begin{subfigure}{0.4\textwidth}
        \caption{}
        \includegraphics[width=\textwidth]{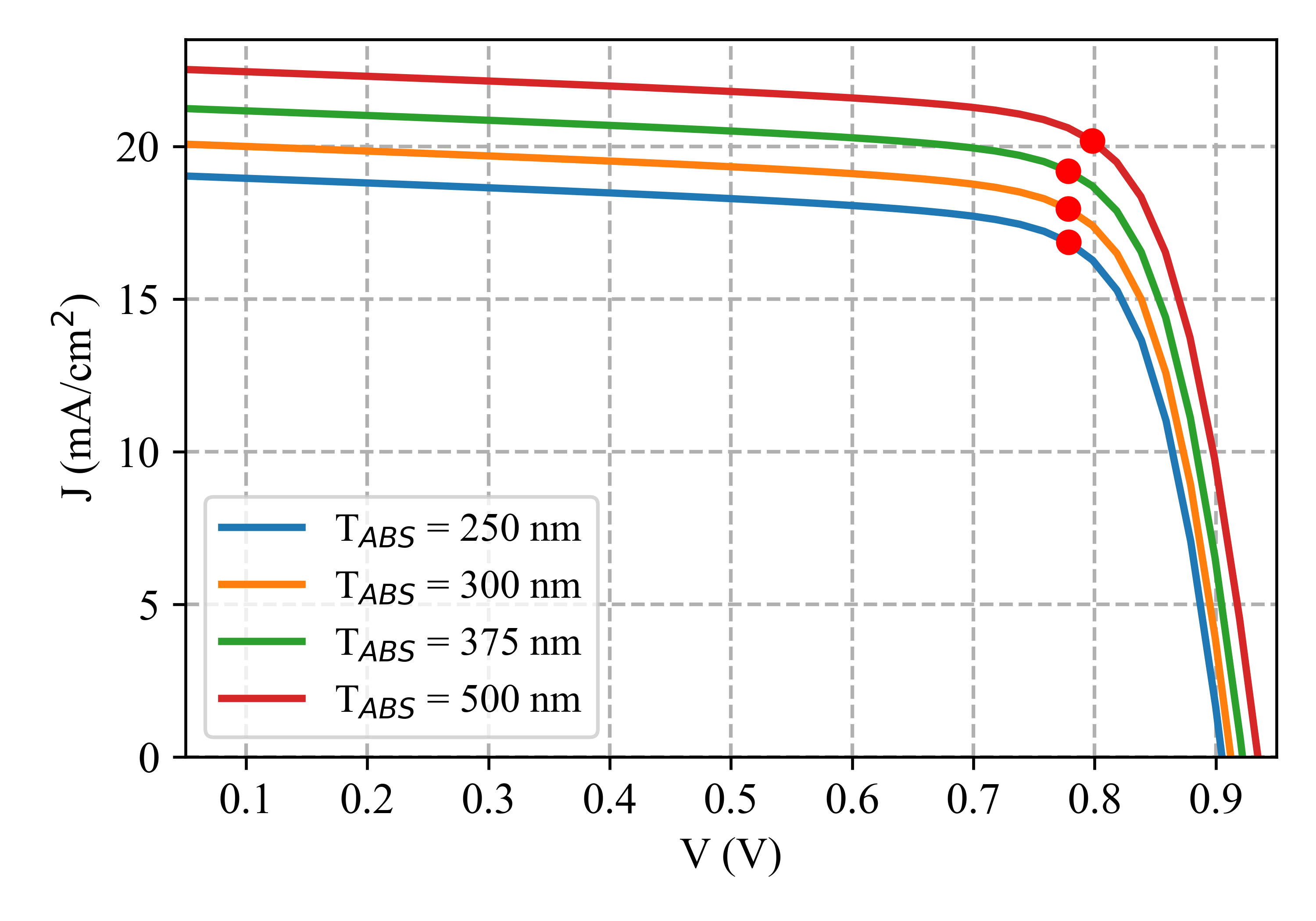}
        \label{fig:7c}
    \end{subfigure}
    \begin{subfigure}{0.4\textwidth}
        \caption{}
        \includegraphics[width=\textwidth]{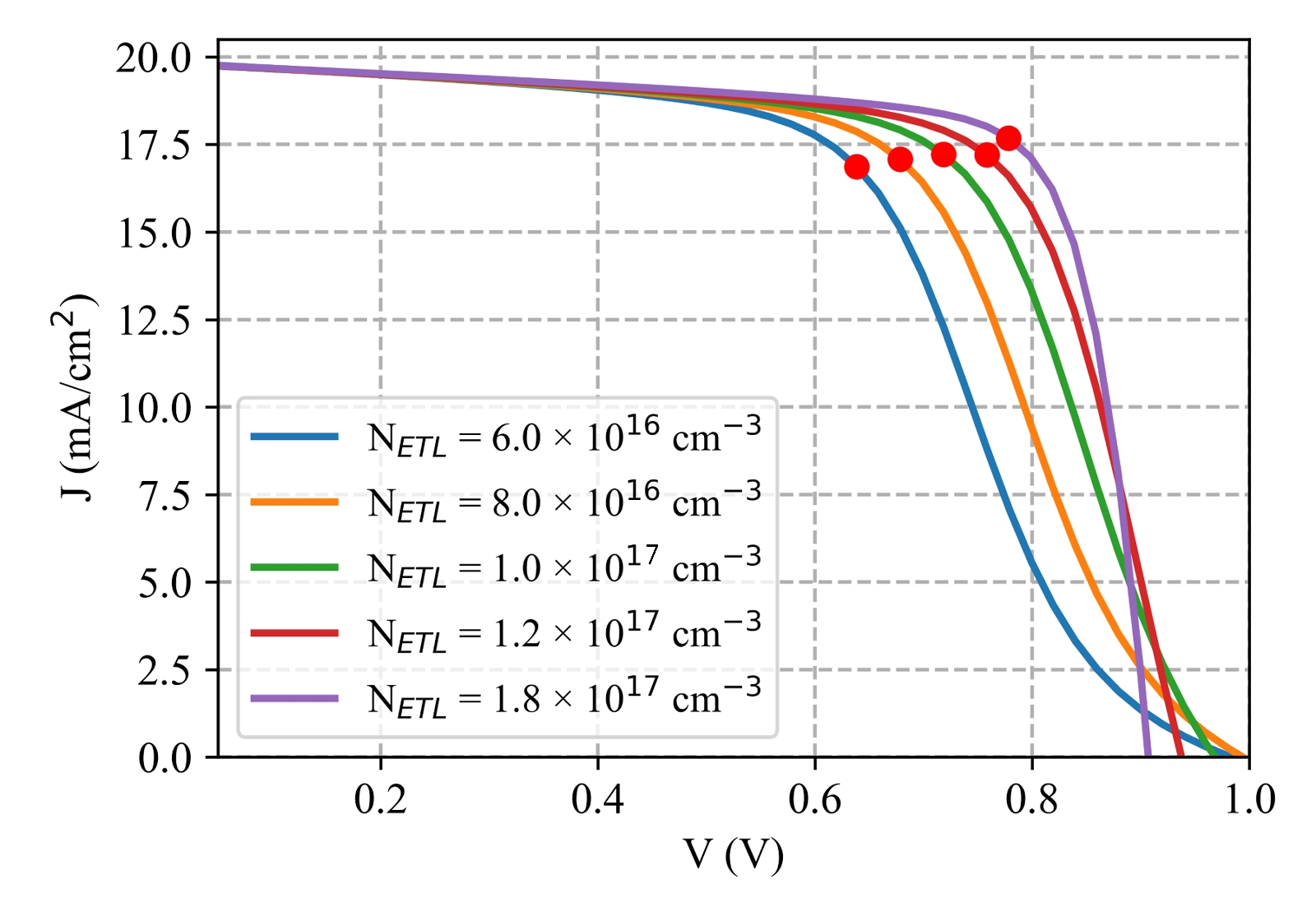}
        \label{fig:7d}
    \end{subfigure}
    \caption{J–V characteristic curves with variations in (a) molar fraction, (b) absorber defect density, (c) absorber layer thickness, and (d) ETL dopant concentration; the MPP is indicated by red markers.}
    \label{fig:jvft}
\end{figure*}

Varying the 4 key fabrication variables as features listed in Table \ref{tab:1}, within their minimum to maximum values and following the mentioned steps, results in a total of \(11 \times 6 \times 5 \times 5 = 1650\) samples in the dataset. These samples have 5 performance parameters as targets: \( J_{SC} \), \( V_{OC} \), FF, \( \eta \), and \( \Delta \). In the dataset, \( \Delta \) represents degradation after about 50 hours of aging. To observe how these features affect the performance, the J-V curve and degradation curve were examined by varying these features, as shown in Figure~\ref{fig:jvft} and Figure~\ref{fig:stbft}, respectively. 

\subsubsection{Variation in Characteristic Curves}

In Figure~\ref{fig:jvft}a, it is evident that increasing the first feature, molar fraction \(x\), raises \(V_{OC}\), as lowering \(x\) reduces the bandgap of the absorption layer, as shown in Figure~\ref{fig:eg}. This decrease in bandgap results in a reduction in \(V_{OC}\), as reported in~\cite{yan2022simultaneously}. The MPP also shifts with increasing \(x\), consistent with the rise in \(V_{OC}\). Moving to Figure~\ref{fig:jvft}b, we observe that a higher absorber defect density decreases efficiency due to increased recombination rates, leading to energy losses, as discussed in~\cite{motti2019defect}. Correspondingly, the MPP drops with increasing defect density, highlighting the degradation in power output. In Figure~\ref{fig:jvft}c, an increase in the absorber layer thickness results in a rise in \(J_{SC}\), attributed to enhanced light absorption and greater charge carrier generation, as observed in~\cite{mykytyuk2012limitations}. The MPP moves upward with thicker absorber layers, primarily driven by the improvement in \(J_{SC}\). Finally, in Figure~\ref{fig:jvft}d, increasing the ETL doping concentration improves the fill factor, which aligns with the findings in~\cite{jeyakumar2020influence}, due to enhanced charge carrier transport. A slight reduction in \(V_{OC}\) is observed, likely due to band alignment issues~\cite{malek2024machine} at the ETL interface. Nonetheless, the MPP improves, benefiting from the significant enhancement in the fill factor.

\subsubsection{Variation in Degradation Curves}

\begin{figure*}[b]
    \centering
    \begin{subfigure}{0.4\textwidth}
        \caption{}
        \includegraphics[width=\textwidth]{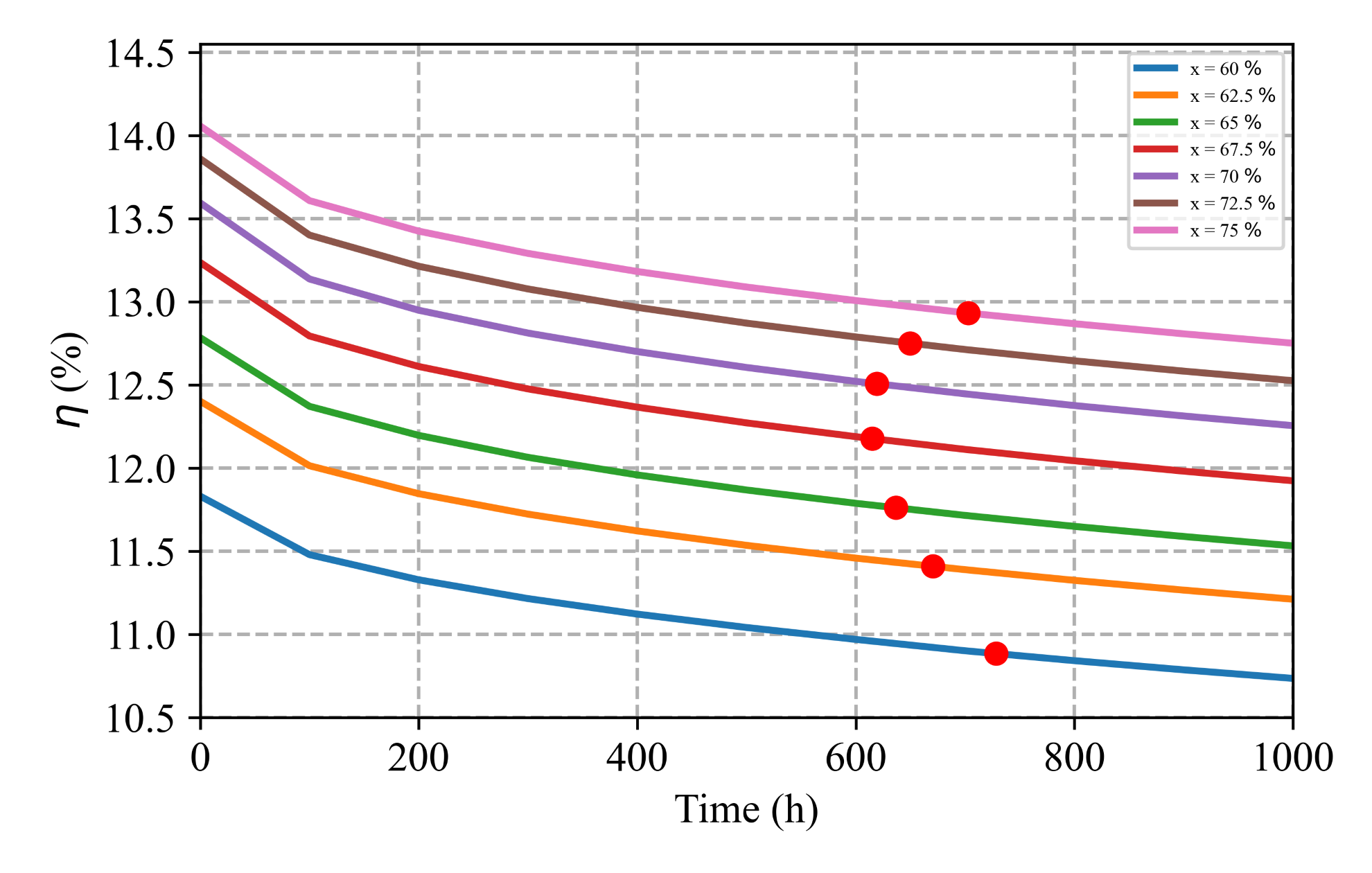}
        \label{fig:8a}
    \end{subfigure}
    \begin{subfigure}{0.4\textwidth}
        \caption{}
        \includegraphics[width=\textwidth]{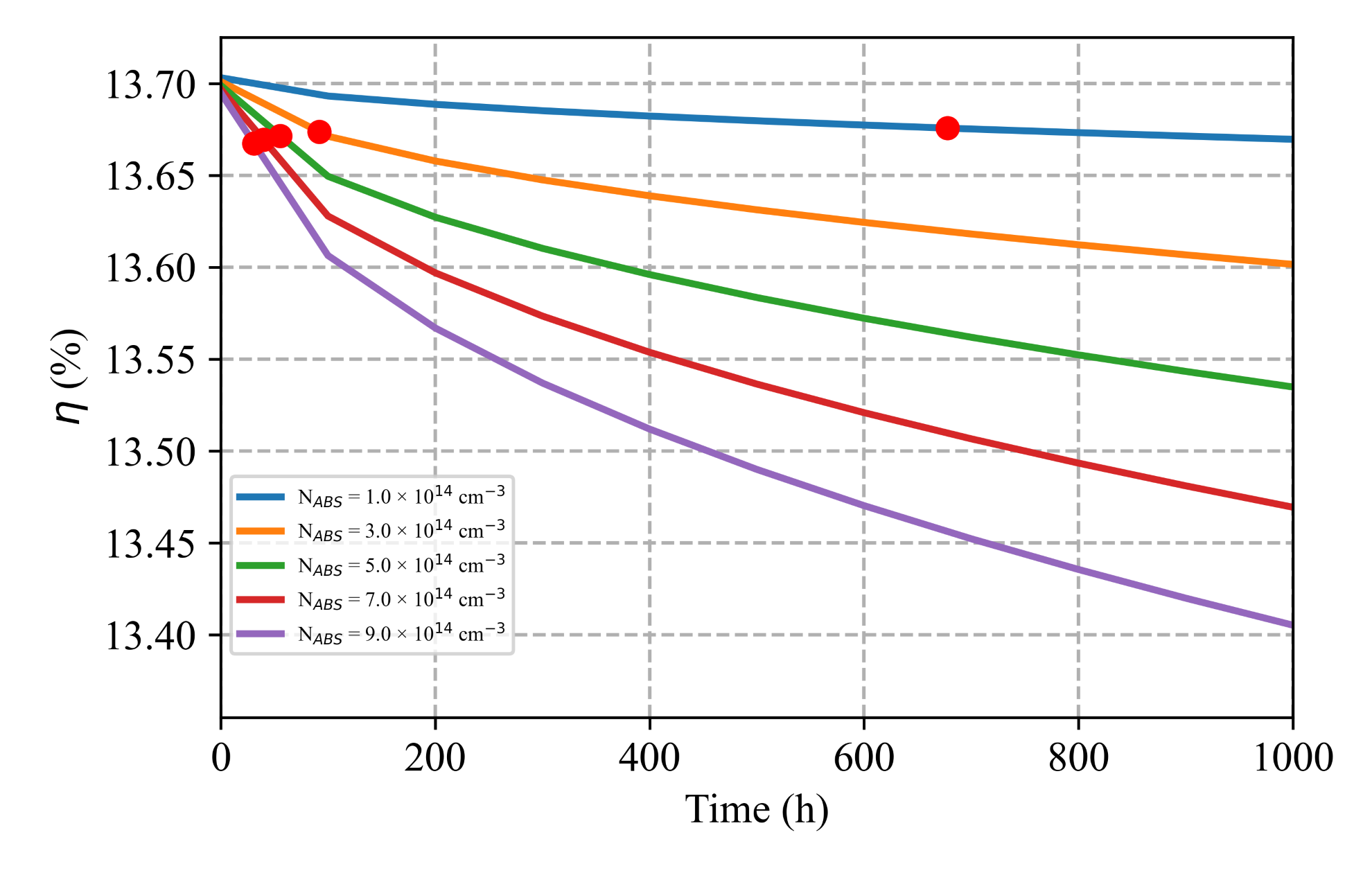}
        \label{fig:8b}
    \end{subfigure}
    \begin{subfigure}{0.4\textwidth}
        \caption{}
        \includegraphics[width=\textwidth]{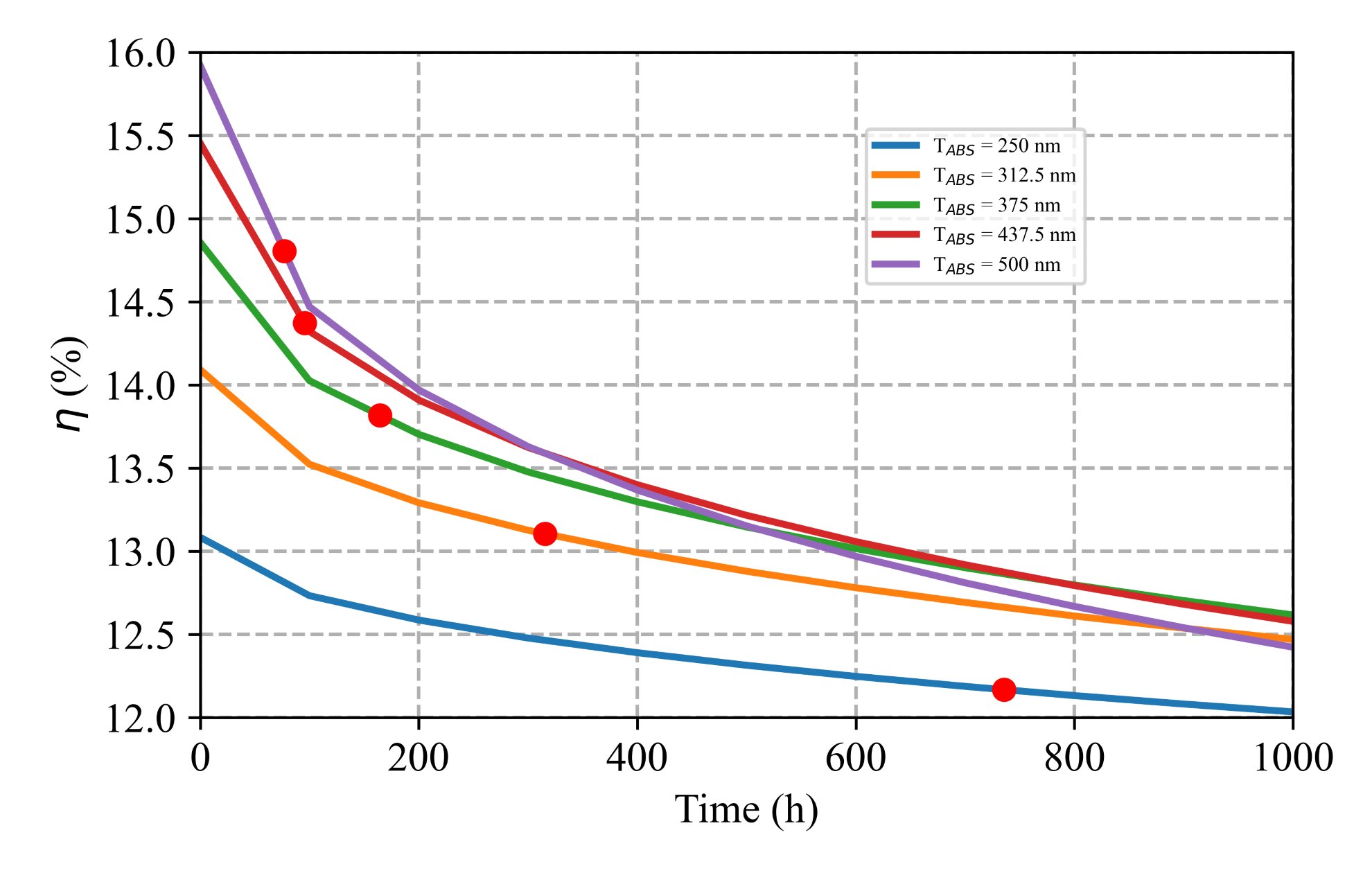}
        \label{fig:8c}
    \end{subfigure}
    \begin{subfigure}{0.4\textwidth}
        \caption{}
        \includegraphics[width=\textwidth]{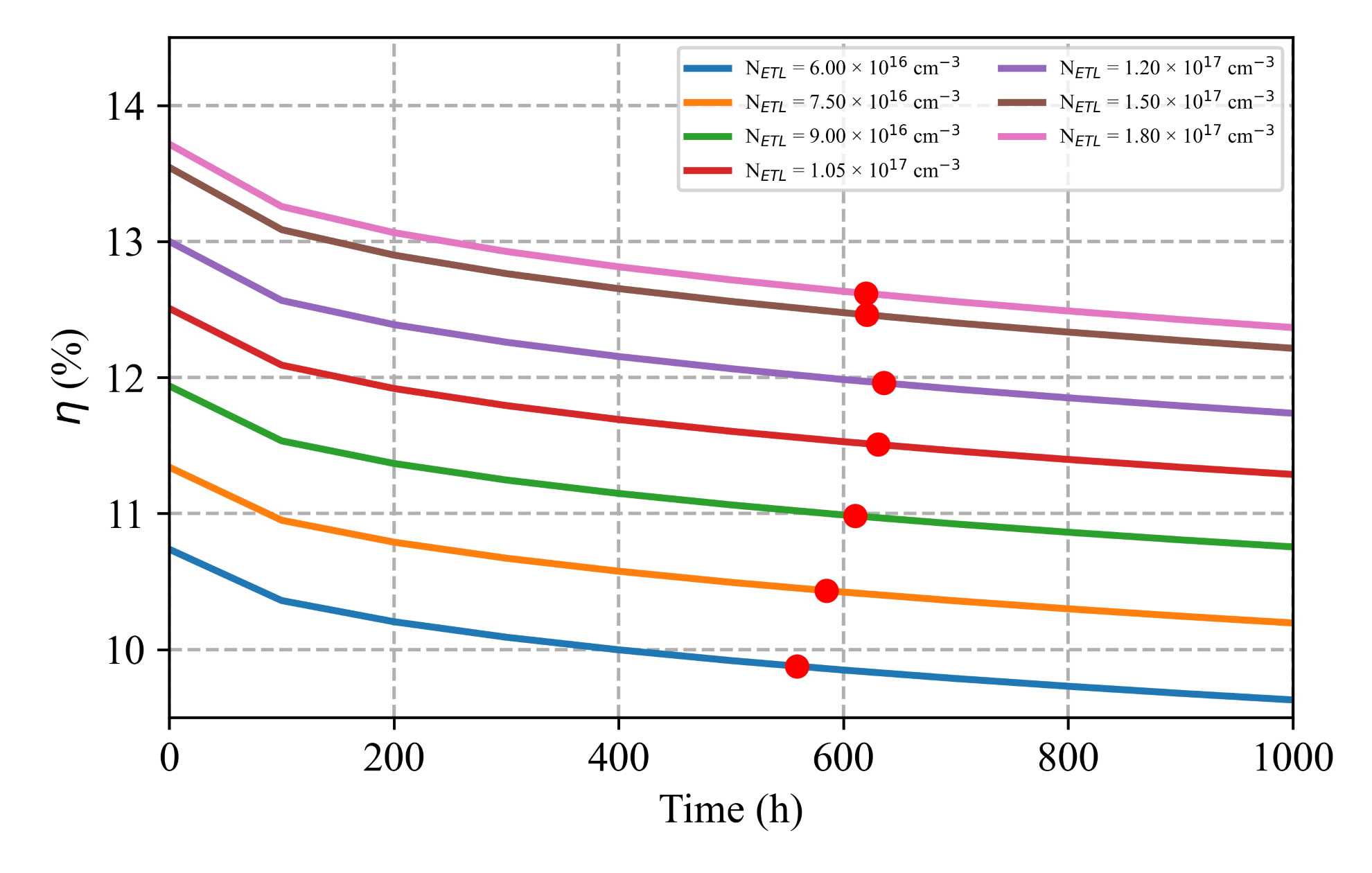}
        \label{fig:8d}
    \end{subfigure}
    \caption{Efficiency degradation curves with variations in (a) molar fraction, (b) absorber defect density, (c) absorber layer thickness, and (d) ETL dopant concentration; the $T_{92}$, $T_{99.8}$, $T_{93}$, and $T_{92}$ points indicated by red markers, respectively.}
    \label{fig:stbft}
\end{figure*}

In Figure~\ref{fig:stbft}a, we observe that increasing \(x\) increases the \(\eta\), but results in a slight change in \( T_{92} \). The minor change in \( T_{92} \) can be attributed to the relatively stable degradation process despite the increase in efficiency. With increasing \(x\), \(T_{92}\) initially decreases and then starts increasing, following a parabolic trend. However, this variation occurs within a narrow span of approximately 600–750 hours. In Figure~\ref{fig:stbft}b, we observe that increasing the defect density leads to a significant decrease in \( T_{99.8} \) from about 650 hours to nearly 10 hours, varying over a wide range, as the higher defect density promotes faster recombination, accelerating degradation. A slight change in \(\eta\) with defect density is observed due to the low variation ranges in defects compared to Figure~\ref{fig:jvft}b. In Figure~\ref{fig:stbft}c, we observe that an increase in absorber layer thickness improves efficiency but reduces \( T_{93} \) from about 750 hours to 50 hours, varying over a broad range. This is because, while thicker layers enhance light absorption and carrier generation, they also increase the recombination rates, which, over time, leads to quicker degradation. Finally, in Figure~\ref{fig:stbft}d, we observe that increasing the ETL doping concentration results in an improvement in PCE but has only a slight effect on \( T_{92} \), as the ETL doping concentration improvement enhances charge transport but does not significantly impact the degradation rate over time. Here, \(T_{92}\) initially increases with ETL doping concentration and then begins to decrease, following a concave parabolic trend within a narrow window of approximately 550–650 hours.

It is worth mentioning that variation in ETL thickness (\( T_{\mathrm{ETL}} \)) has a negligible impact on both the \( J\text{-}V \) characteristic curves and stability, as shown in Figure~\ref{fig:tetl}a and b, respectively. Therefore, \( T_{\mathrm{ETL}} \) is not considered as a feature.

\subsubsection{Data Clustering, Labelling and Sampling}

\begin{figure}[tb]
    \centering
    \includegraphics[width=\linewidth]{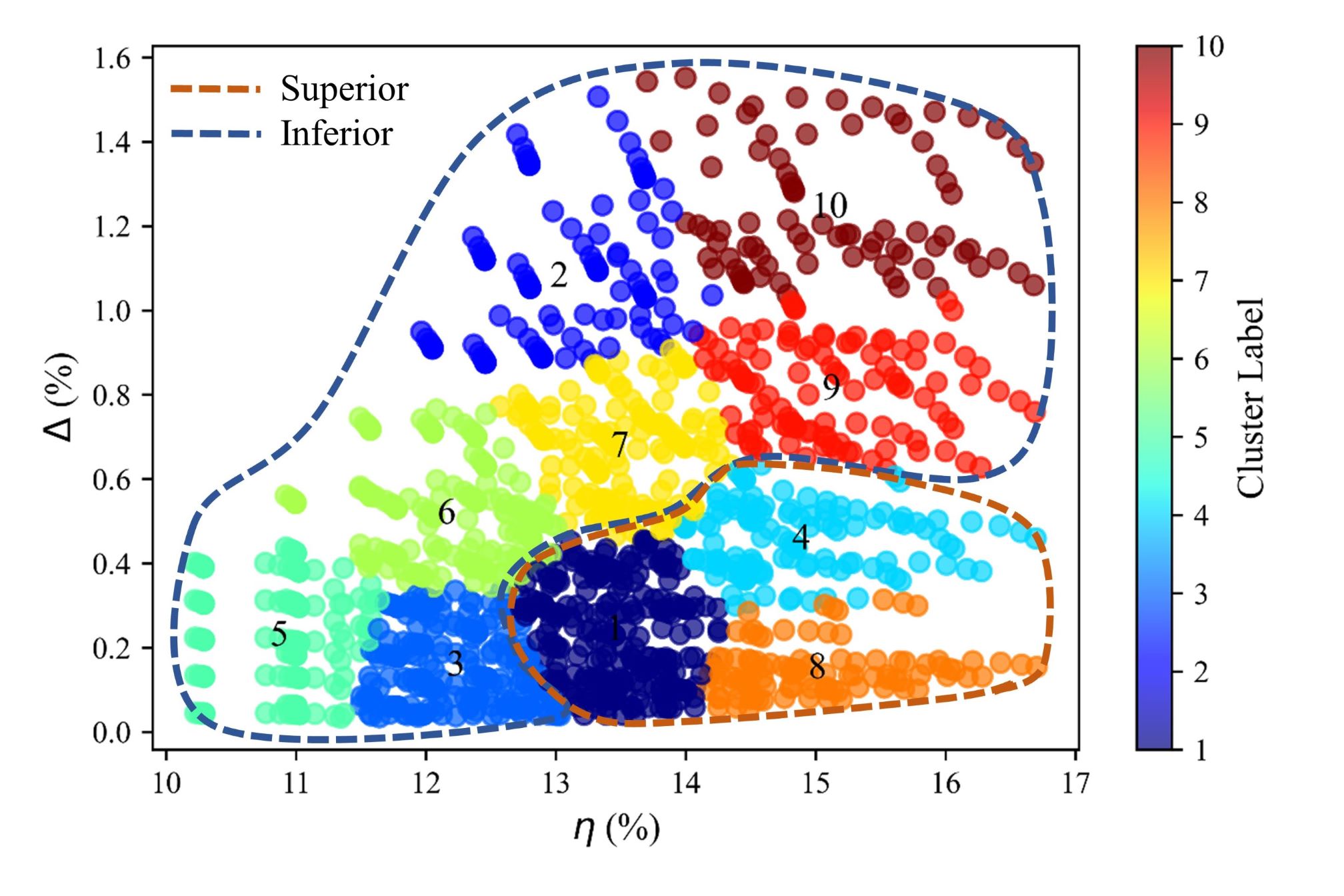}
    \caption{K-means clustering results with 'Superior' instances in green and 'Inferior' instances in orange.}
    \label{fig:clusters}
\end{figure}

The data labeling process involved classifying the dataset using two key targets, $\eta$ and $\Delta$, which were selected for the clustering analysis. Since KMeans clustering \cite{sinaga2020unsupervised} operates on numerical data, we applied MinMax scaling to these target variables to normalize them within a range of 0 to 1 \cite{pagan2023investigating}. This ensures that both features contribute equally to the clustering process. The normalized data was then passed through KMeans clustering with ten clusters, as shown in Figure~\ref{fig:clusters}. This number (ten) was selected based on heuristic experimental tuning to ensure balanced granularity for downstream classification tasks, with clusters effectively separating samples exhibiting both higher PCE and stability. KMeans works by iteratively partitioning the data into clusters, with data points assigned to the nearest centroids and centroids updated until convergence \cite{ghosh2009k}.

Upon analyzing the figure, it is observed that three out of the ten clusters (i.e., clusters 1, 4 and 8 in Figure~\ref{fig:clusters}) corresponded to cells with higher efficiency and lower degradation, while the other clusters represented cells with lower efficiency and higher degradation. This aligns with the expected behavior, where cells with higher initial PCE tend to have lower degradation over time, which is a desirable feature in PSCs with superior behavior. Based on these observed characteristics and our defined benchmark, the first, fourth, and eighth clusters are labeled as \textit{Superior}, and the remaining clusters are labeled as \textit{Inferior}, representing cells with lower efficiency and higher degradation. The labels for the two classes are then added to the original dataset, allowing for future machine learning model training focused on classifying cell performance as either \textit{Superior} or \textit{Inferior}. 

\begin{figure}[tb]
    \centering
    \begin{subfigure}{.95\linewidth}
        \includegraphics[width=\textwidth]{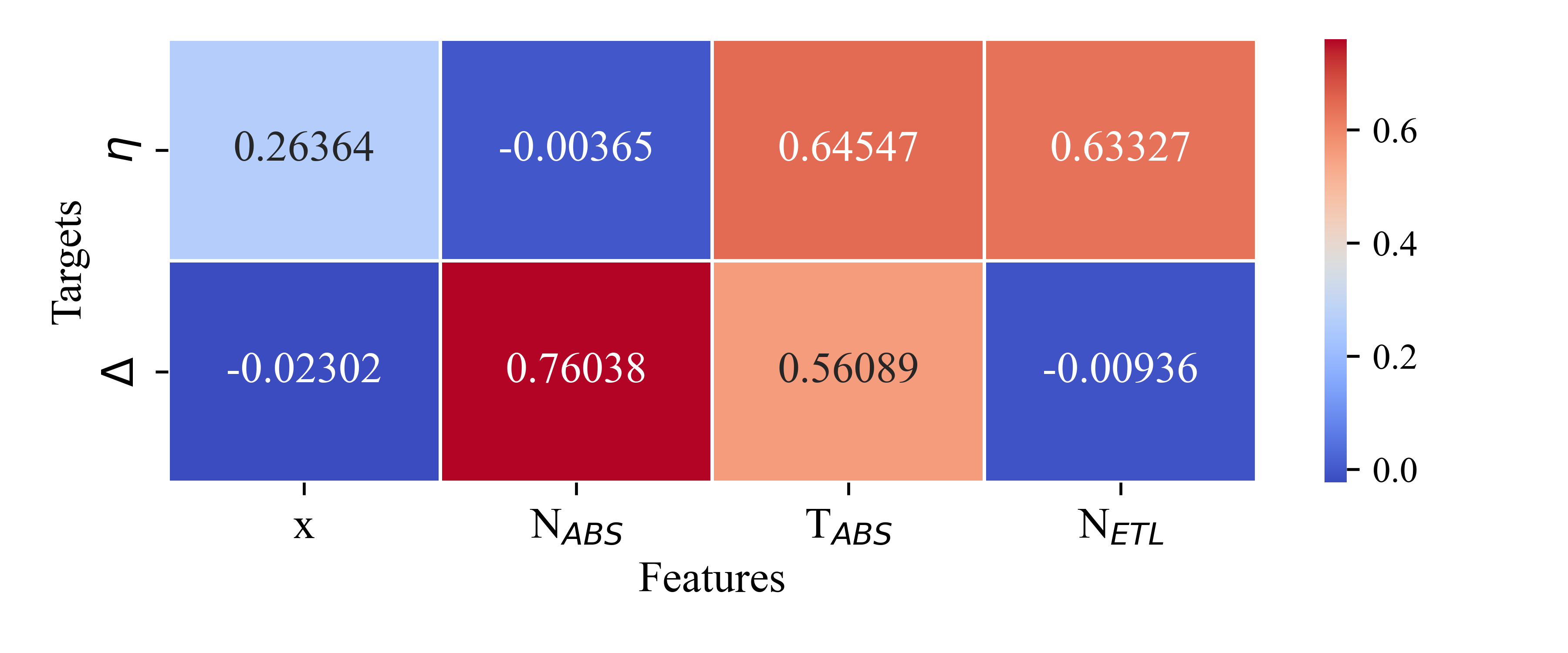}
        \caption{Correlation scores}
        \label{fig:corr}
    \end{subfigure}
    \begin{subfigure}{.95\linewidth}
        \includegraphics[width=\textwidth]{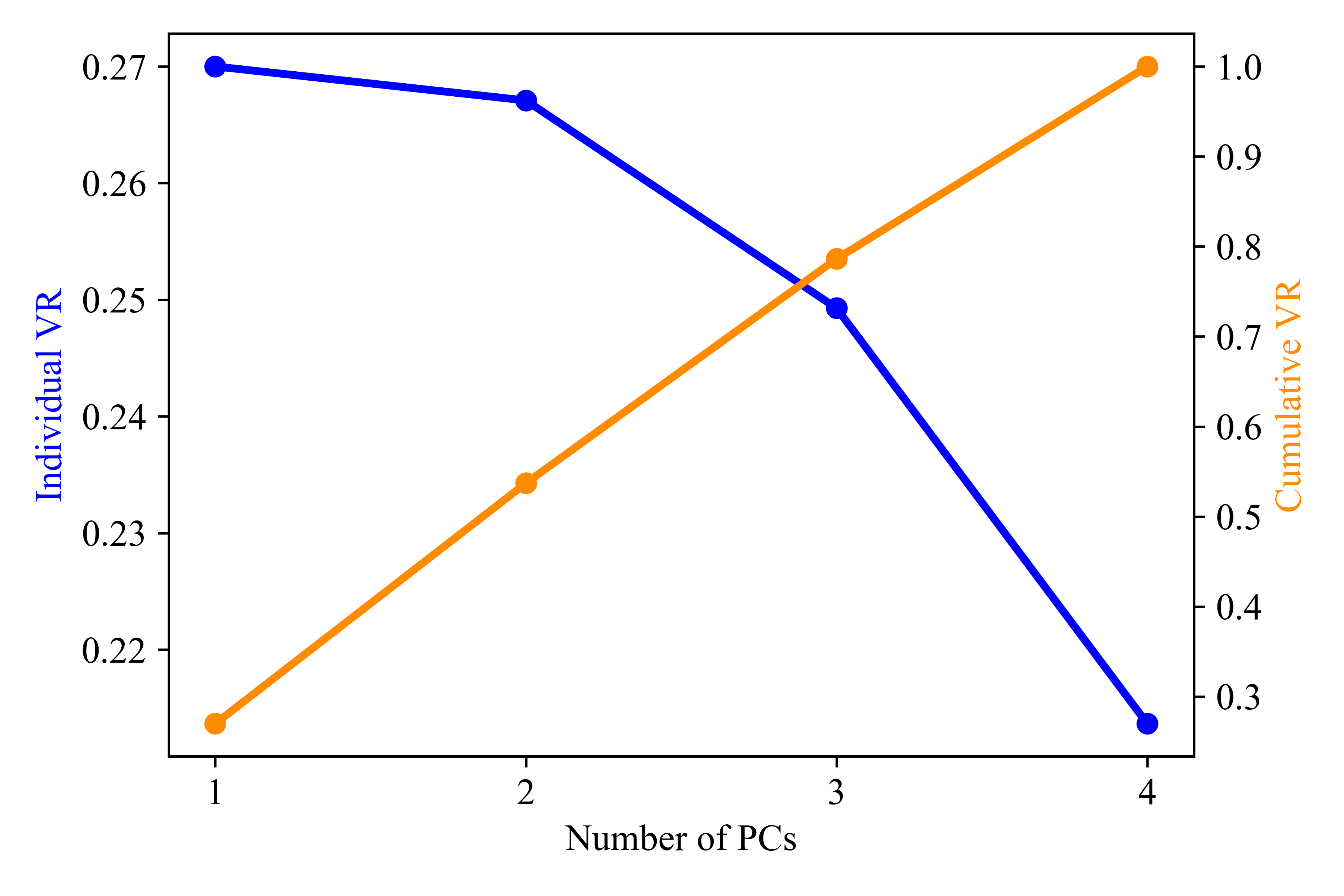}
        \caption{Variance ratio of PCs}
        \label{fig:pcs}
    \end{subfigure}
    \begin{subfigure}{.95\linewidth}
        \includegraphics[width=\textwidth]{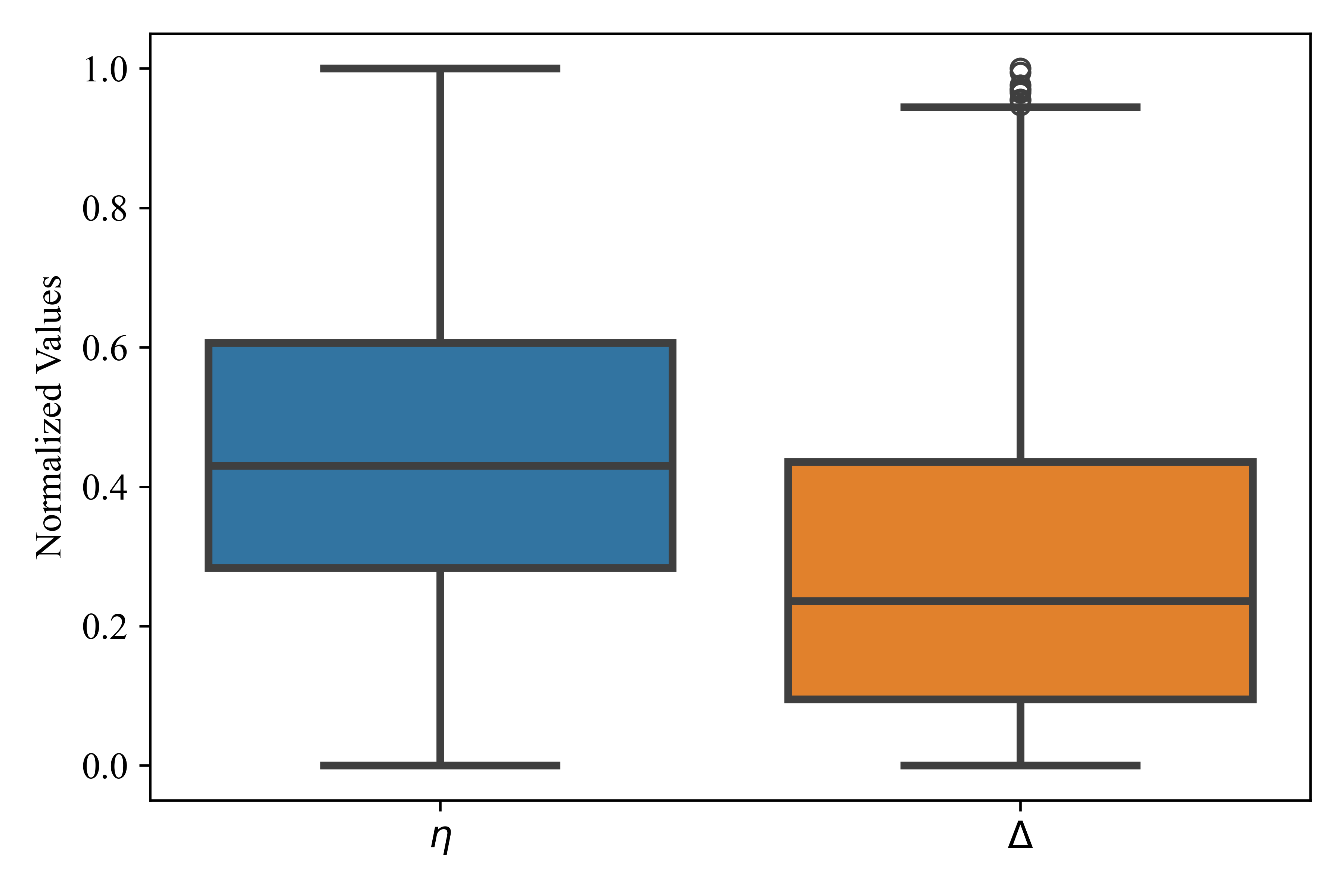}
        \caption{Boxplot of targets}
        \label{fig:bxplot}
    \end{subfigure}
    \caption{Data analysis.}
    \label{fig:cpb}
\end{figure}

After labeling, it is observed that the new dataset had 477 samples of the \textit{Superior} class and 1173 samples of the \textit{Inferior} class, resulting in an imbalanced dataset that could lead to biased predictions when using ML models \cite{chakraborty2021bias}. To address this class imbalance, we applied the Synthetic Minority Over-sampling Technique (SMOTE), which generated synthetic samples for the minority class \cite{chawla2002smote}, in this case, for the \textit{Superior} class, thereby balancing the dataset. The resampled dataset, containing almost equal representations of both \textit{Superior} and \textit{Inferior} classes, was then stored for use in training machine learning models. This balanced dataset can now serve as a reference for testing the performance of trained classifiers to predict the quality of optimized results, potentially validating them as \textit{Superior} or \textit{Inferior} by cross-verifying the optimized data through different approaches.

\begin{comment}
    
\begin{figure}[tb]
    \centering
    % Left side (two images stacked)
    \begin{minipage}{0.44\linewidth}
        \centering
        \subfloat[Correlation scores]{%
            \includegraphics[width=\linewidth]{figs/corr.png}
        } \\
        \subfloat[Variance ratio of PCs]{%
            \includegraphics[width=\linewidth]{figs/pcs.png}
        }
    \end{minipage}
    % Right side (one image)
    \begin{minipage}{0.54\linewidth}
        \centering
        \subfloat[Boxplot of targets]{%
            \includegraphics[width=\linewidth]{figs/bxplot.png}
        }
    \end{minipage}

    \caption{Data analysis.}
    \label{fig:cpb}
\end{figure}
    
\end{comment}

\subsubsection{Exploratory Data Evaluation}

These relationships between features and targets are observed in Figure~\ref{fig:cpb}a, where the correlation between the target \(\eta\) and \(\Delta\) with the 4 features indicates a very high correlation in degradation with absorber thickness and defect density, while PCE is highly correlated with all features except defect density, as it was not varied widely to precisely observe its impact on stability within a practical range of values \cite{jan2022influence}. However, from Figure~\ref{fig:jvft}b, it is evident that with a wide range of variation in defect density, there is a significant effect on the fill factor and PCE. In Figure~\ref{fig:cpb}b, the principal components (PCs) represent the projection of the data onto the PCs, highlighting the most significant variance directions in the dataset \cite{malek2024predicting}, with their individual variance ratios (VR) and cumulative VR observed. The individual VRs showed minimal variation, indicating similar contributions from each principal component, while the cumulative VR increased almost linearly, suggesting a well-structured dataset where most of the variance is captured progressively by the principal components, with each component adding incremental information. In Figure~\ref{fig:cpb}c, the boxplot of \(\eta\) and \(\Delta\) is shown, where the interquartile range is almost the same for both. However, the minimum to maximum range is slightly higher for \(\eta\), while \(\Delta\) has a few outliers, suggesting that the prediction of \(\Delta\) would be more challenging than predicting \(\eta\). These outliers indicate greater variability and noise in \(\Delta\), making it harder for ML models to predict precisely.

\subsection{ML Regressor Performance}

\begin{table*}[tb]
\centering
\caption{Model Performance: $R^2$ and RMSE values for different regressors.}
\label{tab:ml_prfrm}
\resizebox{\linewidth}{!}{
\begin{tabular}{lccccccccccccc}
\hline
\multirow{2}{*}{\textbf{Types}} & \multirow{2}{*}{\textbf{Model}} & \multicolumn{6}{c}{\textbf{$R^2$}} & \multicolumn{6}{c}{\textbf{RMSE}} \\
\cline{3-14}
& & \textbf{$J_{SC}$} & \textbf{$V_{OC}$} & \textbf{$FF$} & \textbf{$\eta$} & \textbf{$\Delta$} & \textbf{$\frac{\eta - \Delta}{2}$} & \textbf{$J_{SC}$} & \textbf{$V_{OC}$} & \textbf{$FF$} & \textbf{$\eta$} & \textbf{$\Delta$} & \textbf{$\frac{\eta - \Delta}{2}$} \\
\hline
\multirow{5}{*}{\rotatebox{90}{\textbf{Classical}}} 
& LR & 0.974 & 0.939 & 0.893 & 0.881 & 0.901 & 0.846 & 0.1867 & 0.0209 & 2.6935 & 0.4717 & 0.1144 & 0.235\\
& DT & \textbf{\textit{1.00}} & \textbf{\textit{1.00}} & \textbf{\textit{1.00}} & \textbf{\textit{1.00}} & 0.998 & 0.999 & \textbf{\textit{2.38$\times$10$^{-5}$}} & \textbf{\textit{3.30$\times$ 10$^{-5}$}} & \textbf{\textit{0.0211}} & \textbf{\textit{0.0046}} & 0.0160 & 0.0221 \\
& SVR & 0.996 & 0.592 & 0.995 & 0.998 & 0.980 & 0.986 & 0.0716 & 0.0540 & 0.5547 & 0.0662 & 0.0511 & 0.0721 \\
& K-NN & 0.979 & 0.970 & 0.952 & 0.953 & 0.994 & 0.969 & 0.1681 & 0.0146 & 1.7994 & 0.2954 & 0.0275 & 0.1048 \\
& PR-4 & 0.999 & 0.997 & 0.998 & \textit{1.00} & \textit{0.999} & \textbf{1.00} & 3.15 $\times 10^{-4}$ & 0.0045 & 0.4080 & \textit{0.0179} & \textit{0.0117} & \textbf{0.0124} \\
\hline
\multirow{3}{*}{\textbf{\shortstack{Ense-\\mble}}}
& RF & 1.00 & 0.999 & 1.00 & 1.00 & 0.999 & 0.999 & $2.39 \times 10^{-5}$ & 8.25$\times 10^{-4}$ & 0.0321 & 0.0182 & 0.0134 & 0.0163 \\
& XGB & 1.00 & 0.999 & 1.00 & 1.00 & 0.999 & 0.999 & $1.13 \times 10^{-4}$ & 0.0011 & 0.0221 & 0.0182 & 0.0198 & 0.0194\\
& AB & 1.00 & 0.942 & 0.965 & 0.948 & 0.968 & 0.955 & 0.0052 & 0.0204 & 1.5415 & 0.3106 & 0.0649 & 0.1275 \\
\hline
\multirow{2}{*}{\textbf{Deep}} 
& MLP & 0.995 & 0.984 & 1.00 & 0.997 & \textbf{\textit{1.00}} & 0.998 & 0.0805 & 0.0106 & 0.1779 & 0.0754 & \textbf{\textit{0.0094}} & 0.0295 \\
& CNN & 0.678 & 0.976 & 0.973 & 0.997 & 0.995 & 0.991 & 0.6630 & 0.0132 & 1.3407 & 0.0785 & 0.0255 & 0.0575 \\
\hline
\end{tabular}}
\end{table*}

We evaluated and compared the performance of various machine learning models, including classical, ensemble, and neural network models, on the dataset with five targets: \(J_{SC}\), \(V_{OC}\), \(FF\), \(\eta\), and \(\Delta\). Each model was assessed based on two key metrics: the coefficient of determination (\(R^2\)) and RMSE, to evaluate its predictive performance for these targets. The results of these evaluations are presented in Table~\ref{tab:ml_prfrm}. Classical models, such as Decision Tree (DT), achieved perfect \(R^2\) values for most targets, making it the best-performing model for all targets except \(\Delta\). For \(\Delta\), the MLP model outperformed the others, achieving the highest \(R^2\) and lowest RMSE. The 4th degree polynomial regression (PR 4) ranked second in performance for both \(\eta\) and \(\Delta\). Models like LR, Support Vector (SVR), and K-Nearest Neighbors (K-NN) showed higher RMSE values, indicating potential for improvement. Among the ensemble models, Random Forest (RF) and XGBoost (XGB) delivered excellent performance, with high \(R^2\) and low RMSE for almost all targets, outperforming classical models. AdaBoost (AB), while performing well for \(J_{SC}\), showed relatively lower performance for other targets. Neural network-based models, such as Convolutional Neural Networks (CNN), demonstrated strong performance, particularly for \(\eta\) and \(\Delta\), but exhibited higher RMSE values for the other targets, suggesting that ensemble methods provided the best balanced performance. To observe the combined impact of $\eta$ and $\Delta$, we derived the $R^2$ and RMSE of the objective function $f(x, N_{\mathrm{ABS}}, T_{\mathrm{ABS}}, N_{\mathrm{ETL}}) = w_{\eta}\eta - w_{\Delta}\Delta$, which simplifies to $\frac{\eta - \Delta}{2}$ when $w_{\eta} = w_{\Delta} = 0.5$, as given in Eq.(15). The results show that PR-4 outperformed all other models in predicting this objective target.

\subsubsection{ML Predicted Characteristic Curves}

\begin{figure}[tb]
    \centering
    \begin{subfigure}{\linewidth}
        \includegraphics[width=\textwidth]{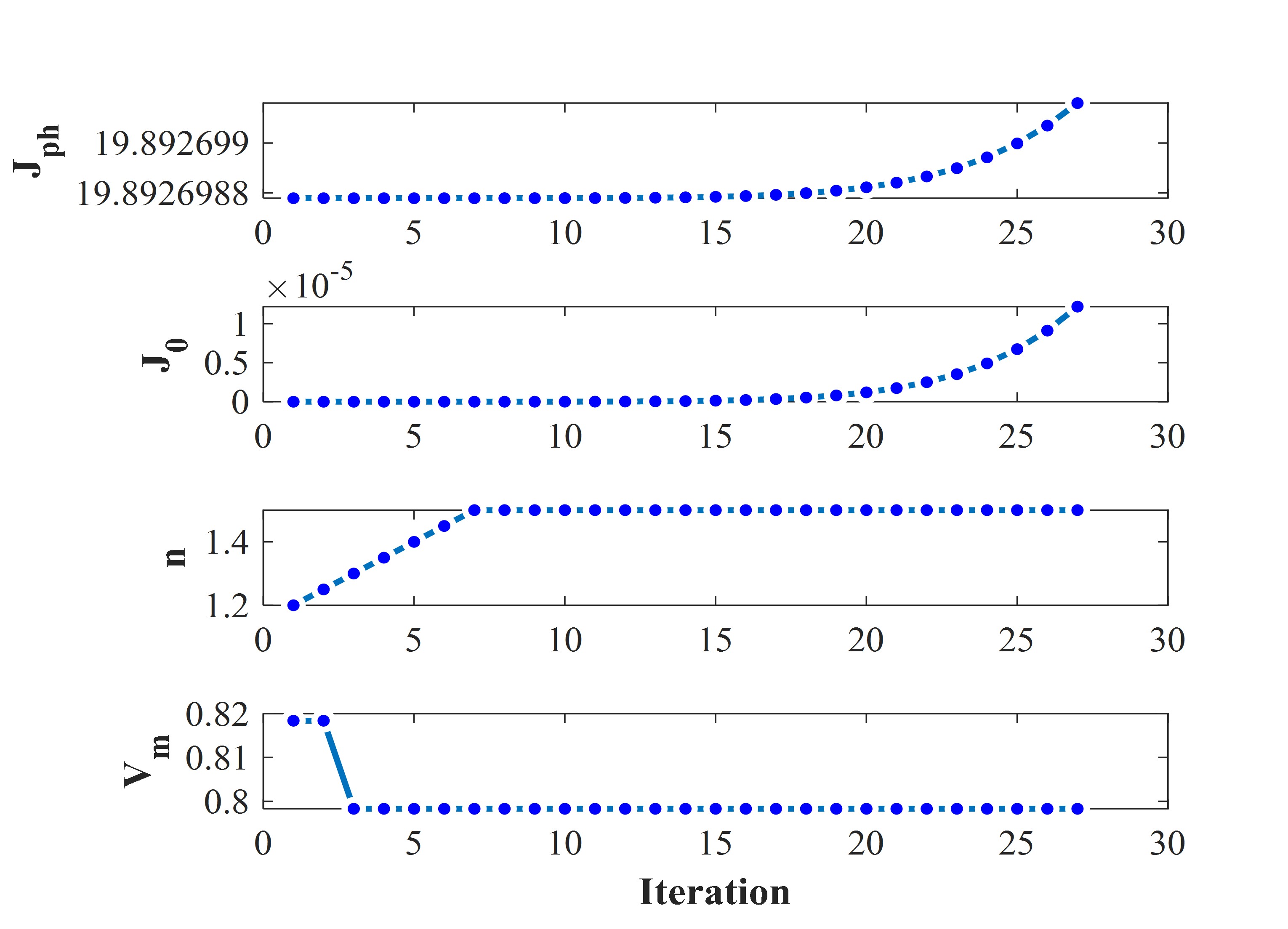}
        \caption{Values of unknowns solved using the Newton-Raphson method at each iteration until convergence.}
    \label{fig:jv_itr}
    \end{subfigure}
    \begin{subfigure}{\linewidth}
        \includegraphics[width=\textwidth]{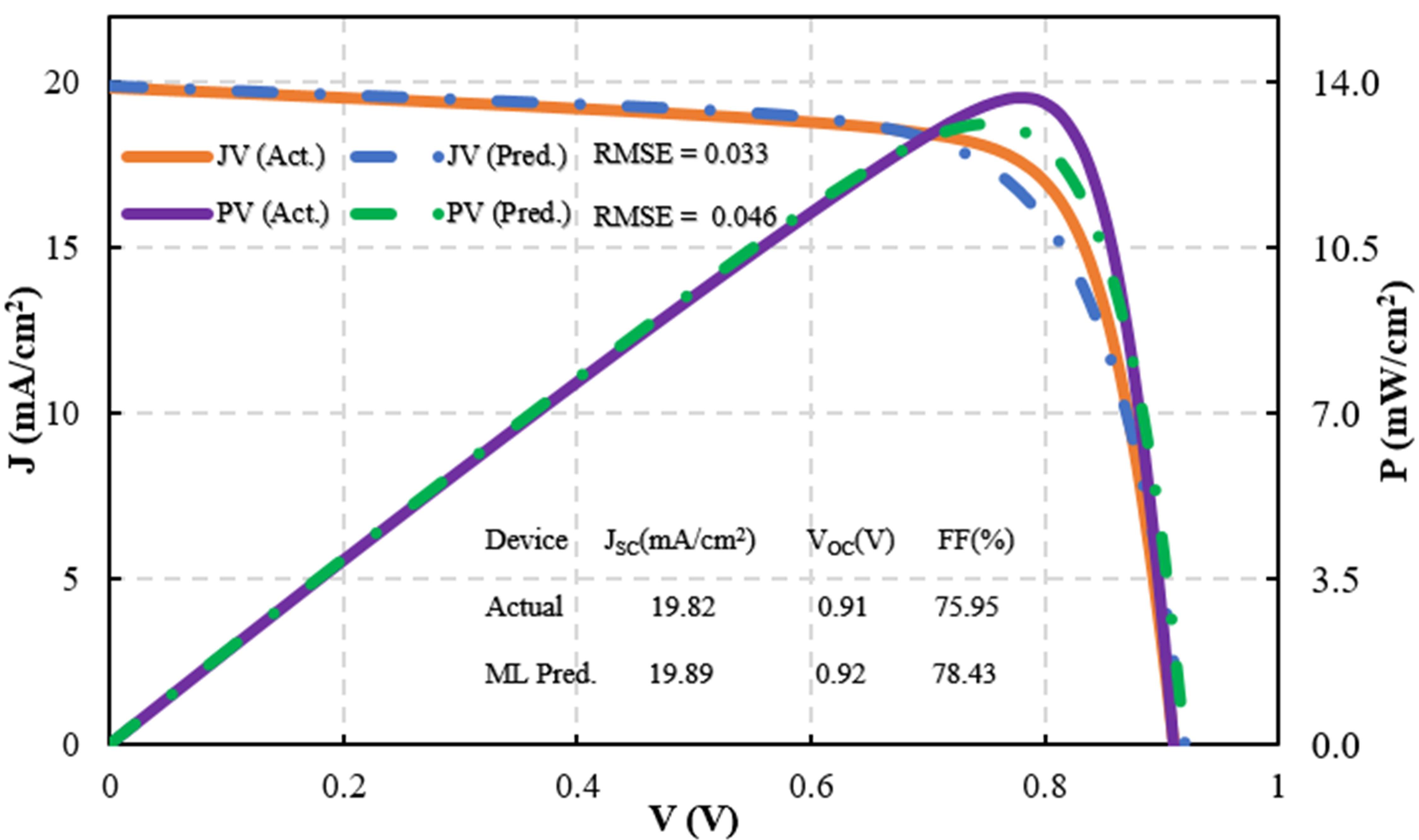}
        \caption{Characteristic curves generated by solving unknowns using ML-predicted targets.}
        \label{fig:apjv}
    \end{subfigure}
    \caption{ML predicted curves.}
    \label{fig:curve_pred}
\end{figure}

As described in the methodology section, the known values of \(J_{SC}\), \(V_{OC}\), and \(FF\) can be used for the reconstruction of the J-V and P-V curves using the Newton-Raphson method. Since the DT model achieved the highest \(R^2\) and the lowest RMSE for these targets, the trained DT model is used to predict \(J_{SC}\), \(V_{OC}\), and \(FF\) for the feature values in Table~\ref{table:layer_properties}. Using these predicted targets, the four unknowns—\(J_{ph}\), \(J_0\), \(n\), and \(V_m\)—were calculated through the Newton-Raphson method. Figure~\ref{fig:curve_pred}a shows how these four unknowns converge with each iteration. The reconstructed J-V and P-V curves are shown in Figure~\ref{fig:curve_pred}b, with RMSE values of 0.033 and 0.046 for the J-V and P-V curves, respectively. This error includes both the ML prediction and the curve reconstruction error. The low RMSE values demonstrate the effectiveness of the ML model in accurately predicting the target characteristics, emphasizing the contribution of ML and Newton-Raphson method in enhancing the precision of J-V and P-V curve reconstruction.

\subsubsection{ML Models for PCE and Degradation Prediction}

\begin{figure}[tb]
    \centering
    \begin{subfigure}{\linewidth}
        \includegraphics[width=\textwidth]{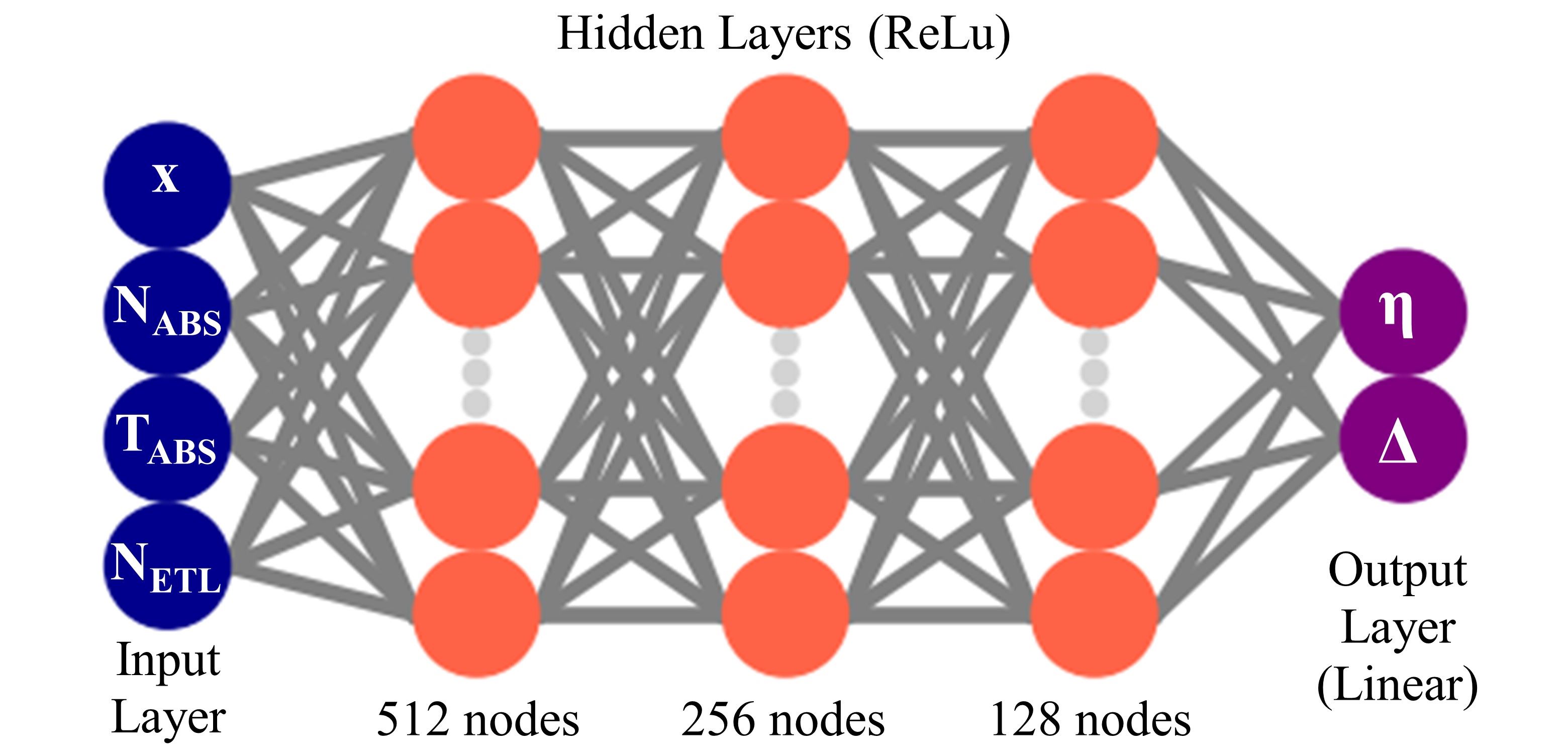}
        \caption{MLP model used with backpropagation algorithm and Adam optimizer.}
        \label{fig:mlp_arch}
    \end{subfigure}
    \begin{subfigure}{\linewidth}
        \includegraphics[width=\textwidth]{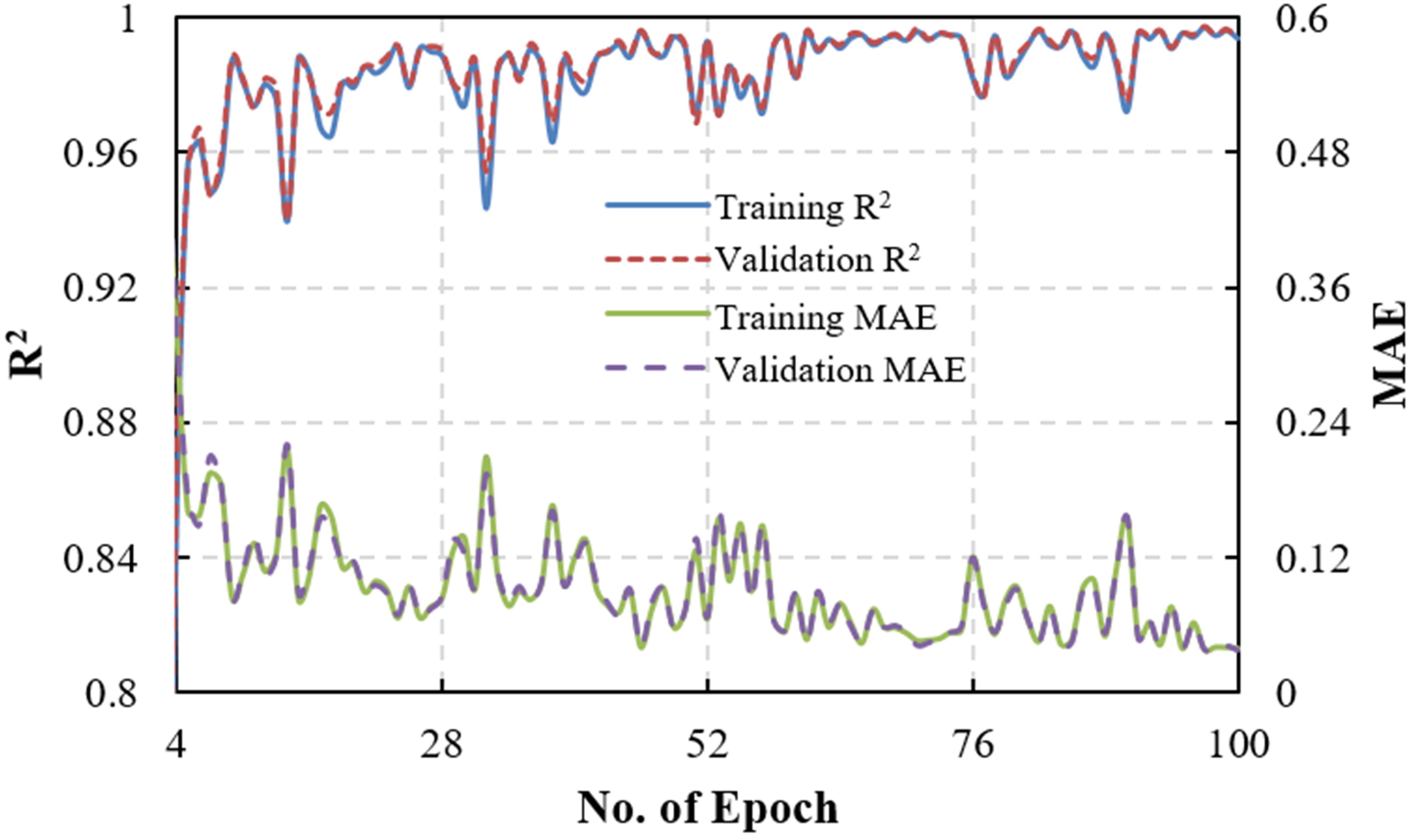}
        \caption{Performance of MLP model with epochs.}
        \label{fig:mlp_perf}
    \end{subfigure}
    \caption{MLP architecture and MLP performance.}
    \label{fig:mlp}
    
\end{figure}

As the primary goal of this study is to optimize both $\eta$ and $\Delta$, we need to identify an ML model that accurately predicts these targets. Table~\ref{tab:ml_prfrm} shows that the DT model provides the best predictions for $\eta$, while the Multilayer Perceptron (MLP) model performs best for $\Delta$. The DecisionTreeRegressor was used with the squared error criterion, best splitter, a minimum samples split of 2, and a minimum samples leaf of 1. The MLP model used in this study follows the architecture presented in Figure~\ref{fig:mlp}a, with its training and validation $R^2$ (primary y-axis) and mean absolute error (MAE) (secondary y-axis) across epochs shown in Figure~\ref{fig:mlp}b. The results indicate that validation performance remains slightly higher than training performance for most epochs, suggesting good generalization and minimal overfitting.

\begin{figure}[tb]
    \includegraphics[width=\linewidth]{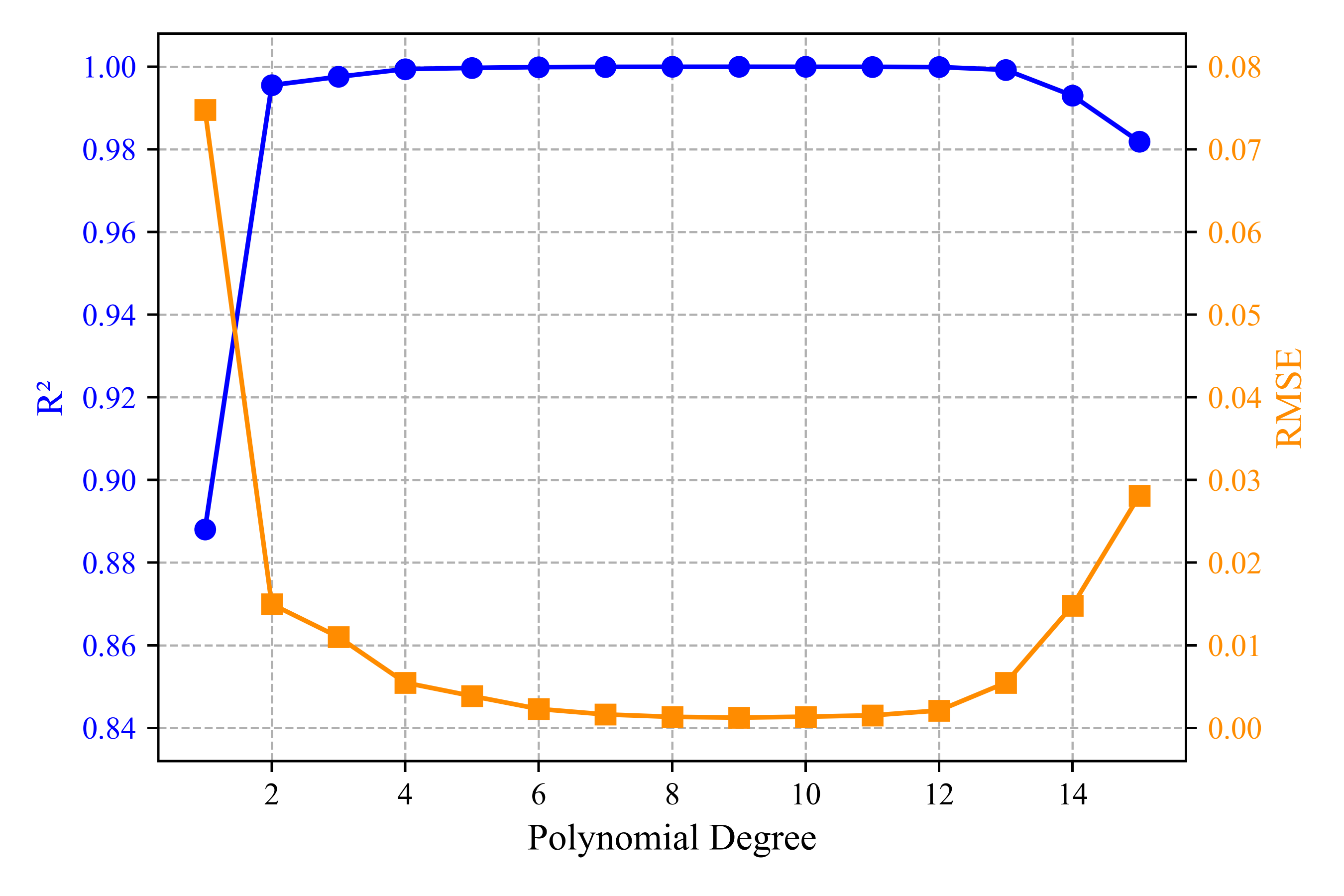}
    \caption{$R^2$ and RMSE vs. polynomial degree for test data.}
    \label{fig:poly_deg}
\end{figure}

However, since both $\eta$ and $\Delta$ are equally important, a model that balances accuracy for both targets is required. From Table~\ref{tab:ml_prfrm}, it is evident that the Polynomial Regressor (PR) provides the second-best performance for both $\eta$ and $\Delta$. Moreover, PR allows for deriving a continuous differentiable function of the four input features. The key question, however, is determining the optimal degree of the polynomial model. To address this, Figure~\ref{fig:poly_deg} illustrates the $R^2$ (primary y-axis) and RMSE (secondary y-axis) values for polynomial degrees ranging from 1 to 15. The results show an improvement in performance with increasing degree, but beyond degree 10, the performance starts degrading due to overfitting and reduced interpretability. For polynomial degrees between 3 and 14, the RMSE remains within 1\%. Though the 9th degree polynomial regression yields the best performance with the minimum RMSE and maximum $R^2$, it also entails a higher computational burden and an increased risk of overfitting when predicting beyond the dataset range, thereby reducing generalizability. In contrast, the 4th degree polynomial offers a comparable performance with a difference of only 0.00423 in RMSE and 0.00056 in $R^2$. A fourth degree polynomial was therefore selected as it provides a balance between low RMSE and model complexity, helping to mitigate overfitting for predictions beyond the training data boundary.

\begin{figure*}[b]
    \centering
    \begin{subfigure}{0.4\textwidth}
        \includegraphics[width=\textwidth]{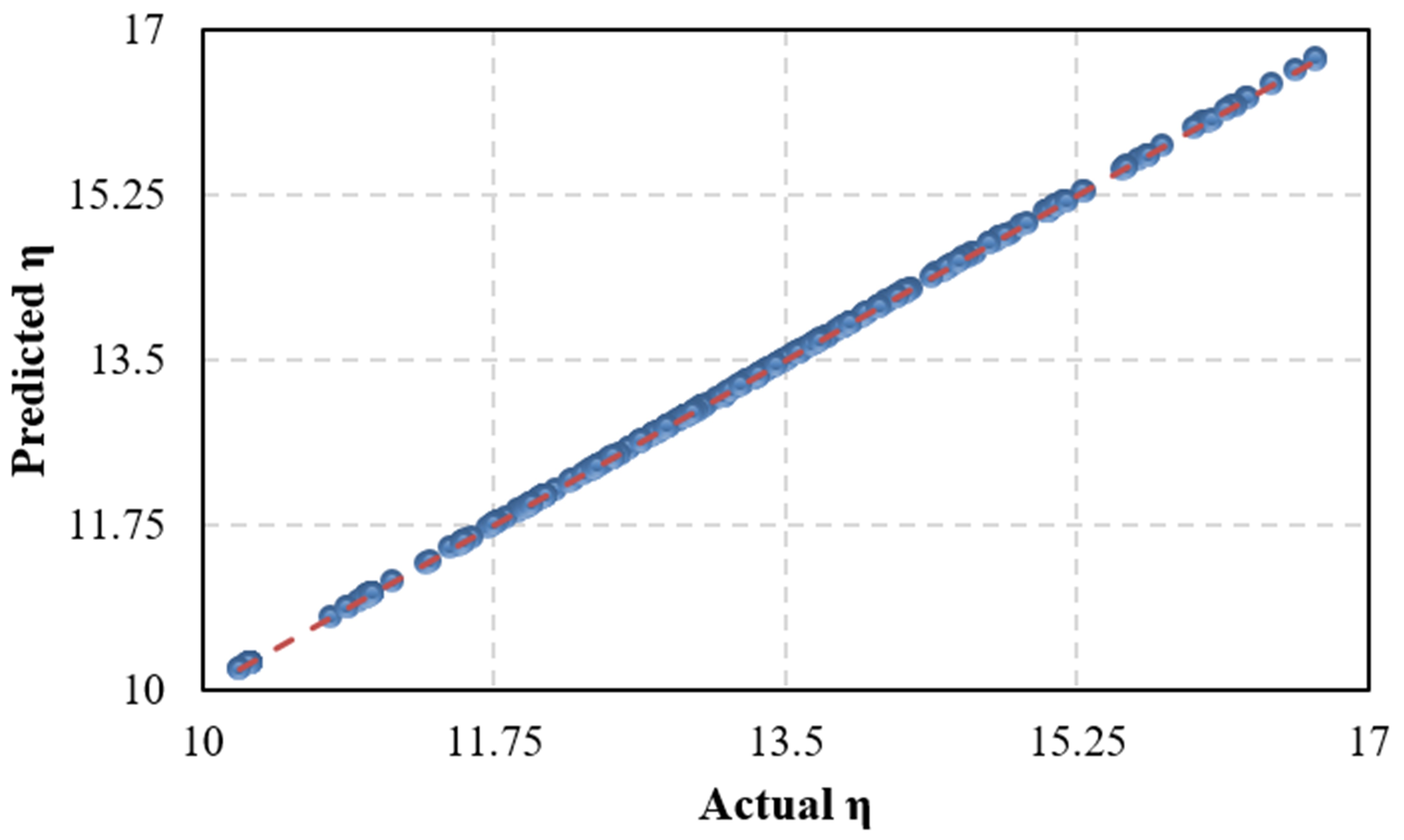}
        \caption{Prediction of $\eta$ using DT ($y = 0.9998x + 0.0023$)}
        \label{fig:etadt}
    \end{subfigure}
    \begin{subfigure}{0.4\textwidth}
        \includegraphics[width=\textwidth]{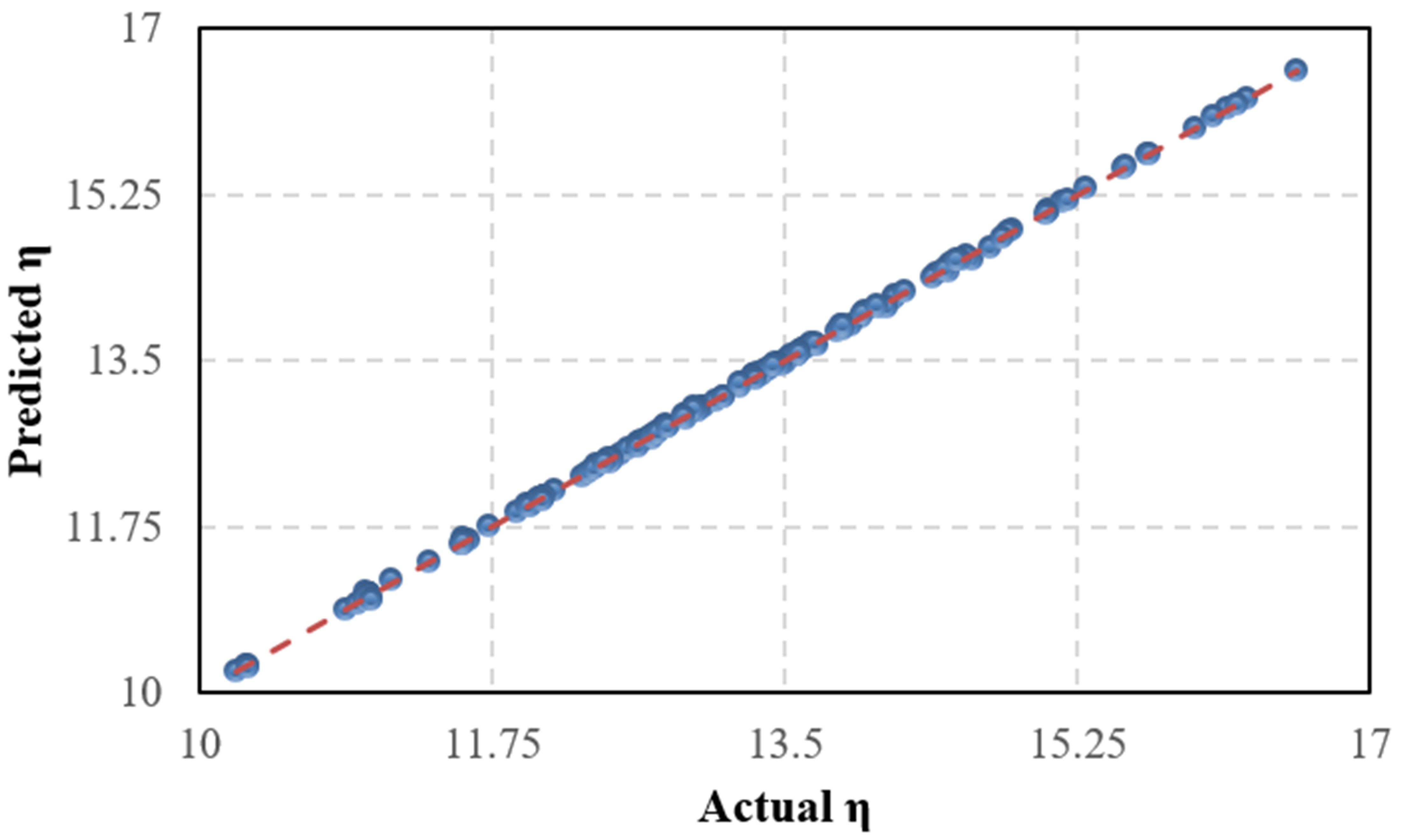}
        \caption{Prediction of $\eta$ using PR ($y = 0.9991x + 0.012$)}
        \label{fig:etapoly}
    \end{subfigure}
    \begin{subfigure}{0.4\textwidth}
        \includegraphics[width=\textwidth]{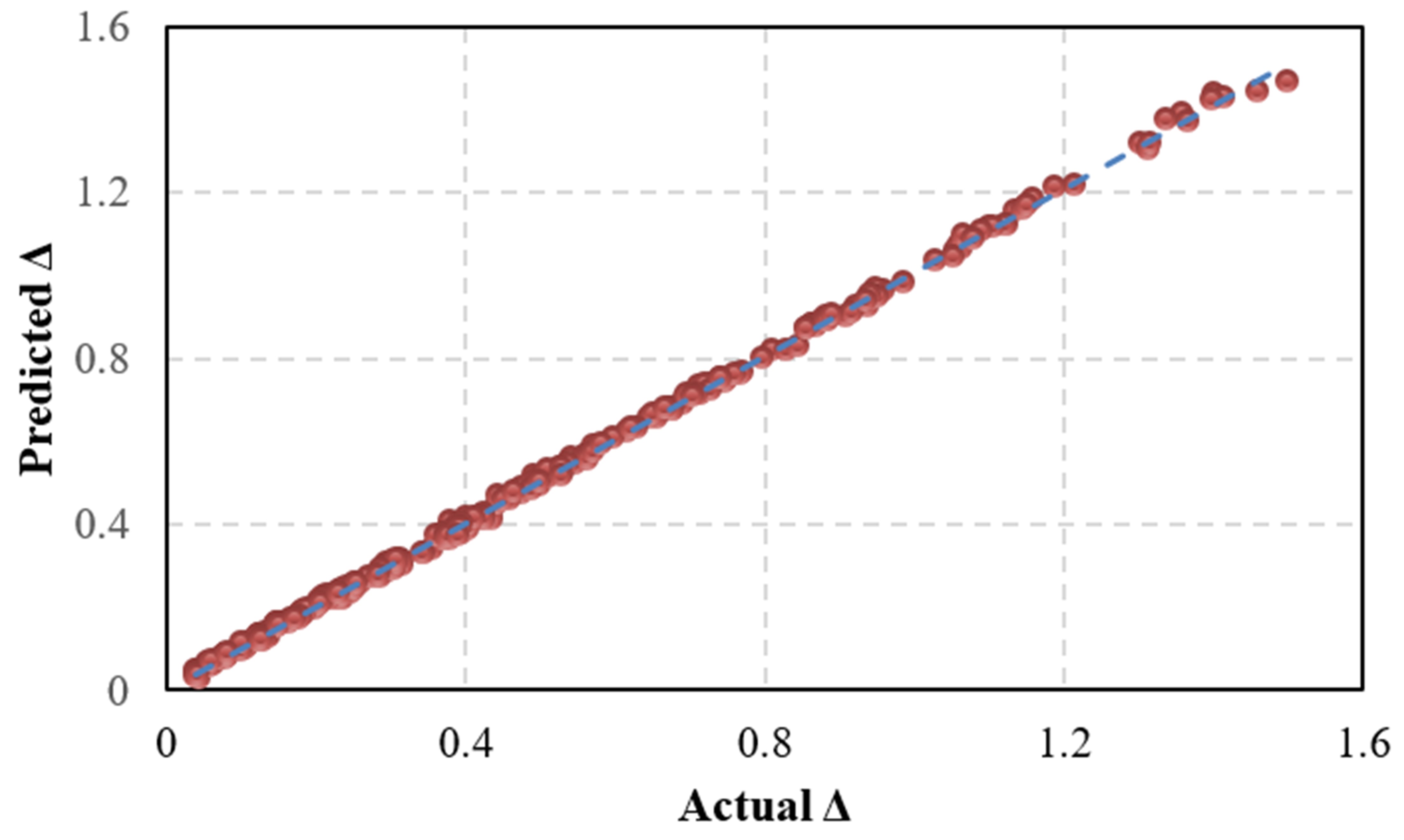}
        \caption{Prediction of $\Delta$ using MLP ($y = 1.0x + 2\times10^{-5}$)}
        \label{fig:stabmlp}
    \end{subfigure}
    \begin{subfigure}{0.4\textwidth}
        \includegraphics[width=\textwidth]{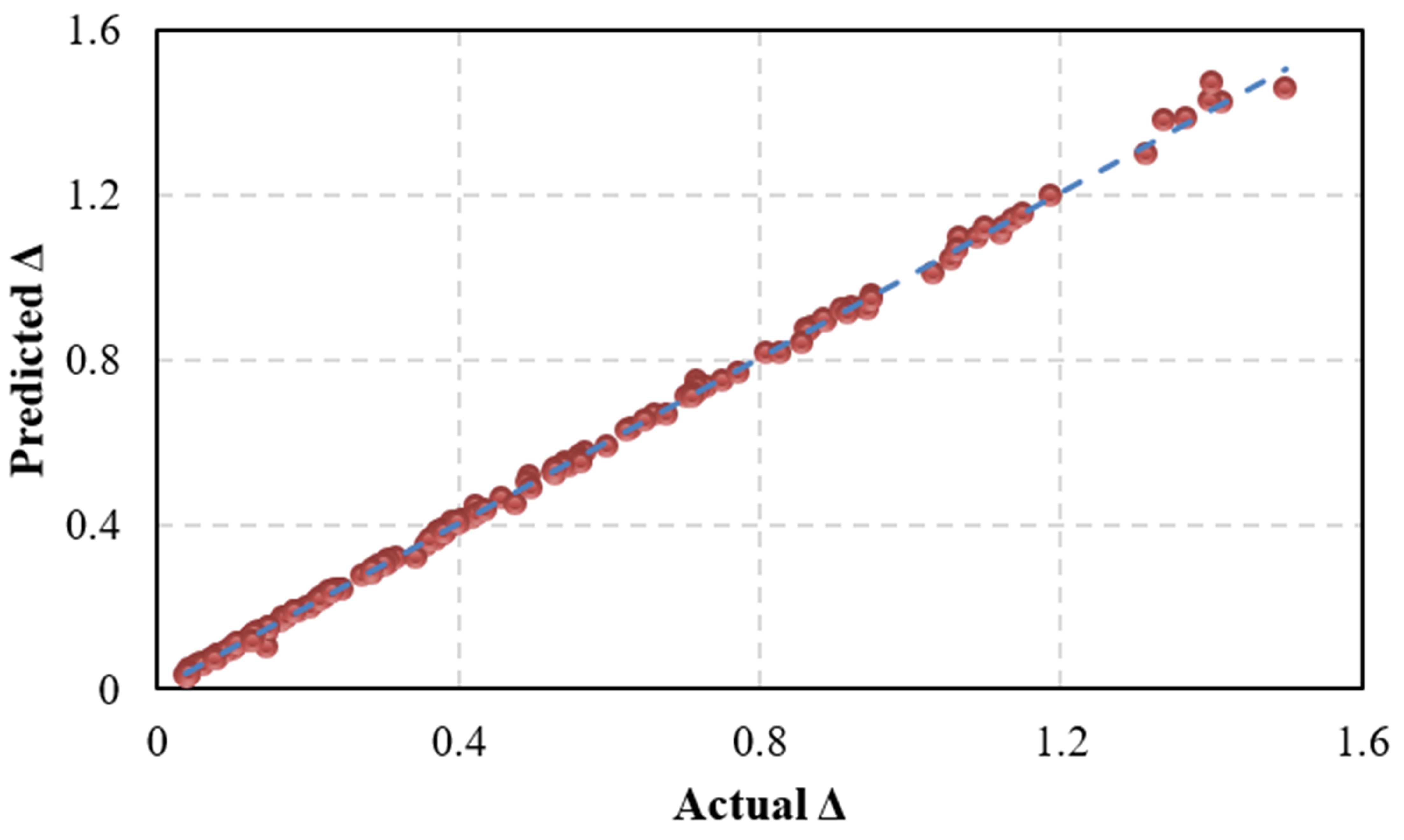}
        \caption{Prediction of $\Delta$ using PR ($y = 1.0x - 0.0023$)}
        \label{fig:stabpoly}
    \end{subfigure}
    \caption{Actual and predicted values using Decision Tree (DT), Multilayer Perceptron (MLP), and Polynomial Regressor (PR) with linearly fitted equations shown in brackets.}
    \label{fig:actprd}
\end{figure*}

To further analyze model performance, a comparison between actual and predicted values of $\eta$ (for DT and PR 4) and $\Delta$ (for MLP and PR 4) is presented in Figure~\ref{fig:actprd}. Figure~\ref{fig:actprd}a shows that the DT-predicted $\eta$ values align slightly better with the actual data compared to Figure~\ref{fig:actprd}b, where PR 4 was used. However, both models exhibit a similar fitted line slope. Similarly, Figure~\ref{fig:actprd}c indicates that MLP predicts $\Delta$ more accurately than PR 4, as shown in Figure~\ref{fig:actprd}d. The fitted line for MLP predictions has an almost perfect slope, with a slightly higher y-axis offset for PR 4, indicating a minor systematic deviation. 

Overall, these results highlight that DT and MLP provide the most accurate predictions for $\eta$ and $\Delta$, respectively. However, PR 4 serves as a robust alternative, offering a well-balanced prediction accuracy for both targets while maintaining interpretability. The result obtained indicate that ML-based predictions effectively capture underlying trends.

\subsection{Optimization of PCE and Degradation}

\begin{table}[ht]
\centering
\caption{Optimized Feature Values and Targets for PR, DT, and SCAPS.}
\begin{tabular}{lccc}
\hline
\multicolumn{2}{c}{\textbf{ML Method}} & \textbf{PR-4} & \textbf{DT} \\
\multicolumn{2}{c}{\textbf{Optimization Method}} & \textbf{L-BFGS-B} & \textbf{GA} \\
\hline
\multirow{4}{*}{Features} & $x(\%)$ & 77.5 & 74.5 \\
& $N_{ABS} \,(cm^{-3})$ & $3 \times 10^{14}$ & $3 \times 10^{14}$ \\
& $T_{ABS} \,(nm)$ & 537.5 & 487.5 \\
& $N_{ETL} \,(cm^{-3})$ & $1.75 \times 10^{17}$ & $1.64 \times 10^{17}$ \\
\hline
\multirow{2}{*}{Targets} & $\eta (\%)$ & 17.02 & 16.69 \\
& $\Delta (\%)$ & 0.48 & 0.46 \\
\hline
SCAPS & $\eta(\%)$ & 16.84 & 16.42 \\
validation & $\Delta(\%)$ & 0.50 & 0.52 \\
\hline
\end{tabular}
\label{table:features_targets}
\end{table}

\begin{figure}[b]
\centering
    \includegraphics[width=\linewidth]{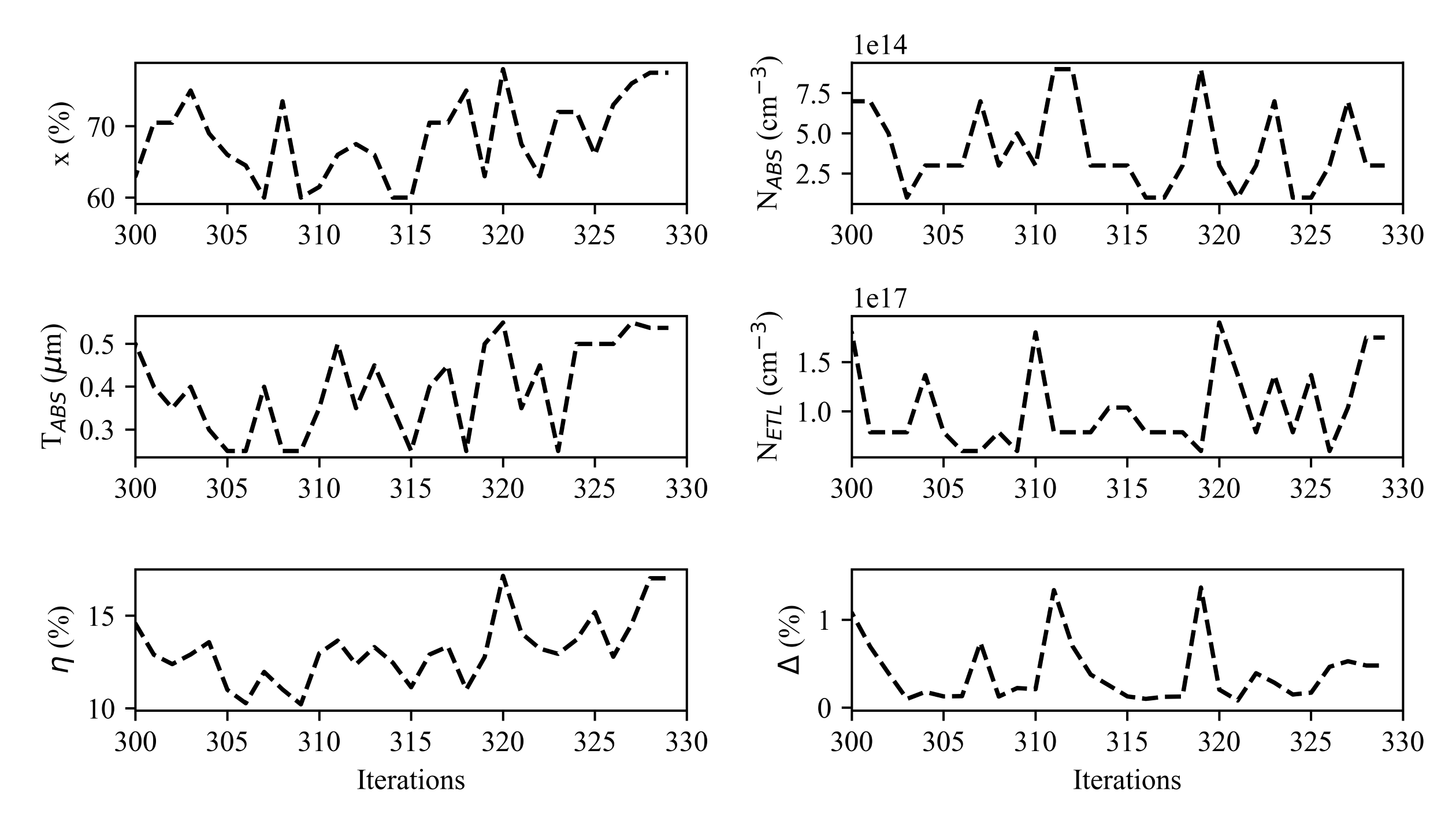}
    \caption{Key variables and targets for last few iterations during optimization using the L-BFGS-B method.}
\label{fig:opt_itr}
\end{figure}

After successfully training the PR-4 ML model, the equations for PCE, $\eta(x, N_{ABS}, T_{ABS}, N_{ETL})$ and degradation, $\Delta(x, N_{ABS}, T_{ABS}, N_{ETL})$ were derived, as provided in the Appendix. The weighted sum of these two targets, as defined in Eq. (15), was used as the objective function for optimization. The optimization was carried out using the L-BFGS-B method within the parameter ranges of $x \in [55, 78]$, $N_{ABS} \in [3\times10^{14}, 1\times10^{15}]$, $T_{ABS} \in [0.2, 0.55]$, and $N_{ETL} \in [2\times10^{16}, 1.9\times10^{17}]$. Notably, these values extend beyond the dataset range used for training the ML models.

\begin{figure}[tb]
    \centering
    \begin{subfigure}{\linewidth}
        \includegraphics[width=0.97\textwidth]{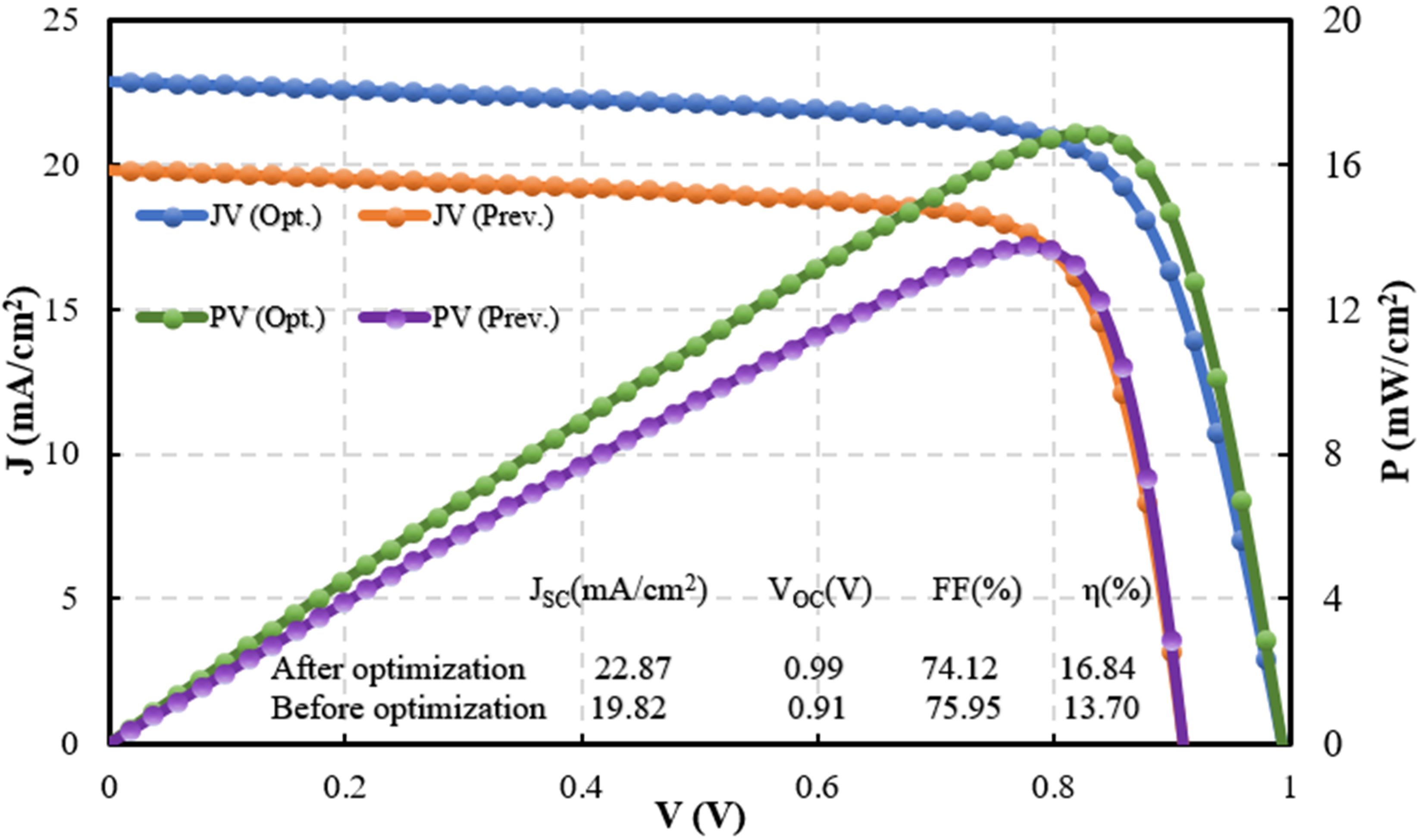}
        \caption{JV curve after and before optimization.}
        \label{fig:jvopt}
    \end{subfigure}
    \begin{subfigure}{\linewidth}
        \includegraphics[width=0.97\textwidth]{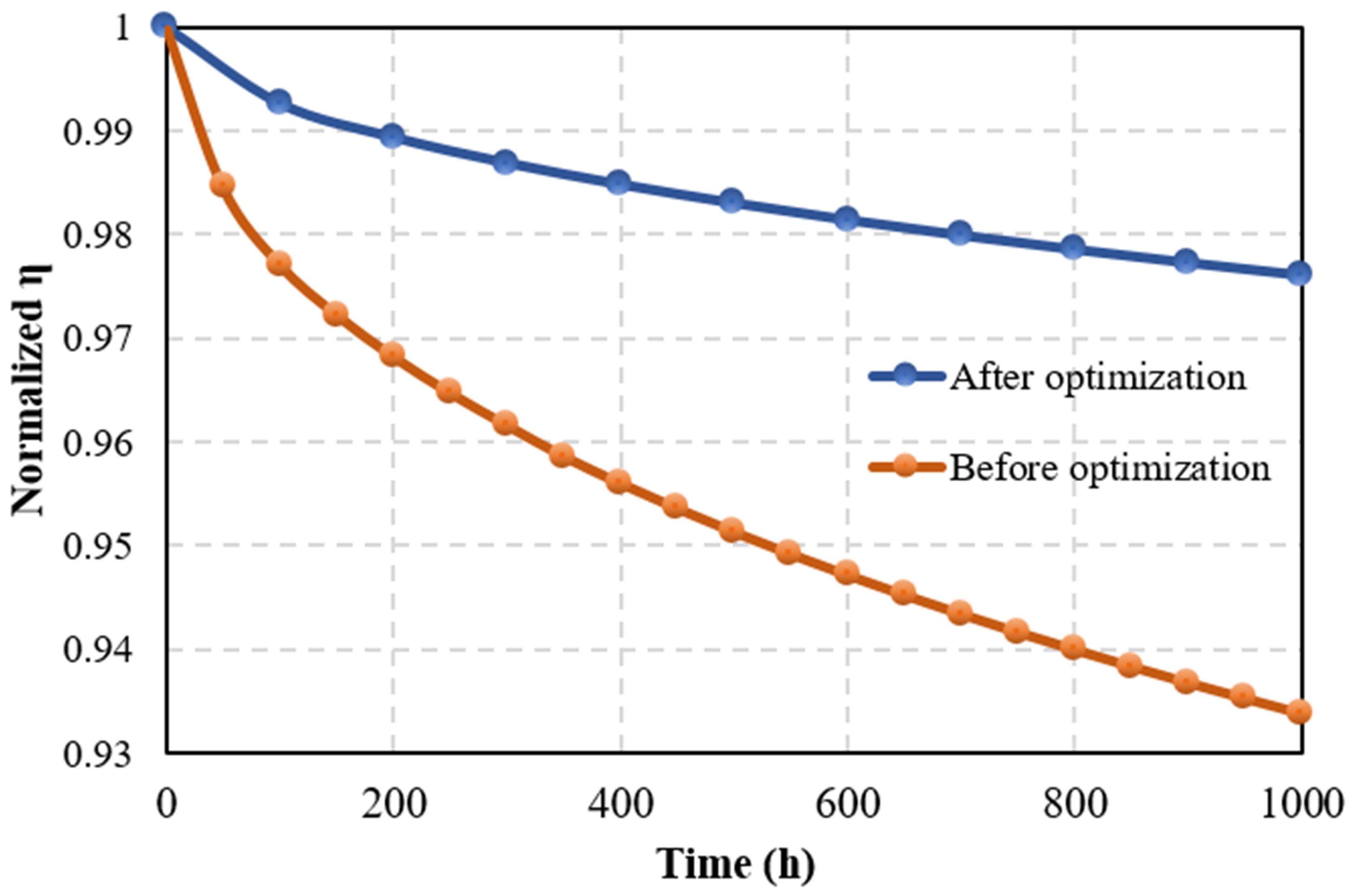}
        \caption{Stability curve after and before optimization.}
        \label{fig:stabopt}
    \end{subfigure}
    \caption{Optimized performances.}
    \label{fig:optjvstab}
\end{figure}

To compare the performance of the PR-4 and L-BFGS-B combination with other ML and optimization approaches, we used DT as the ML model—since it provided the best performance for most targets (Table~\ref{tab:ml_prfrm})—and Genetic Algorithm (GA) as the optimization method. Table~\ref{table:features_targets} presents the feature values corresponding to the best performance, i.e., maximum PCE and minimum degradation. These results were then validated using SCAPS to assess prediction errors. The table indicates slight differences in the optimized feature values, but the SCAPS validation confirms that the PR-4 and L-BFGS-B combination exhibits superior accuracy, yielding absolute prediction errors of 1.07\% for $\eta$ and 4.17\% for $\Delta$. In contrast, the DT and GA combination resulted in significantly higher absolute errors of 1.62\% for $\eta$ and 11.54\% for $\Delta$. This demonstrates the robustness of our chosen approach, highlighting PR-4’s superior generalization capability, even beyond the training data range.

Figure~\ref{fig:opt_itr} illustrates the values of features and targets over the last 20 iterations, showcasing the optimization process and its convergence trend. Using the optimized feature values, SCAPS was employed to generate characteristic curves and stability curves before and after optimization, as depicted in Figure~\ref{fig:optjvstab}a and Figure~\ref{fig:optjvstab}b, respectively. These results indicate a substantial performance improvement, with $\eta$ increasing from 13.7\% to 16.84\% and $\Delta$ decreasing from 6.615\% to 2.393\% after 1000 hours of aging.

With the optimized feature values for PCE and degradation shown in Table~\ref{table:features_targets}, the quantum efficiency is evaluated before and after optimization as a function of wavelength and photon energy, as shown in Figure~\ref{fig:qe}. The results indicate that, after optimization, the quantum efficiency improves for wavelengths ranging from 450 nm to 750 nm, demonstrating that the optimization enhances the device's ability to absorb light in this range, leading to improved performance and efficiency in the visible spectrum.

\subsection{ML Classifier Performance}

\begin{table}[tb]
\centering
\caption{Classification performances for Different Models}
\begin{tabular}{ccccc}
\hline
\textbf{Model} & \textbf{Precision} & \textbf{Recall} & \textbf{F1-Score} & \textbf{Accuracy} \\
\hline
LR & 0.83 & 0.84 & 0.83 & 0.8255 \\
SVM & 0.96 & 0.99 & 0.98 & 0.9745 \\
KNN & 0.92 & 1.00 & 0.96 & 0.9532 \\
DT & 0.97 & 0.99 & 0.98 & 0.9787 \\
RF & 0.99 & 1.00 & 0.99 & 0.9936 \\
GB & 0.97 & 1.00 & 0.98 & 0.9830 \\
AB & 0.84 & 0.94 & 0.89 & 0.8745 \\
MLP & \textbf{1.00} & \textbf{1.00} & \textbf{1.00} & \textbf{1.0000} \\
\hline
\end{tabular}
\label{tab:cls}
\end{table}

The Table~\ref{tab:cls} compares the performance of different machine learning models, including Logistic Regression (LR), SVM, KNN, DT, RF, Gradient Boosting (GB), AdaBoost, and MLP, in classifying a dataset where the 'Superior' class is treated as the true class. The features considered here are $x$, $N_{ABS}$, $T_{ABS}$, and $N_{ETL}$.

Precision for the 'Superior' class reflects the proportion of correct predictions among all instances classified as superior. MLP achieved the highest precision of 1.00, demonstrating perfect accuracy in identifying superior cases. SVM followed closely with 0.96 precision, indicating strong performance. Other models such as RF, KNN, and DT also showed high precision values above 0.92, signifying good prediction accuracy for the superior class.

Recall for the 'Superior' class represents the proportion of actual superior instances correctly identified by each model. MLP again excelled with a recall of 1.00, correctly identifying all superior instances. SVM and RF also performed well with recall values of 0.99, ensuring a low number of false negatives. KNN showed perfect recall, reflecting its ability to identify all superior cases.

F1-Score, which balances precision and recall, was highest for MLP with a score of 1.00, demonstrating its best overall performance in correctly identifying superior cells. The RF model followed closely with an F1-score of 0.99, indicating it provided a very well-balanced performance. SVM and KNN achieved F1-scores of 0.98 and 0.96, respectively, reflecting their reliable performance in the classification task.

Accuracy, which measures the overall correctness of the model, was highest for MLP with a perfect accuracy of 100\%, followed by RF with 99.36\%. SVM and DT also showed strong performance with accuracies of 97.45\% and 97.87\%, respectively. Models like AdaBoost, while providing solid results, had lower accuracy scores (87.45\%) compared to the other models. The trained MLP model can therefore be used to predict whether a similar device will exhibit superior performance based on its corresponding fabrication parameters.

For the features corresponding to the optimized performance shown in Table~\ref{table:features_targets}, all the machine learning classifiers successfully classified the optimized result as a \textit{Superior} type cell, further validating the models' reliability in distinguishing between superior and inferior instances.

\begin{figure}[tb]
    \centering
    \begin{subfigure}{\linewidth}
        \includegraphics[width=.98\textwidth]{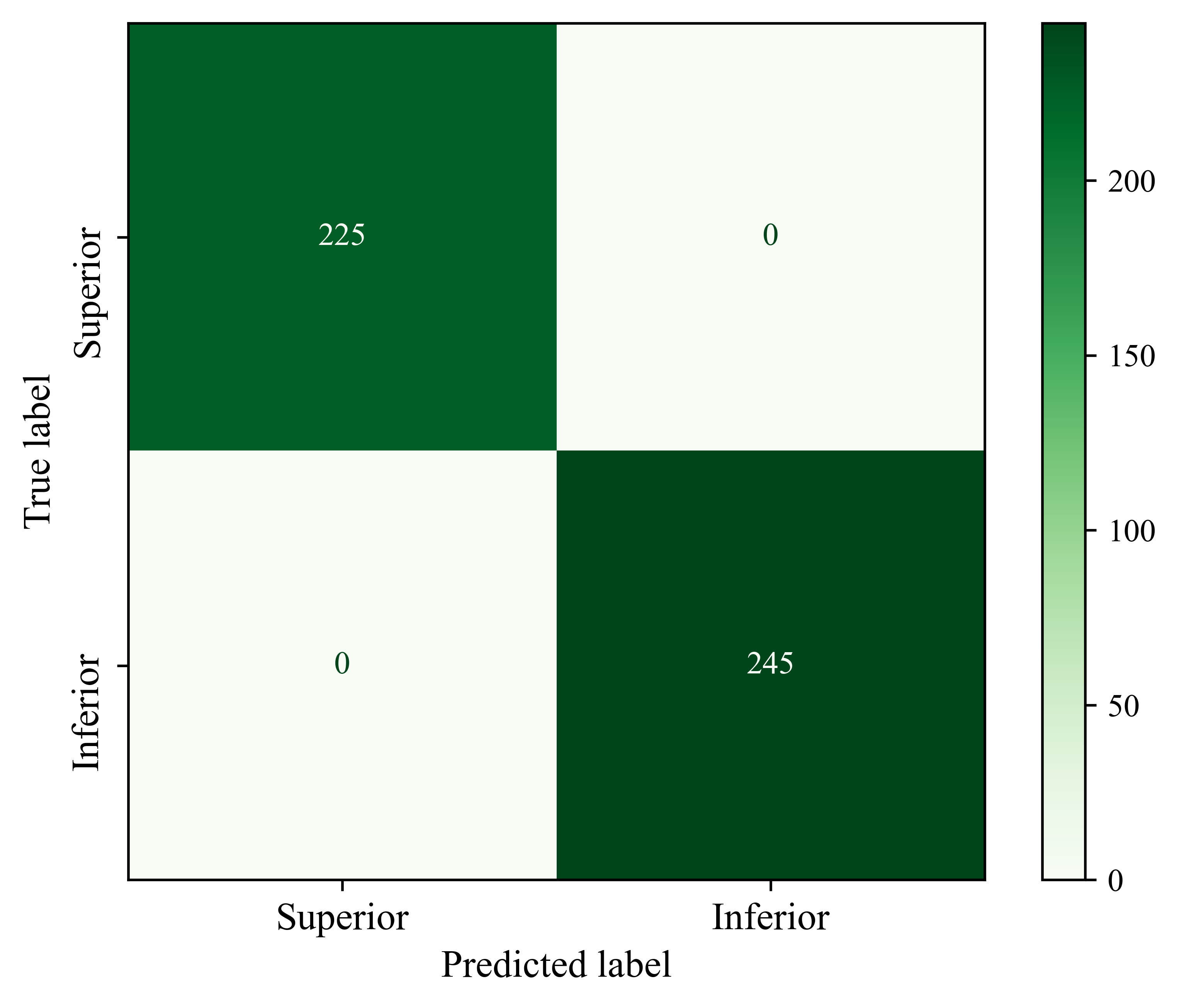}
        \caption{Confusion matrix using the MLP classifier.}
        \label{fig:mlpcm}
    \end{subfigure}
    \begin{subfigure}{\linewidth}
        \includegraphics[width=.98\textwidth]{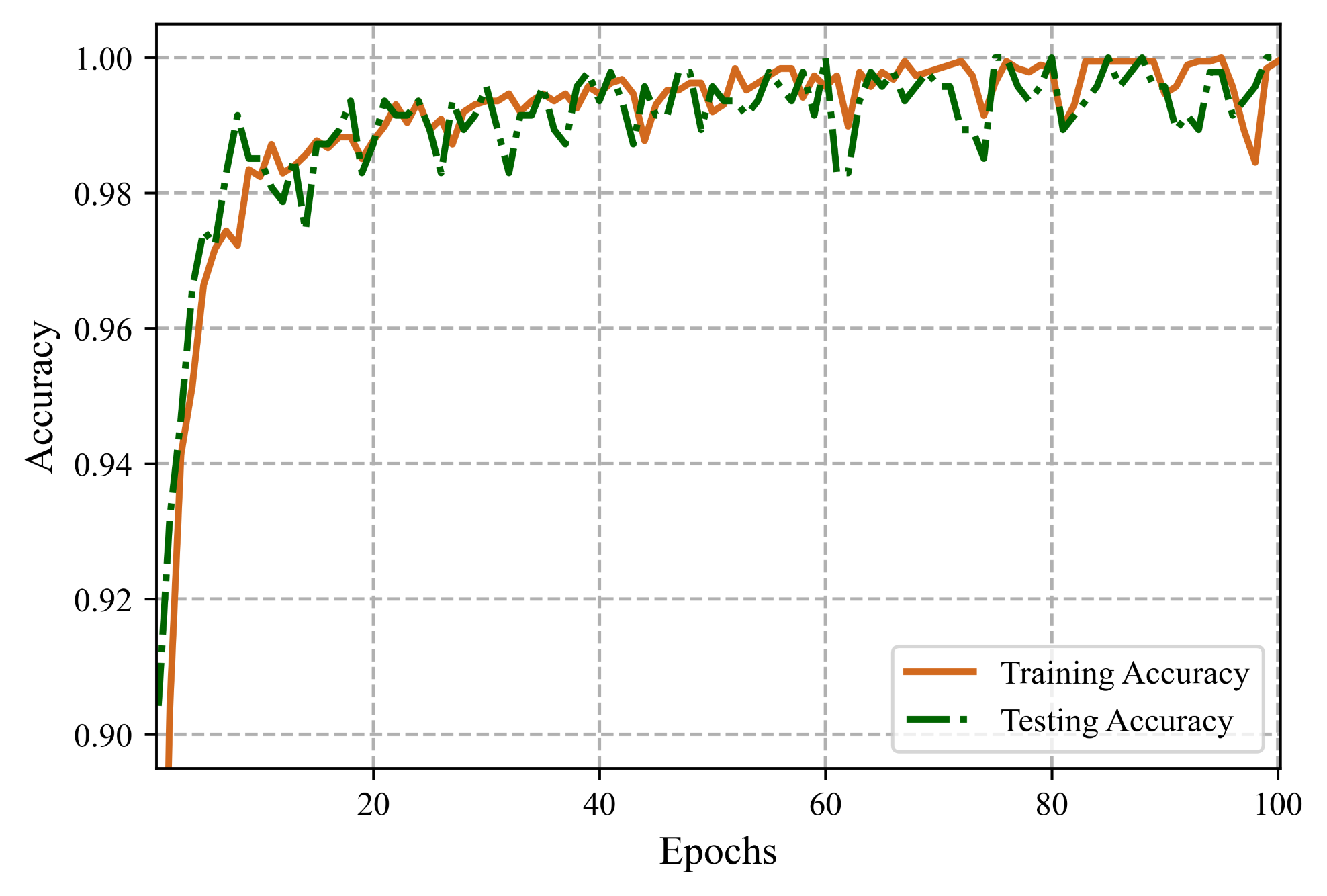}
        \caption{Training and testing accuracy vs. epoch for MLP classifier.}
        \label{fig:epoch}
    \end{subfigure}
    \caption{Performance evaluation of the MLP classifier.}
    \label{fig:mlp_class_perf}
\end{figure}

The confusion matrix for the best-performing MLP model, shown in Figure~\ref{fig:mlpcm}, provides a detailed view of its classification results. The matrix has two rows: the first represents the true 'Superior' class, and the second represents the true 'Inferior' class. The diagonal elements indicate correct classifications, while the off-diagonal elements represent misclassifications.

In the MLP model, the top-left element shows that all superior instances were correctly classified, while the bottom-right element indicates that all inferior instances were correctly identified. The MLP model demonstrates strong performance with zero false positives and false negatives, which is reflected in its perfect accuracy, precision, and recall for the 'Superior' class. MLP outperformed other models in all metrics, making it the top choice for this task. 

Regarding training and testing accuracy, the MLP model showed consistent improvement across epochs, as shown in Figure~\ref{fig:epoch}. While the training accuracy steadily increased, it did not exceed 90\% by the 90th epoch, which suggests that the model was not overfitting. On the other hand, the testing accuracy reached 100\% multiple times throughout the epochs, indicating that the model generalized well to unseen data. This behavior is indicative of a well-regularized model that avoids overfitting and performs exceptionally well on both training and testing datasets. SVM, DT, and RF also performed well, with high accuracy and recall, while AdaBoost and LR showed relatively lower effectiveness in identifying superior instances.

The parameters used in the MLP model are displayed in Table~\ref{tab:mlp}. The model has 3 hidden layers with 64, 32, and 16 units, using ReLU activation. The output layer uses softmax for binary-class classification. The optimizer is Adam, and the loss function is sparse categorical cross-entropy. The model was trained with a batch size of 16 for 100 epochs, with an 80-20 train-test split. 

\begin{table}[tb]
\centering
\caption{The parameters used in the MLP classifier}
\label{tab:mlp}
\begin{tabular}{ll}
\hline
\textbf{Parameter}        & \textbf{Value} \\ \hline
Input Dimensions          & 4              \\
Input Features            & $x, N_{ABS}, T_{ABS}, N_{ETL}$ \\ 
Hidden Layers             & 3              \\ 
Units per Hidden Layer    & 64, 32, 16 \\ 
\multirow{2}{*}{Activation Function}       & ReLU (hidden layers)\\
& Softmax (output) \\ 
Optimizer                 & Adam           \\ 
Loss Function             & Categorical Crossentropy \\
Batch Size                & 16             \\
Epochs                   & 100           \\
Data Split       & 80\% training, 20\% testing \\ \hline
\end{tabular}
\end{table}

\subsection{Performance Comparison with Prior Work}

\begin{table*}[tb]
\centering
\caption{Regression and classification performance of existing work.}
\resizebox{\linewidth}{!}{
\begin{tabular}{c c c c c c c c}
\hline
\multirow{2}{*}{\textbf{Targets}} & \multirow{2}{*}{\textbf{Features}} & \multirow{2}{*}{\textbf{ML Model}} & \multirow{2}{*}{\textbf{$R^2$/Accuracy}} & \multirow{2}{*}{\textbf{Samples}} & \multirow{2}{*}{\textbf{Train:Test}} & \multirow{2}{*}{\textbf{PSC architecture}} & \multirow{2}{*}{\textbf{References}} \\ 
& & & & & & \\ \hline
\multirow{4}{*}{$J_{SC}$} & 3 & NN & 0.8378 & 280 & 90:10 & $ITO/PEDOT:PSS/MASn_xPb_{1-x}I_3/C_{60}/BCP/Ag$ & \cite{cai2022data}  \\
 & 33 & RF & 0.6262 & 60 & - & $FTO/SnO_2/FAPbI_3/Spiro-OMeTAD/Au$ & \cite{hasanzadeh2023scaps} \\ 
 & 8 & MLP & 1 & 20,480 & 75:25 & $FTO/TiO_2/MAPbI_3/SWCNTs/m-SWCNTs$ & \cite{malek2024machine} \\ 
 & 4 & DT & 1 & 1,650 & 80:20 & $FTO/TiO_2/MAPb_{1-x}Sb_{2x/3}I_3/MWCNTs$ & This work \\ \hline
\multirow{4}{*}{$V_{OC}$} & 3 & NN & 0.9024 & 280 & 90:10 & $ITO/PEDOT:PSS/MASn_xPb_{1-x}I_3/C_{60}/BCP/Ag$ & \cite{cai2022data} \\ 
 & 33 & RF & 0.8925 & 60 & - & $FTO/SnO_2/FAPbI_3/Spiro-OMeTAD/Au$ &  \cite{hasanzadeh2023scaps} \\
 & 8 & MLP & 0.9998 & 20,480 & 75:25 & $FTO/TiO_2/MAPbI_3/SWCNTs/m-SWCNTs$ & \cite{malek2024machine} \\
 & 4 & DT & 1 & 1,650 & 80:20 & $FTO/TiO_2/MAPb_{1-x}Sb_{2x/3}I_3/MWCNTs$ &  This work \\ \hline
\multirow{4}{*}{$FF$} & 3 & NN & 0.9096 & 280 & 90:10 & $ITO/PEDOT:PSS/MASn_xPb_{1-x}I_3/C_{60}/BCP/Ag$ &  \cite{cai2022data} \\ 
 & 33 & RF & 0.7753 & 60 & - & $FTO/SnO_2/FAPbI_3/Spiro-OMeTAD/Au$ & \cite{hasanzadeh2023scaps} \\
 & 8 & MLP & 0.9993 & 20,480 & 75:25 & $FTO/TiO_2/MAPbI_3/SWCNTs/m-SWCNTs$ & \cite{malek2024machine} \\ 
 & 4 & DT & 1 & 1,650 & 80:20 & $FTO/TiO_2/MAPb_{1-x}Sb_{2x/3}I_3/MWCNTs$ & This work \\ \hline
\multirow{4}{*}{$\eta$} & 3 & NN & 0.9026 & 280 & 90:10 & $ITO/PEDOT:PSS/MASn_xPb_{1-x}I_3/C_{60}/BCP/Ag$ &  \cite{cai2022data} \\ 
 & 33 & RF & 0.8356 & 60 & - &  $FTO/SnO_2/FAPbI_3/Spiro-OMeTAD/Au$ & \cite{hasanzadeh2023scaps} \\ 
 & 8 & MLP & 0.9995 & 20,480 & 75:25 & $FTO/TiO_2/MAPbI_3/SWCNTs/m-SWCNTs$ & \cite{malek2024machine} \\ 
 & 4 & DT & 1 & 1,650 & 80:20 & $FTO/TiO_2/MAPb_{1-x}Sb_{2x/3}I_3/MWCNTs$ & This work \\ \hline
\multirow{4}{*}{$\Delta/T_P$} & 9 & SVR & 0.683 & 4,018 & 80:20 & $FTO/TiO_2-c/TiO_2-mp/MAPbI_3/Spiro-OMeTAD/Au$ & \cite{alsulami2024application} \\ 
 & 67 & RF & 0.18 & 1,834 & 75:25 &  - & \cite{graniero2023challenge} \\ 
 & 24 & XGB & 0.751 & 1,541 & 80:20 & FTO/TiO$_2$-c/TiO$_2$-mp/Cs$_{0.1}$FA$_{0.7}$MA$_{0.2}$PbBr$_{1.5}$I$_{1.5}$/Spiro-OMeTAD/Ag & \cite{kusuma2025optimizing}  \\ 
 & 4 & MLP & 1 & 1,650 & 80:20 & $FTO/TiO_2/MAPb_{1-x}Sb_{2x/3}I_3/MWCNTs$ & This work \\ \hline
\multirow{4}{*}{Classifier} & 10 & CC & 0.932 & 2,226 & - & - & \cite{wang2024prediction} \\
 & 24 & RF & 0.9409 & 3,469 & 80:20 & - & \cite{talapatra2021machine} \\
 & 24 & XGB & 0.848 & 15,513 & 80:20 & FTO/TiO$_2$-c/TiO$_2$-mp/Cs$_{0.1}$FA$_{0.7}$MA$_{0.2}$PbBr$_{1.5}$I$_{1.5}$/Spiro-OMeTAD/Ag & \cite{kusuma2025optimizing} \\
 & 4 & MLP & 1 & 1,650 & 80:20 & $FTO/TiO_2/MAPb_{1-x}Sb_{2x/3}I_3/MWCNTs$ & This work \\ \hline
\end{tabular}}

\label{tab:comparison}
\end{table*}

The comparative analysis presented in Table~\ref{tab:comparison} highlights the superiority of the proposed work over existing studies across several key dimensions. First, in terms of input features, this work utilizes only four highly relevant parameters to predict all major performance and stability metrics of PSCs. In contrast, prior studies often employed significantly higher feature dimensions—up to 33 and 67 in \cite{hasanzadeh2023scaps} and \cite{graniero2023challenge}, respectively—leading to increased model complexity and potential overfitting.

Second, the proposed models exhibit perfect regression performance, achieving $R^2$ values of 1.0 for all essential photovoltaic metrics: $J_{SC}$, $V_{OC}$, FF, $\eta$, and degradation indicators (such as $\Delta$ or, $T_P$). This surpasses the performance of earlier models such as RF, support vector regression (SVR), and neural networks (NN) reported in \cite{hasanzadeh2023scaps}, \cite{alsulami2024application}, \cite{cai2022data}, and \cite{malek2024machine}, and underscores the robustness of the proposed approach in capturing device physics with minimal pre-processing and tuning. Notably, the Extreme Gradient Boosting (XGB) model in \cite{kusuma2025optimizing} achieved a solid $R^2$ of 0.751 for degradation prediction.

Third, in the case of stability classification, the proposed MLP model achieves 100\% accuracy using just four features, significantly outperforming existing works. For instance, \cite{kusuma2025optimizing} reported 84.8\% accuracy with 24 features, while the RF model in \cite{talapatra2021machine} achieved 94.09\% with 3,469 samples and 24 features. The Classifier Chain (CC) model in \cite{wang2024prediction} attained 93.2\% accuracy, yet still fell short of the perfect classification achieved in this work with fewer features. These results reinforce the efficiency of our approach in delivering perfect classification with minimal data input.

\begin{figure*}[tb]
    \centering
    \hspace*{\fill}
    \begin{subfigure}{.42\textwidth}
        \centering
        \includegraphics[width=\textwidth]{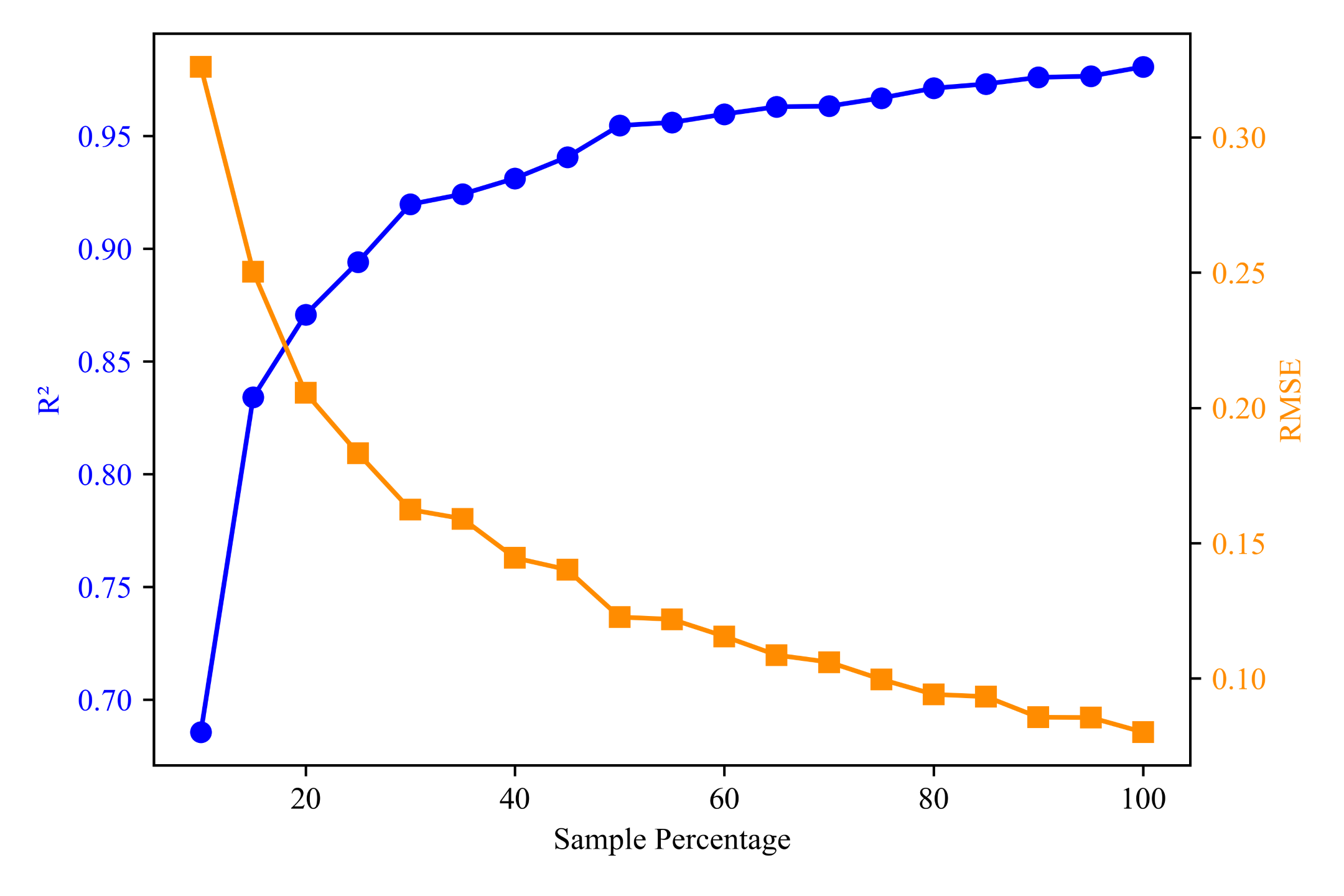}
        \caption{$R^2$ (primary y-axis) and RMSE (secondary y-axis) with increasing sample size using the DT regressor.}
        \label{fig:data_frac}
    \end{subfigure}
    \hfill
    \begin{subfigure}{.42\textwidth}
        \centering
        \includegraphics[width=\textwidth]{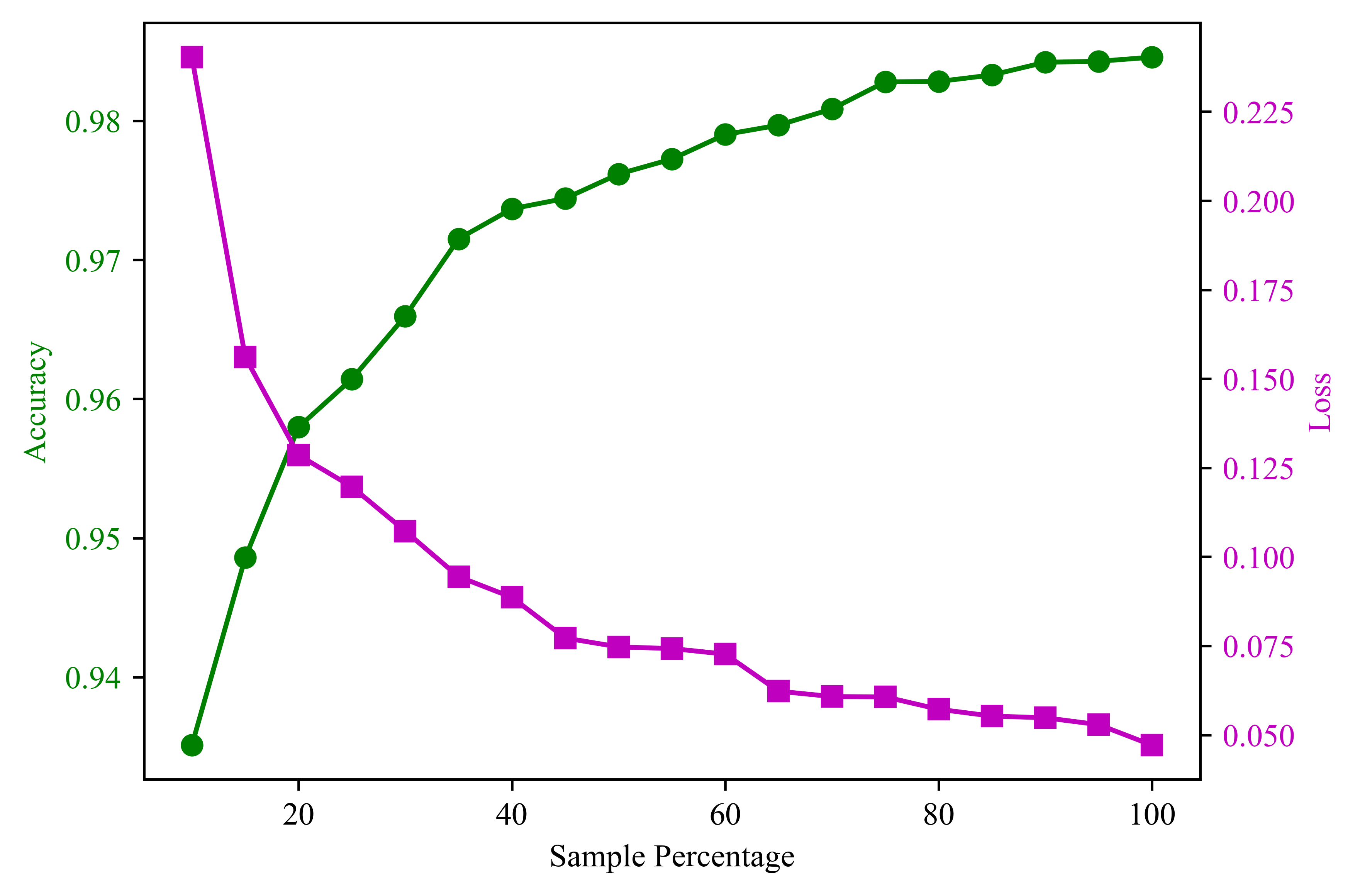}
        \caption{Accuracy (primary y-axis) and Loss (secondary y-axis) with increasing sample size using the RF classifier.}
        \label{fig:data_cls}
    \end{subfigure}
    \hspace*{\fill}
    \caption{Performance variation with sample size.}
    \label{fig:data_combined}
\end{figure*}

Moreover, while many prior models rely on substantially larger datasets (e.g., over 20,000 samples), the proposed models attain superior performance with only 1,650 samples. This demonstrates strong generalizability and data efficiency—particularly valuable in scenarios where data collection is costly or limited. 

Finally, by leveraging lightweight yet interpretable models such as DT and MLP, this work achieves both high accuracy and model transparency, in contrast to black-box or ensemble-based methods used in earlier studies. Collectively, these advantages position the proposed approach as a robust, scalable, and practical framework for modeling and stability assessment of perovskite solar cells.

\subsection{Possible Future Works}

Our findings indicate that advanced machine learning models, such as MLP and CNN, exhibited lower performance compared to traditional models like DT, likely due to the limitation of the dataset size. While this dataset size is sufficient for classical models like PR-4, which offer computational efficiency and accurate predictions, advanced models would benefit from larger datasets. For predicting $\eta$ and $\Delta$, we gradually varied the dataset size in 5\% increments, ranging from 10\% to 100\%, with the DT model selected as the baseline due to its relatively strong performance, as shown in Table~\ref{tab:ml_prfrm}. This process was repeated 10 times, and the average $R^2$ and RMSE values for each dataset fraction were computed and plotted in Figure~\ref{fig:data_frac}. Similarly, for stability classification, the RF model was evaluated using the same dataset fractions, with accuracy and loss plotted in Figure~\ref{fig:data_cls}. The RF model was chosen for its strong performance among classical ML models, as shown in Table~\ref{tab:cls}. Both figures show similar trends, where performance improves with increasing dataset size, but neither reveals a distinct inflection point for determining an optimal dataset fraction. This suggests that a significantly larger dataset is required to fully exploit the potential of more advanced deep learning models.

\begin{table}[tb]
    \centering
    \caption{RMSE and $R^2$ values for different feature combinations}
    \begin{tabular}{cccc}
        \hline
        Feature 1 & Feature 2 & RMSE & $R^2$ \\
        \hline
        \textit{$N_{ABS}$} & \textbf{$T_{ABS}$} & 0.0257 & 0.9944 \\
        $N_{ABS}$ & $N_{ETL}$ & 0.2194 & 0.5945 \\
        $x$ & $N_{ABS}$ & 0.2261 & 0.5691 \\
        $T_{ABS}$ & $N_{ETL}$ & 0.2854 & 0.3134 \\
        $x$ & $T_{ABS}$ & 0.2933 & 0.2749 \\
        $x$ & $N_{ETL}$ & 0.3542 & 0.0573 \\
        \hline
    \end{tabular}
    \label{tab:rmse_r2}
\end{table}

For simulation-based studies, we examined the influence of different features on the prediction of $\Delta$ by selecting two at a time from the four-dimensional input feature space, resulting in six feature combinations listed in Table~\ref{tab:rmse_r2}. The results show that the best R² and RMSE values were obtained for \(N_{ABS}\) and \(T_{ABS}\), suggesting that physics-based simulation techniques such as molecular dynamics or density functional theory should focus on the absorber layer. Simulating the entire device is computationally expensive, making a targeted approach more practical for capturing the most significant effects.

In addition, new combinations of materials, device architectures, and interlayers can be explored to expand the dataset space. Enhanced feature dimensions may include detailed material compositions and layer-specific physical parameters, which could further improve model accuracy and interpretability. Furthermore, future work may focus on generating datasets from experimental measurements rather than relying solely on simulation-based data. This would allow machine learning models to better capture real-world variability and device performance, ultimately enabling more reliable predictions and practical design insights.

The PCE and degradation can be explored for each pair of weight values \(w_{\eta}\) and \(w_{\Delta}\). As the weight for efficiency (\(w_{\eta}\)) increases and that for stability (\(w_{\Delta}\)) decreases, the performance (\(\eta\) and \(\Delta\)) evolve accordingly. The Table~\ref{tab:w1w2} shows the detailed results for each configuration. From the table, it is evident that as \(w_{\eta}\) increases, \(\eta\) improves, whereas a decrease in \( w_{\Delta} \) leads to an enhancement in \( \Delta \). Accordingly, the fabrication parameters (\(x, N_{\mathrm{ABS}}, T_{\mathrm{ABS}}, N_{\mathrm{ETL}}\)) also vary based on Eq. 15, which in turn affects the overall fabrication cost. This interdependence highlights the importance of selecting suitable weights that balance performance with economic considerations for specific application needs. By selecting appropriate values of \(w_{\eta}\) and \(w_{\Delta}\), it is possible to optimize efficiency, stability, and cost, facilitating the commercialization of perovskite solar cells.

\begin{table}[tb]
    \centering
    \caption{Performance for each weight pair}
    \begin{tabular}{cccc}
        \hline
        \textbf{$w_{\eta}$} & \textbf{$w_{\Delta}$} & \textbf{$\eta (\%)$} & \textbf{$\Delta (\%)$} \\
        \hline
        0.25 & 0.75 & 14.93 & 0.19 \\
        0.30 & 0.70 & 15.57 & 0.25 \\
        0.35 & 0.65 & 15.99 & 0.29 \\
        0.40 & 0.60 & 16.35 & 0.35 \\
        0.45 & 0.55 & 16.68 & 0.41 \\
        0.55 & 0.45 & 17.04 & 0.49 \\
        0.60 & 0.40 & 17.07 & 0.52 \\
        0.65 & 0.35 & 17.10 & 0.54 \\
        0.70 & 0.30 & 17.11 & 0.55 \\
        0.75 & 0.25 & 17.12 & 0.56 \\
        \hline
    \end{tabular}
    \label{tab:w1w2}
\end{table}

\section{Conclusion}

Carbon nanotube-based PSCs show great promise for achieving higher PCE and improved stability. Further optimization of device architecture and fabrication parameters could additionally enhance their performance. SCAPS has proven to be a valuable tool for simulating CNT-PSC devices, with its parameters validated against experimental data by matching both characteristic and stability curves. By varying key fabrication parameters, a dataset can be generated to establish relationships between these parameters and the primary targets—efficiency and degradation.

To model these relationships, a fourth-order polynomial regressor shows robust performance, providing explicit, continuous, and differentiable equations for efficiency and degradation as functions of the fabrication parameters. Using a weighted sum of these two targets as an objective function, optimization can be performed using the L-BFGS-B method, which led to an efficiency boost to \textbf{16.84\%} while reducing degradation to \textbf{2.39\%}. Through this optimization approach, a balance between high efficiency and long-term stability is possible to achieve, demonstrating the potential of data-driven modeling and optimization in advancing CNT-based PSC technology.

\appendix
\section{Appendix}

The following equations represent the trained fourth-order polynomial regression models for the target variables $\eta$ and $\Delta$ as functions of $x$, $N_{\text{ABS}}$, $T_{\text{ABS}}$, and $N_{\text{ETL}}$. These equations were derived from the regression model after training and provide an explicit mathematical representation of the learned relationships (thickness in $\mu$m). To evaluate the target values using these equations, MinMax-normalized inputs are required.

\begin{figure}[tb]
    \renewcommand{\thefigure}{A1} % Set custom figure number
    \centering
    \begin{subfigure}{0.9\linewidth}
        \includegraphics[width=\textwidth]{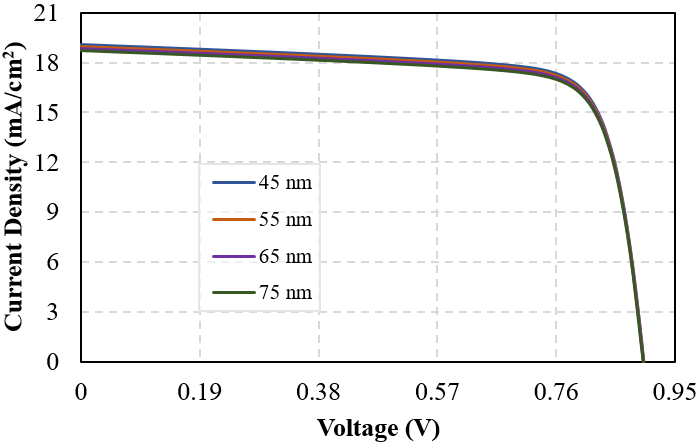}
        \caption{J-V curve with varying ETL thickness.}
        \label{fig:jv_tetl}
    \end{subfigure}
    \begin{subfigure}{0.9\linewidth}
        \includegraphics[width=\textwidth]{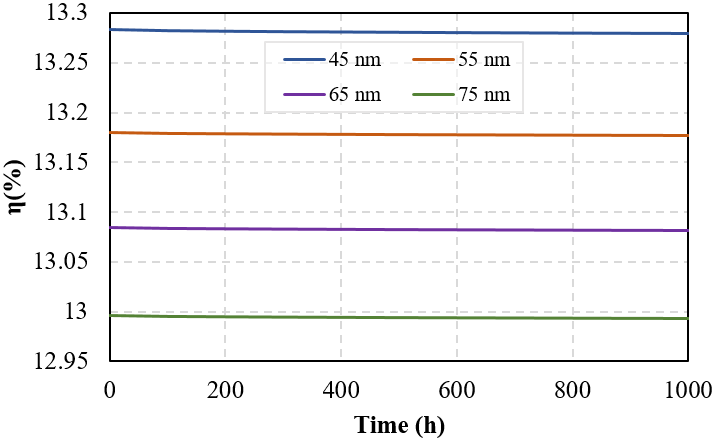}
        \caption{Stability curve with varying ETL thickness.}
        \label{fig:stab_tetl}
    \end{subfigure}
    \caption{Performance curves with varying ETL thickness.}
    \label{fig:tetl}
\end{figure}

\begin{figure}[tb]
    \renewcommand{\thefigure}{A2} 
    \includegraphics[width=\linewidth]{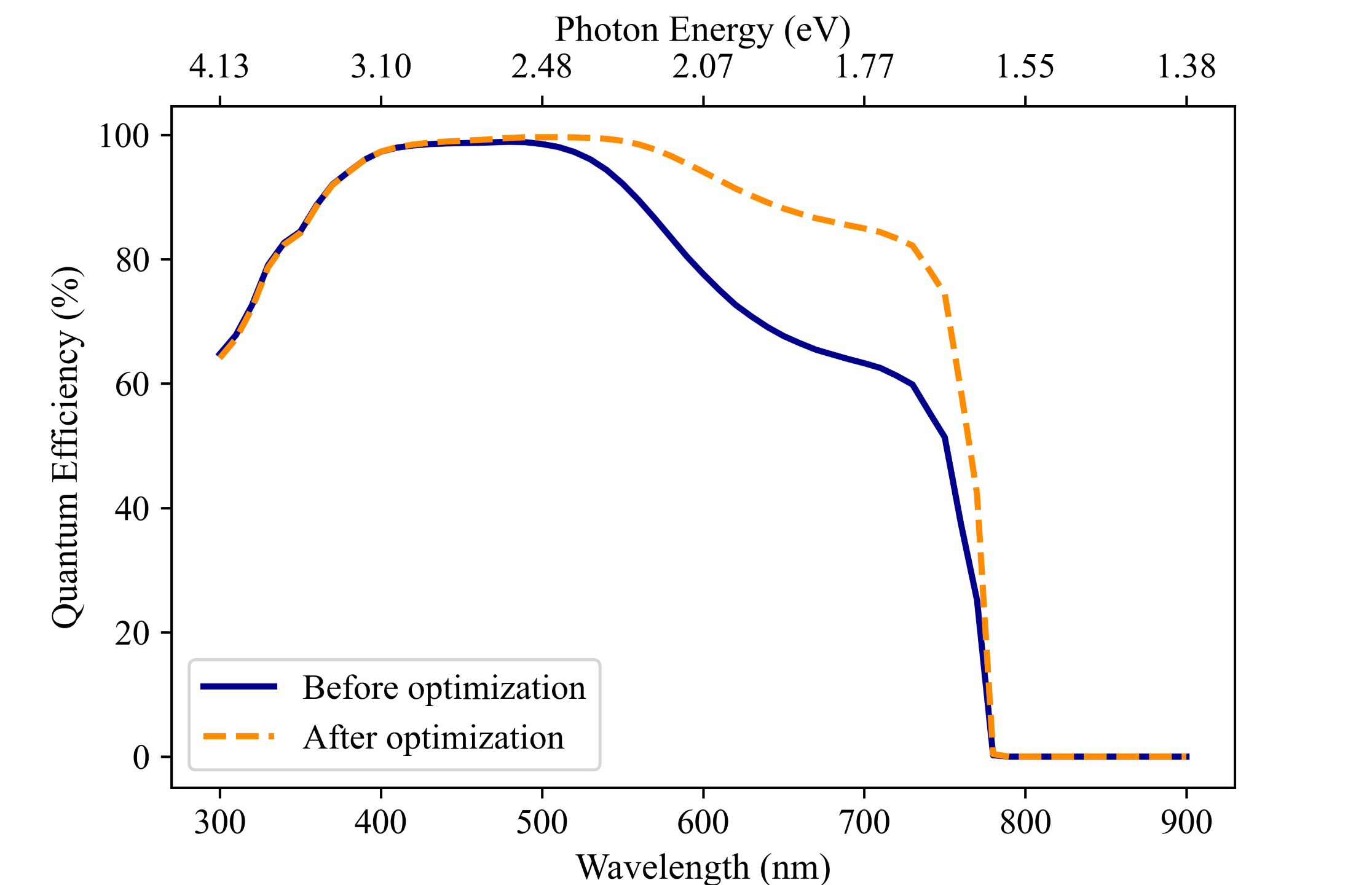}
    \caption{Quantum efficiency vs. wavelength and photon energy before and after optimization.}
    \label{fig:qe}
\end{figure}

$\eta (x, N_{ABS}, T_{ABS}, N_{ETL}) = 0.102211 + 0.0648 \times \Big(
0.0005 - 6.29 \times 10^{-13} \cdot 1 + 7.37 \times 10^{-2} x - 8.27 \times 10^{-4} N_{ABS} + 6.02 \times 10^{-1} T_{ABS} + 6.15 \times 10^{-1} N_{ETL} - 1.51 \times 10^{-1} x^2 + 3.52 \times 10^{-4} x N_{ABS} + 4.93 \times 10^{-2} x T_{ABS} + 5.39 \times 10^{-1} x N_{ETL} + 4.91 \times 10^{-5} N_{ABS}^2 - 1.76 \times 10^{-3} N_{ABS} T_{ABS} - 5.27 \times 10^{-4} N_{ABS} N_{ETL} - 3.32 \times 10^{-1} T_{ABS}^2 + 1.57 \times 10^{-1} T_{ABS} N_{ETL} - 4.86 \times 10^{-1} N_{ETL}^2 + 4.77 \times 10^{-2} x^3 - 4.51 \times 10^{-4} x^2 N_{ABS} - 4.58 \times 10^{-2} x^2 T_{ABS} - 3.69 \times 10^{-1} x^2 N_{ETL} + 6.67 \times 10^{-5} x N_{ABS}^2 - 1.75 \times 10^{-4} x N_{ABS} T_{ABS} - 4.21 \times 10^{-4} x N_{ABS} N_{ETL} - 3.26 \times 10^{-2} x T_{ABS}^2 + 2.27 \times 10^{-1} x T_{ABS} N_{ETL} + 8.41 \times 10^{-1} x N_{ETL}^2 - 9.02 \times 10^{-5} N_{ABS}^3 - 7.54 \times 10^{-5} N_{ABS}^2 T_{ABS} + 2.97 \times 10^{-5} N_{ABS}^2 N_{ETL} - 1.22 \times 10^{-3} N_{ABS} T_{ABS}^2 - 5.73 \times 10^{-4} N_{ABS} T_{ABS} N_{ETL} + 6.53 \times 10^{-4} N_{ABS} N_{ETL}^2 + 1.58 \times 10^{-1} T_{ABS}^3 - 8.18 \times 10^{-2} T_{ABS}^2 N_{ETL} - 2.02 \times 10^{-1} T_{ABS} N_{ETL}^2 - 4.48 \times 10^{-1} N_{ETL}^3 + 5.09 \times 10^{-2} x^4 - 5.09 \times 10^{-5} x^3 N_{ABS} - 8.74 \times 10^{-4} x^3 T_{ABS} - 3.05 \times 10^{-1} x^3 N_{ETL} - 4.53 \times 10^{-5} x^2 N_{ABS}^2 + 5.27 \times 10^{-4} x^2 N_{ABS} T_{ABS} + 1.63 \times 10^{-3} x^2 N_{ABS} N_{ETL} + 1.90 \times 10^{-2} x^2 T_{ABS}^2 - 3.43 \times 10^{-2} x^2 T_{ABS} N_{ETL} + 6.40 \times 10^{-1} x^2 N_{ETL}^2 - 1.16 \times 10^{-5} x N_{ABS}^3 - 1.24 \times 10^{-6} x N_{ABS}^2 T_{ABS} - 4.09 \times 10^{-6} x N_{ABS}^2 N_{ETL} + 1.66 \times 10^{-4} x N_{ABS} T_{ABS}^2 - 1.08 \times 10^{-3} x N_{ABS} T_{ABS} N_{ETL} - 1.52 \times 10^{-3} x N_{ABS} N_{ETL}^2 + 8.22 \times 10^{-3} x T_{ABS}^3 - 3.87 \times 10^{-2} x T_{ABS}^2 N_{ETL} - 5.53 \times 10^{-2} x T_{ABS} N_{ETL}^2 - 1.02 x N_{ETL}^3 + 4.49 \times 10^{-5} N_{ABS}^4 + 3.32 \times 10^{-5} N_{ABS}^3 T_{ABS} - 1.55 \times 10^{-5} N_{ABS}^3 N_{ETL} + 4.57 \times 10^{-5} N_{ABS}^2 T_{ABS}^2 - 4.09 \times 10^{-5} N_{ABS}^2 T_{ABS} N_{ETL} + 2.31 \times 10^{-5} N_{ABS}^2 N_{ETL}^2 + 1.24 \times 10^{-4} N_{ABS} T_{ABS}^3 - 6.32 \times 10^{-4} N_{ABS} T_{ABS}^2 N_{ETL} + 9.73 \times 10^{-4} N_{ABS} T_{ABS} N_{ETL}^2 + 1.61 \times 10^{-4} N_{ABS} N_{ETL}^3 - 4.15 \times 10^{-2} T_{ABS}^4 + 1.52 \times 10^{-2} T_{ABS}^3 N_{ETL} + 4.70 \times 10^{-2} T_{ABS}^2 N_{ETL}^2 + 6.00 \times 10^{-2} T_{ABS} N_{ETL}^3 + 4.94 \times 10^{-1} N_{ETL}^4
\Big)$

$\Delta (x, N_{ABS}, T_{ABS}, N_{ETL}) = 0.03385 + 1.51773 \times \Big(-4.39 \times 10^{-4} + 7.16 \times 10^{-13} \cdot 1 + 7.38 \times 10^{-3} x - 1.06 \times 10^{-5} N_{ABS} + 4.07 \times 10^{-3} T_{ABS} - 5.14 \times 10^{-3} N_{ETL} - 6.25 \times 10^{-4} x^2 - 4.19 \times 10^{-7} x N_{ABS} + 3.04 \times 10^{-4} x T_{ABS} - 2.47 \times 10^{-3} x N_{ETL} + 2.14 \times 10^{-7} N_{ABS}^2 + 1.84 \times 10^{-6} N_{ABS} T_{ABS} - 1.63 \times 10^{-7} N_{ABS} N_{ETL} + 1.12 \times 10^{-4} T_{ABS}^2 - 2.31 \times 10^{-5} T_{ABS} N_{ETL} + 3.68 \times 10^{-4} N_{ETL}^2  5.45 \times 10^{-5} x^3 - 1.37 \times 10^{-7} x^2 N_{ABS} - 4.51 \times 10^{-6} x^2 T_{ABS} + 1.16 \times 10^{-5} x^2 N_{ETL} + 2.23 \times 10^{-8} x N_{ABS}^2 - 1.02 \times 10^{-7} x N_{ABS} T_{ABS} + 6.91 \times 10^{-8} x N_{ABS} N_{ETL} + 1.01 \times 10^{-6} x T_{ABS}^2 - 3.15 \times 10^{-7} x T_{ABS} N_{ETL} - 1.57 \times 10^{-6} x N_{ETL}^2 + 1.07 \times 10^{-8} N_{ABS}^3 + 1.96 \times 10^{-9} N_{ABS}^2 T_{ABS} - 2.90 \times 10^{-10} N_{ABS}^2 N_{ETL} - 1.67 \times 10^{-8} N_{ABS} T_{ABS}^2 + 3.22 \times 10^{-9} N_{ABS} T_{ABS} N_{ETL} - 1.67 \times 10^{-8} N_{ABS} N_{ETL}^2 - 4.13 \times 10^{-7} T_{ABS}^3 + 1.09 \times 10^{-7} T_{ABS}^2 N_{ETL} + 1.42 \times 10^{-7} T_{ABS} N_{ETL}^2 + 7.54 \times 10^{-7} N_{ETL}^3 + 2.47 \times 10^{-7} x^4 + 2.28 \times 10^{-9} x^3 N_{ABS} + 5.81 \times 10^{-9} x^3 T_{ABS} - 2.96 \times 10^{-8} x^3 N_{ETL} + 3.35 \times 10^{-10} x^2 N_{ABS}^2 + 1.24 \times 10^{-9} x^2 N_{ABS} T_{ABS} + 1.74 \times 10^{-9} x^2 N_{ABS} N_{ETL} - 1.32 \times 10^{-8} x^2 T_{ABS}^2 + 3.58 \times 10^{-9} x^2 T_{ABS} N_{ETL} + 4.03 \times 10^{-8} x^2 N_{ETL}^2 - 2.08 \times 10^{-11} x N_{ABS}^3 - 1.11 \times 10^{-11} x N_{ABS}^2 T_{ABS} - 5.84 \times 10^{-12} x N_{ABS}^2 N_{ETL} + 7.86 \times 10^{-11} x N_{ABS} T_{ABS}^2 + 6.33 \times 10^{-11} x N_{ABS} T_{ABS} N_{ETL} - 1.04 \times 10^{-10} x N_{ABS} N_{ETL}^2 - 6.41 \times 10^{-10} x T_{ABS}^3 + 1.80 \times 10^{-10} x T_{ABS}^2 N_{ETL} + 1.92 \times 10^{-10} x T_{ABS} N_{ETL}^2 + 6.33 \times 10^{-10} x N_{ETL}^3 - 2.19 \times 10^{-9} N_{ABS}^4 - 1.33 \times 10^{-10} N_{ABS}^3 T_{ABS} + 7.00 \times 10^{-11} N_{ABS}^3 N_{ETL} - 1.22 \times 10^{-9} N_{ABS}^2 T_{ABS}^2 + 7.94 \times 10^{-10} N_{ABS}^2 T_{ABS} N_{ETL} - 1.41 \times 10^{-9} N_{ABS}^2 N_{ETL}^2 + 5.08 \times 10^{-9} N_{ABS} T_{ABS}^3 - 1.79 \times 10^{-9} N_{ABS} T_{ABS}^2 N_{ETL} + 1.65 \times 10^{-9} N_{ABS} T_{ABS} N_{ETL}^2 + 1.15 \times 10^{-9} N_{ABS} N_{ETL}^3 + 1.32 \times 10^{-8} T_{ABS}^4 - 4.06 \times 10^{-9} T_{ABS}^3 N_{ETL} - 1.16 \times 10^{-8} T_{ABS}^2 N_{ETL}^2 - 1.39 \times 10^{-8} T_{ABS} N_{ETL}^3 - 1.08 \times 10^{-7} N_{ETL}^4
\Big)$

The variation in J-V characteristic curves and stability curves with changes in $T_{ETL}$ is shown in Figure~\ref{fig:tetl}. In Figure~\ref{fig:qe}, the quantum efficiency as a function of wavelength and photon energy before and after optimization is shown.

\printcredits

\section*{Acknowledgments}
The authors acknowledge the facility and the support provided by the Department of Electrical and Electronic Engineering at Bangladesh University of Engineering and Technology. 

\section*{Data Availability}
All relevant data that support the findings of this study are presented in the manuscript. The source data are available on GitHub and can be accessed at \href{https://zenodo.org/records/15478813}{this link}.

\section*{Declaration of Competing Interest}

The authors declare that they have no known competing financial interests or personal relationships that could have influenced the work reported in this paper.

\bibliographystyle{model1-num-names}

\bibliography{cas-refs}

\balance


\end{document}